\documentclass[pdflatex,sn-mathphys]{sn-jnl}% Math and Physical Sciences Reference Style
\jyear{2021}%

\theoremstyle{thmstyleone}%
\theoremstyle{thmstyletwo}%

\theoremstyle{thmstylethree}%

\usepackage{graphicx}
\graphicspath{ {./images/} }
\usepackage{caption}
\usepackage{subcaption}
\usepackage[section]{placeins}
\usepackage{amsmath}

\begin{document}

\title[ ]{Identifying the Leading Factors of Significant Weight Gains Using a New Rule Discovery Method}


\author[1]{\fnm{Mina} \sur{Samizadeh}}\email{minasmz@udel.edu}

\author[2]{\fnm{Jessica} \sur{C Jones-Smith}}\email{jjoness@uw.edu}

\author[3]{\fnm{Bethany} \sur{Sheridan}}\email{bethany.sheridan.g@gmail.com}
\author[1]{\fnm{Rahmatollah} \sur{Beheshti}}\email{rbi@udel.edu}


\affil[1]{ \orgname{University of Delaware},\orgaddress{ \state{Delaware}, \country{USA}}}

\affil[2]{\orgname{University of Washington},\orgaddress{ \state{Washington}, \country{USA}}}

\affil[3]{\orgname{athenahealth, Inc.},\orgaddress{ \state{Massachusetts}, \country{USA}}}

%\affil[3]{\orgdiv{Department}, \orgname{Organization}, \orgaddress{\street{Street}, \city{City}, \postcode{610101}, \state{State}, \country{Country}}}


\FloatBarrier

\abstract{Overweight and obesity remain a major global public health concern and identifying the individualized patterns that increase the risk of future weight gains has a crucial role in preventing obesity and numerous subsequent diseases associated with obesity. In this work, we use a rule discovery method to study this problem, by presenting  an approach that offers genuine interpretability and concurrently optimizes the accuracy (being correct often) and support (applying to many samples) of the identified patterns. Specifically, we extend an established subgroup-discovery method to generate the desired rules of type {\it X → Y}, and show how  top features can be extracted from the {\it X} side, functioning as the best predictors of {\it Y}. In our obesity problem, {\it X} refers to the extracted features from a very large  and multi-site EHR data, and {\it Y} indicates significant weight gains. Using our  method, we also extensively compare the differences and inequities in patterns across 22 strata determined by the individuals’ gender, age, race, insurance type, neighborhood type, and income level. Through extensive series of experiments, we show new and complementary findings regarding the predictors of future dangerous weight gains\footnote{Our code is available on our \href{https://github.com/minasmz/identifying_obesity_factors}{GitHub repository}.}}.


\FloatBarrier
%\begin{figure}[b]
%\centering
%\includegraphics[width=11cm]{images/graphical abstract 1.PNG}
%\caption{Overview}
%\label{Fig:graph_abs}
%\end{figure}



\maketitle

\section{Introduction}\label{sec1}

Obesity and overweight continue to prevail around the world. It is estimated that more than 30\% of the world population has been overweight or obese in recent years \citep{centers2020adult}, and the forecast trends do not show signs of dampening \citep{dobbs2014overcoming}. If the growth rates in body-weights stay similar, more than 41 \% of the world population are estimated to be overweight or obese by 2030 \citep{shaw2010global}. At the same time, the costs of the medical support for obesity management have been increasing constantly, for instance, from \$212.4 to \$315.8 billion during the 2005-10 period in the US, showing a 48.7\% spike \citep{biener2017high}. As a major risk factor of chronic diseases such as diabetes \citep{dandona2004inflammation}, cancer \citep{vucenik2012obesity,de2013obesity,barb2006adiponectin}, and cardiovascular diseases \citep{nakamura2014adipokines, ritchie2007link}, as well as infectious diseases such as what the world experienced with the Covid-19 disease \citep{dietz2020obesity,korakas2020obesity}, obesity puts a huge burden on the human societies. Several decades of research on obesity have shown that no single predictor of obesity may exist \citep{malik2013global,mokdad2003prevalence}. Instead, obesity is the product of complex and non-linear interactions among many different factors including biological, environmental, and social ones. 

Like other fields of biomedical research, the obesity field has also faced a potentially overhauling moment, by using large and diverse datasets to study this complex disease. Aligned with the precision medicine aims, applying advanced analytical methods on these large datasets can reveal the role of the complex risk factors of obesity based on the specific characteristics of individuals. One large and popular family of such analytical methods is the family of subgroup discovery (SD) methods \citep{herrera2011overview}, which is closely related to subtyping \citep{wang2009subtyping} and electronic phenotyping methods \citep{blundell2014beyond,hong2019developing,banda2018advances}. SD is a data mining technique that finds desired patterns in a dataset with respect to a certain target variable \citep{herrera2011overview}. This method is generally used to generate {\it if-then} rules of type {\it X → Y}, where {\it X} is a set of input (independent) variables, and {\it Y} is a fixed target (dependent) variable. In our study, {\it Y} indicates an upward shift in the obesity classes or developing obesity (such as transitioning from obese class 1 to obese class 2, or normal weight to obese). We collectively refer to our target {\it Y} variable, as the \textit{dangerous weight gain patterns}. What makes SD methods different from the existing predictive (often classification) and descriptive (often clustering) methods in the field, is their ability to combine both predictive and descriptive inductions. Specifically, these methods can be used to optimize a supervised prediction task (e.g., optimize the sensitivity or specificity of a classification task) and ensure a high support for the identified rules (i.e., the identified rules apply to many of the samples). 

In this study, we use an SD-based method to identify the individuals who experience significant body weight increases, using a longitudinal six-year dataset containing information from more than two million patients and around 13.5 million visits. Our dataset includes information related to demographic, socioeconomic, medication, and a series of major chronic conditions derived from adult population across various sites in the US. Specifically, we customize an established SD technique \citep{clark1989cn2} to generate a pool of desired {\it X → Y} rules, as introduced earlier. Using the generated rules and based a mechanism that we propose, we then determine the most important features of the individuals (from the features appearing on the left side or {\it X}) that can predict the dangerous weight gains. We further study the differences across these across six demographic and socioeconomic characteristics, capturing some of the important social determinants of health (SDOH). These six variables include gender, age, race, insurance type, neighborhood type, and income levels. This way, the specific contributions of our study are:

\begin{itemize}[\qquad z]
    \item[-]	We extend an interpretable SD method, used for identifying the predictive rules, to identify the key features that can predict the outcome of interest. While other SD methods exist in the literature, our method for using the extracted subgropus to identify the important predictive features  is new and applicable to any problem of this type. 
    \item[-] Our study identifies the roots of dangerous weight gains by analyzing a  large, semi-nationally representative, and longitudinal dataset. We show the top features predicting  dangerous weight gains i and compare those features in various stratified groups based on demographic and socioeconomic factors. Our results also demonstrate the inequities with respect to dangerous weight gains. 
\end{itemize}

\section{Related Work}\label{sec2}

\textbf{Similar work to study obesity} – Electronic health records (EHRs) have demonstrated a great potential for studying health problems, and obesity is no exception \citep{bailey2013multi} (a few of these types of studies are reviewed by Baer et al. \citep{baer2013using}).  Considering the common properties of EHR datasets (such as having large volumes and missing elements), machine learning and data mining methods have been among popular choices to work with these datasets, commonly to predict future obesity trends \citep{guptainterpretable,gupta2019obesity,dugan2015machine,zhang2009comparing,campbell2020identification}. From this latter category, a closely related set of studies to our work are the studies that have used association rule mining methods, which like our method involves generating {\it if-then} rules (but not necessarily with a fixed right side). Examples of such studies include the work by Kim et. al to predict the diseases related to obesity \citep{kim2017using} and the work of Hu et. al studying the composition of the gut microbial community in normal and obese adolescents \citep{hu2015obesity}. Among the studies directly comparable to ours, the work by Roth et. al uses multinomial logistic regression to identify community-level factors associated with overweight and obese BMI levels \citep{roth2014community}. In a similar study, Rifat et al. have used a naive Bayes classifier to identify the obesity risk factors using cross-sectional data from 259 individuals \citep{hossain2018prmt}. Besides the predictive models in the field, another group of studies, closely related to our work, use descriptive methods for extracting the characteristics of an outcome of interest. Clustering-based methods are common in this group. To name a few examples, clustering was used to find the groups of patients who could maintain the weight they had lost \citep{ogden2012cluster}, to cluster individuals based on  lifestyle risk factors in studying the links between the clusters and BMIs \citep{schuit2002clustering}, and  in another work to study the  characteristics of the clusters of  people with obesity \citep{green2016obese}. 
% While many studies that use predictive or descriptive methods are available in the literature, combined approaches using a method such as SD for studying obesity trends is rare.

\textbf{SD applications  } – The goal of SD methods is detecting the best subsets of a dataset having a certain property concerning a target feature. Because of having desirable properties such as interpretability, they have been widely used in biomedical domains. Example applications demonstrating the wide scope of their applications include detecting risk groups with coronary heart disease \citep{gamberger2002generating}, extracting useful information in diagnosis and prevention of brain diseases \citep{gamberger2007clinical}, finding important features contributing to cancers  \citep{tan2006data,trajkovski2007learning}, characterizing the patients who visit  emergency departments \citep{carmona2011evolutionary}, and interpreting the medical imaging results  \citep{schmidt2010interpreting}. 
% We are aware of any work using SD methods for studying weight gain or obesity patterns.

\section{Dataset}\label{sec3}
In this study, we use a deidentified dataset from athenahealth, which is a large provider of network-enabled services for hospital and ambulatory clients in the US.  Patients visiting  athenahealth providers are broadly representative of the outpatient visits in the US, when compared to national benchmarks provided by the National Ambulatory Medical Care Survey (NAMCS) \citep{CDC2017}. The dataset is collected from the EHRs of the patients aged 20+ years who had a visit to an athenahealth primary care provider from 2012–2017. Our study was approved by a local institutional review board at the University of Delaware. 

As EHRs generally contain very diverse and heterogeneous data types (such as demographics, measurements, medications, and tests), to reduce the dimension of the dataset, only a subset of  variables (that are known to have relationships with obesity) has been selected. These variables included 1) patient demographics (sex, age, race, and ethnicity), 2) socioeconomic status (type of insurance coverage, urban or rural residence, and median household income in the zip code), 3) provider type (MD, NP, PA, and RN), 4) measurements (systolic and diastolic blood pressure, hemoglobin A1c, and LDL cholesterol), 5) selected medications (from the five categories of antihypertension, antihyperlipidemic, antidepressant, antiobesity, and antidiabetic medications), and 6) selected diagnoses (prior and incident diagnosis of 18 major chronic conditions). Specifically, the 18 chronic conditions included 1) hypothyroidism, 2) stroke, 3) Alzheimer’s or dementia, 4) anemia, 5) asthma, 6) heart failure or ischemic heart disease or acute myocardial infarction (AMI), 7) benign prostatic hyperplasia (BPH), 8) chronic kidney disease (CKD), 9) cancer (breast, colon, prostate, endometrial, or lung), 10) depression, 11) diabetes, 12) hip or pelvic fracture or osteoporosis, 13) hyperlipidemia, 14) hypertension, 15) obesity, 16) rheumatic arthritis or osteoarthritis, 17) Atrial Fibrillation (AFib), and 18) Chronic Obstructive Pulmonary Disease (COPD).
To further reduce the dimensionality of the variables and prepare the dataset for extracting interpretable rules, all numerical values were categorized (and one-hot encoded) in our dataset by following the standard levels used for each variable in the field. Specifically, age was categorized into 10-year buckets, and the income-level into low, medium, and high, as determined by the American Community Survey (ACS) for each zip code. Three categories of low, normal, and high were defined for the systolic blood pressure (low as $\leq98$, high as $\geq166$, and normal as in between), for diastolic blood pressure (low as $\leq58$, high as $\geq 100$, and normal as in between), for hemoglobin A1c (low as $\leq 5$, high as $\geq 11.7$, and normal as in between), and for LDL cholesterol (low as $\leq 44$, high as $\geq 189$, and normal as in between). BMI values were categorized into six categories based on the CDC (Centers for Disease Control and Prevention) categorization, as underweight ($<18.5$), normal ($<25$), overweight ($<30$), and class 1–3 obesity ($<35$, $<40$, and 40 $\leq$). Visit-level (temporal) data was aggregated by calculating the \textit{mode} of the measurements and replacing the chronic condition incidences with zero (if never recorded) and one (if recorded at any visit). Additionally, since prior BMIs are strongly correlated with future BMIs, we omitted features relating to BMI to avoid identifying trivial patterns in our analysis.

Following the preprocessing steps described above, we ended up with 210 variables (including 128 medication variables) per patient.  From this dataset, we define the final cohort by including those patients who have at least two BMI recordings spanning over a minimum of two years. We then assign labels to the patients in the cohort, by considering anyone who develops (new) obesity or shifts upward in the obesity classes as class positive (indicated by {\it class=1}) and rest as class negative ({\it class=0}). We follow the standard definitions by CDC  for classifying adult body weight status \citep{CDCBMI}. This way of defining the positive class is specifically targeting the risks of obesity, either in the form of the new incidence or disease exacerbation, which we refer to as the dangerous weight gains and formally define as:

\begin{equation}
  Class =
    \begin{cases}
      $=1$, & \text{if ($BMI_{s}$ $<30$ AND $\exists BMI_{s'} \geq 30$)}\\
      $=1$, & \text{if ($30 \leq BMI_{s} < 35 $ AND $\exists BMI_{s'} \geq 35$)}\\
      $=1$, & \text{if ($35 \leq BMI_{s} < 40$ AND $\exists BMI_{s'} \geq 40$)}\\
      $=0$, & \text{Otherwise.}\\
    \end{cases} 
    \label{equation:bmi}
\end{equation}

\begin{table}[]
\centering
\caption{The number of individuals in the positive and negative classes across different demographic and socioeconomic variables in our cohort.}
\label{tab:my-table}
\begin{tabular}{l|lll}
Variable     & Strata                                                                                                         & \# of positive cases                                                                       & \# of negative cases                                                                             \\ \hline
Gender       & \begin{tabular}[c]{@{}l@{}}Women \\ Men\end{tabular}                                                           & \begin{tabular}[c]{@{}l@{}}10,331 \\ 7,533\end{tabular}                                    & \begin{tabular}[c]{@{}l@{}}205,683 \\ 152,269\end{tabular}                                       \\ \hline
Race         & \begin{tabular}[c]{@{}l@{}}Latino \\ White\\ African-American \\ Other Races\\ Race Unavailable\end{tabular}   & \begin{tabular}[c]{@{}l@{}}1,048 \\ 13,650 \\ 1,442 \\ 437\\ 1,287\end{tabular}            & \begin{tabular}[c]{@{}l@{}}2,284 \\ 264,839 \\ 33,741 \\ 10,203 \\ 26,885\end{tabular}           \\ \hline
Age          & \begin{tabular}[c]{@{}l@{}}Under-Thirty \\ Thirties \\ Forties \\ Fifties \\ Sixties \\ Seventies\end{tabular} & \begin{tabular}[c]{@{}l@{}}1,307 \\ 1,934 \\ 3,515 \\ 4,493 \\ 3,897 \\ 2,718\end{tabular} & \begin{tabular}[c]{@{}l@{}}14,907 \\ 28,864 \\ 57,898 \\ 89,946 \\ 89,396 \\ 76,941\end{tabular} \\ \hline
Neighborhood & \begin{tabular}[c]{@{}l@{}}Metro \\ Metro-Adjacent \\ Rural\end{tabular}                                       & \begin{tabular}[c]{@{}l@{}}13,572 \\ 3,134 \\ 1,158\end{tabular}                           & \begin{tabular}[c]{@{}l@{}}272,156 \\ 60,945 \\ 24,851\end{tabular}                              \\ \hline
Income       & \begin{tabular}[c]{@{}l@{}}Low-Income \\ Med-Income \\ High-Income\end{tabular}                                & \begin{tabular}[c]{@{}l@{}}9,017 \\ 5,389 \\ 3,458\end{tabular}                            & \begin{tabular}[c]{@{}l@{}}183,946 \\ 105,210 \\ 68,796\end{tabular}                             \\ \hline
Insurance    & \begin{tabular}[c]{@{}l@{}}Medicare \\ Medicaid \\ Commercial \\ Self-Pay\end{tabular}                         & \begin{tabular}[c]{@{}l@{}}6,376 \\ 1,477 \\ 9,444 \\ 540\end{tabular}                     & \begin{tabular}[c]{@{}l@{}}156,113 \\ 24,606 \\ 167,135 \\ 9,601\end{tabular}                    \\ \hline
Total        & All Patients                                                                                                   & 17,864                                                                                     & 357,952                                                                                         
\end{tabular}
\label{table:statistics}
\end{table}

\noindent 
where, $BMI_{s}$ refers to the first available (start) BMI recorded for a patient, and $BMI_{s’}$ refers to ``a'' recorded BMI, following a minimum two-year gap that belongs to a higher obesity class. If such BMI exists, we label that patient as positive and use their data from $BMI_{s}$ to (the first observed) $BMI_{s’}$. If none of the above conditions are met, the patient is labeled as negative, and their entire available data is used. This way, those who maintain their weights (by staying in the same body-weight class) are not considered in the positive class. We study the dangerous weight gain patterns across the entire cohort and in separate 22 strata determined by six demographic and socioeconomic status categories (sex, age, race/ethnicity, insurance type, residence type, and income). Table \ref{table:statistics} shows the number of patients with positive and negative cases in our cohort and each of these strata.

\section{Method}\label{sec2}
\begin{figure}
\centering
\includegraphics[width=11cm]{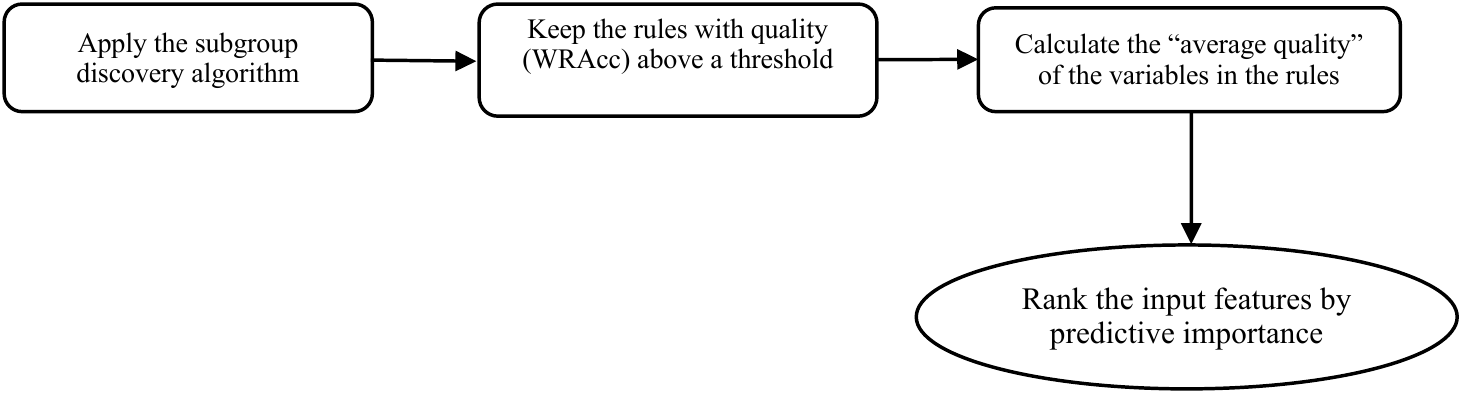}
\caption{Steps of the proposed procedure}
\label{Fig:steps}
\end{figure}

In this work, we adopted the CN2 SD algorithm \citep{clark1989cn2} for the initial extraction of rules. CN2  generats a random pool of {\it X → Y} rules (in our case, {\it Y} is the patients with {\it class=1}), and iteratively continues to improve those rules with respect to a certain quality measure using a heuristic search algorithm called beam search. Considering each generated rule as a graph node, beam search generates new rules to explore in a breadth-first-search manner. We refer the readers to other references for in-depth discussions of the concepts related to CN2 and relevant SD methods \citep{herrera2011overview}. For the quality measure, we use the weighted relative accuracy ({\it WRAcc}), as defined in Equation \ref{Equation:WRAcc},

\setlength{\belowdisplayskip}{0pt} \setlength{\belowdisplayshortskip}{0pt}
\setlength{\abovedisplayskip}{0pt} \setlength{\abovedisplayshortskip}{0pt}
\begin{equation}
WRAcc=Support \times (Confidence - Expected \  Confidence)
\label{Equation:WRAcc}
\end{equation}
\setlength{\belowdisplayskip}{0pt} \setlength{\belowdisplayshortskip}{0pt}
\setlength{\abovedisplayskip}{0pt} \setlength{\abovedisplayshortskip}{0pt}
\begin{equation}
Support = subgroup \ size / dataset \ size
\end{equation}
\setlength{\belowdisplayskip}{0pt} \setlength{\belowdisplayshortskip}{0pt}
\setlength{\abovedisplayskip}{0pt} \setlength{\abovedisplayshortskip}{0pt}
\begin{equation}
Confidence =  positive \ \# \ in \ subgroup / subgroup \ size
\end{equation}
\setlength{\belowdisplayskip}{0pt} \setlength{\belowdisplayshortskip}{0pt}
\setlength{\abovedisplayskip}{0pt} \setlength{\abovedisplayshortskip}{0pt}
\begin{equation}
Expected \ Confidence = positive \ \# \ in \ subgroup / dataset \ size
\end{equation}
\setlength{\belowdisplayskip}{0pt} \setlength{\belowdisplayshortskip}{0pt}
\setlength{\abovedisplayskip}{0pt} \setlength{\abovedisplayshortskip}{0pt}

\noindent
where {\it WRAcc} (of a rule {\it r}) is defined using the {\it support} and {\it confidence} quality measures. Considering the general format of { \it X → Y} for the rule {\it r}, one can consider {\it support} as {\it P(X)} (showing the probability of {\it X}), {\it confidence} as $P(Y \mid X)$, and {\it expected confidence} as {\it P(Y)}. What makes {\it WRAcc} especially relevant for our study is its ability to combine both generalizability ({\it support}) and accuracy ({\it Confidence – Expected Confidence}; the right side of Equation \ref{Equation:WRAcc}) that facilitates capturing both descriptive and predictive patterns across the dataset. The descriptive patterns are captured by the {\it support} (by ensuring that the patterns apply to many), and the predictive patterns are captured by the right side of Equation \ref{Equation:WRAcc}. This right side of the equation compares the accuracy of the rule with respect to a naive classification which considers all the cases as positive. This part normalizes the bias between the classes, since it depends on the proportion of the class that is being analyzed to the size of whole dataset. The higher value for {\it WRAcc} is interpreted as a higher balance between the generality of the pattern and its confidence, and as a result, a higher quality of the subgroup (identified by a rule). This formulation is also well suited for imbalanced datasets, such as ours, as it directly considers the size of each of the classes and adjusts the final value accordingly. After ranking all possible rules (with a certain length)  using {\it WRAcc}, we keep those rules with the {\it WRAcc} more than a threshold, set empirically in our main experiments. From these rules, we then identify the most important features appearing on the left side (of the {\it X → Y} rules), using an average quality approach. In this approach, for each feature, we calculate the weighted average ($A_{W}$) of the {\it WRAcc} of the remained rules (those with {\it WRAcc} above the threshold) in which the feature has appeared, as shown in Equation \ref{equation:WA}.
\setlength{\belowdisplayskip}{0pt} \setlength{\belowdisplayshortskip}{0pt}
\setlength{\abovedisplayskip}{0pt} \setlength{\abovedisplayshortskip}{0pt}
\begin{equation}
A_{W}(f_{i})=(\sum_{rules \ with \ f_{i}} WRAcc)/(\# \ of \ rules \ with \ f_{i})
\label{equation:WA}
\end{equation}
\setlength{\belowdisplayskip}{0pt} \setlength{\belowdisplayshortskip}{0pt}
\setlength{\abovedisplayskip}{0pt} \setlength{\abovedisplayshortskip}{0pt}

\noindent
where  $A_{W}$  shows the weighted average and $f_{i}$ is the $i${th} feature. A higher weighted average quality can indicate the higher importance of a feature in predicting {\it Y} (in our case, dangerous weight gains). A schematic representation of the method we use for generating the rules and identifying the important features from those is presented in Fig \ref{Fig:steps}.

\section{Experiments and Discussion}\label{sec2}
\begin{figure}
\centering
\includegraphics[width=6cm]
{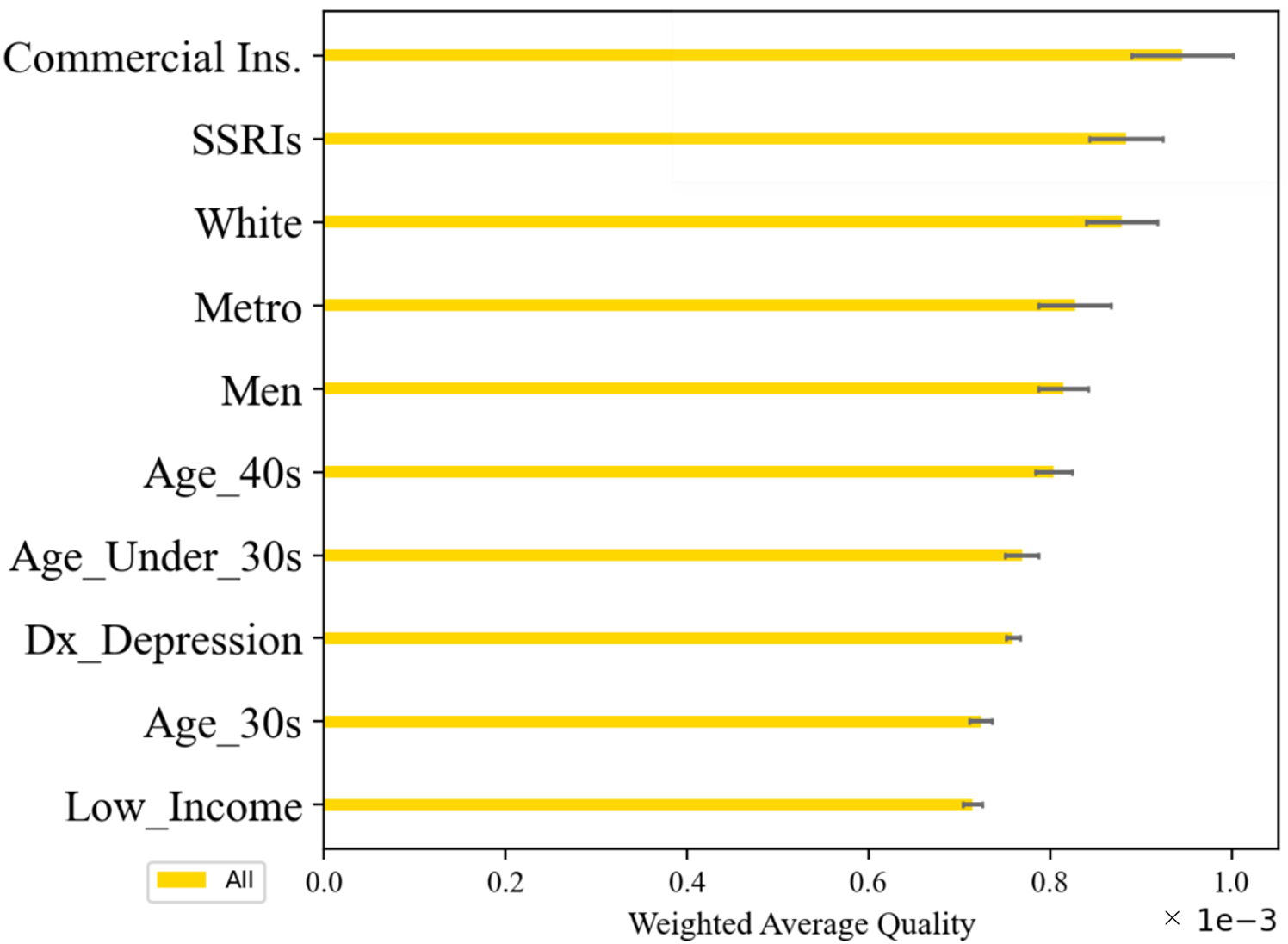}
\caption{Top 10 features with highest weighted average quality across the whole cohort. (Dx=diagnosis, SSRIs= “Selective Serotonin Reuptake Inhibitors”, an antidepression medicine, Ins=insurance, Metro=living in a metro area) }
\label{Fig:2}
\end{figure}

\begin{figure}
    %\captionsetup[subfigure]{farskip=0.1pt,captionskip=0.01pt}
  \begin{subfigure}{\linewidth}
      \centering
  \includegraphics[width=.25\linewidth]{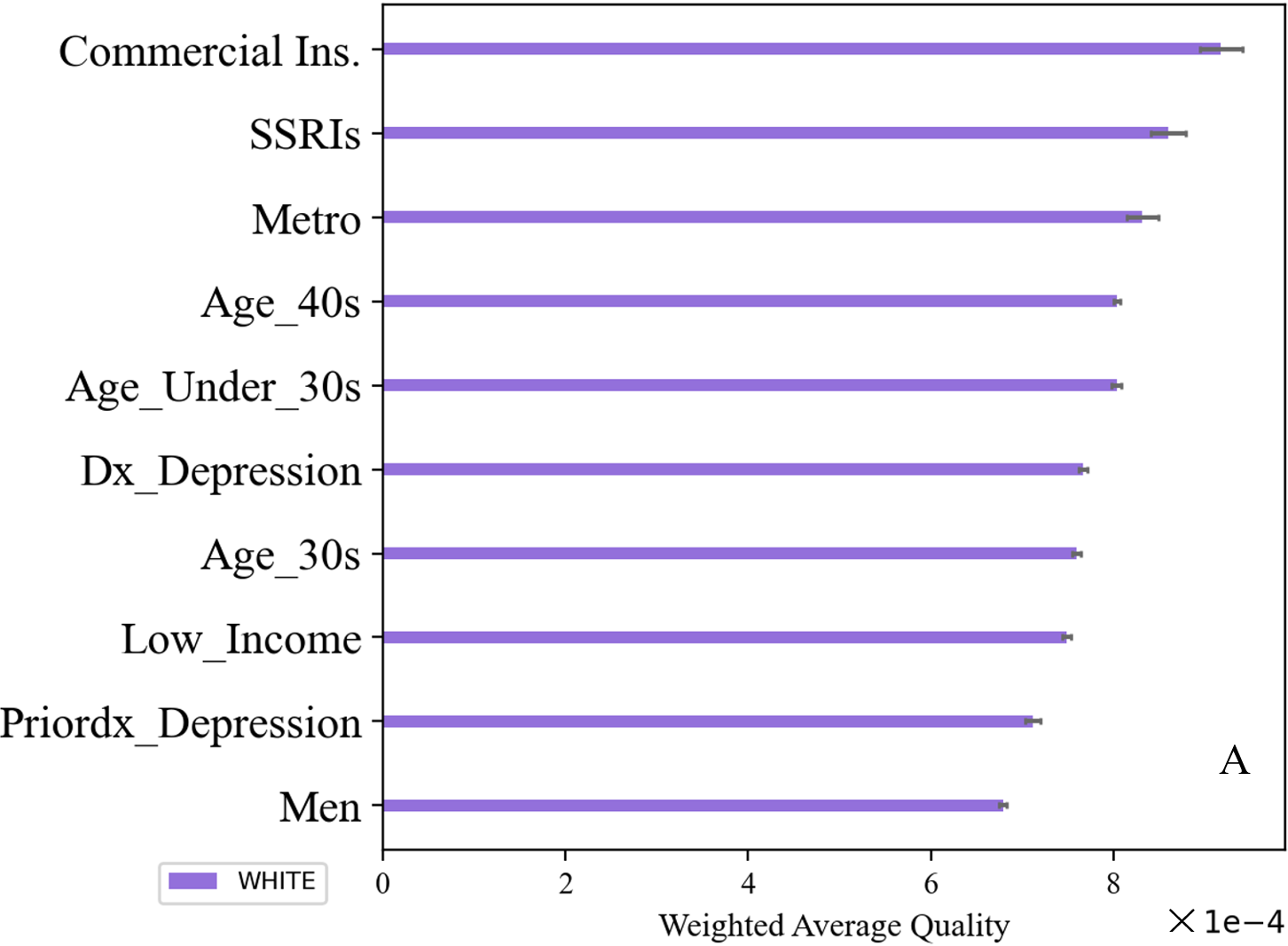}\hfill
  \includegraphics[width=.25\linewidth]{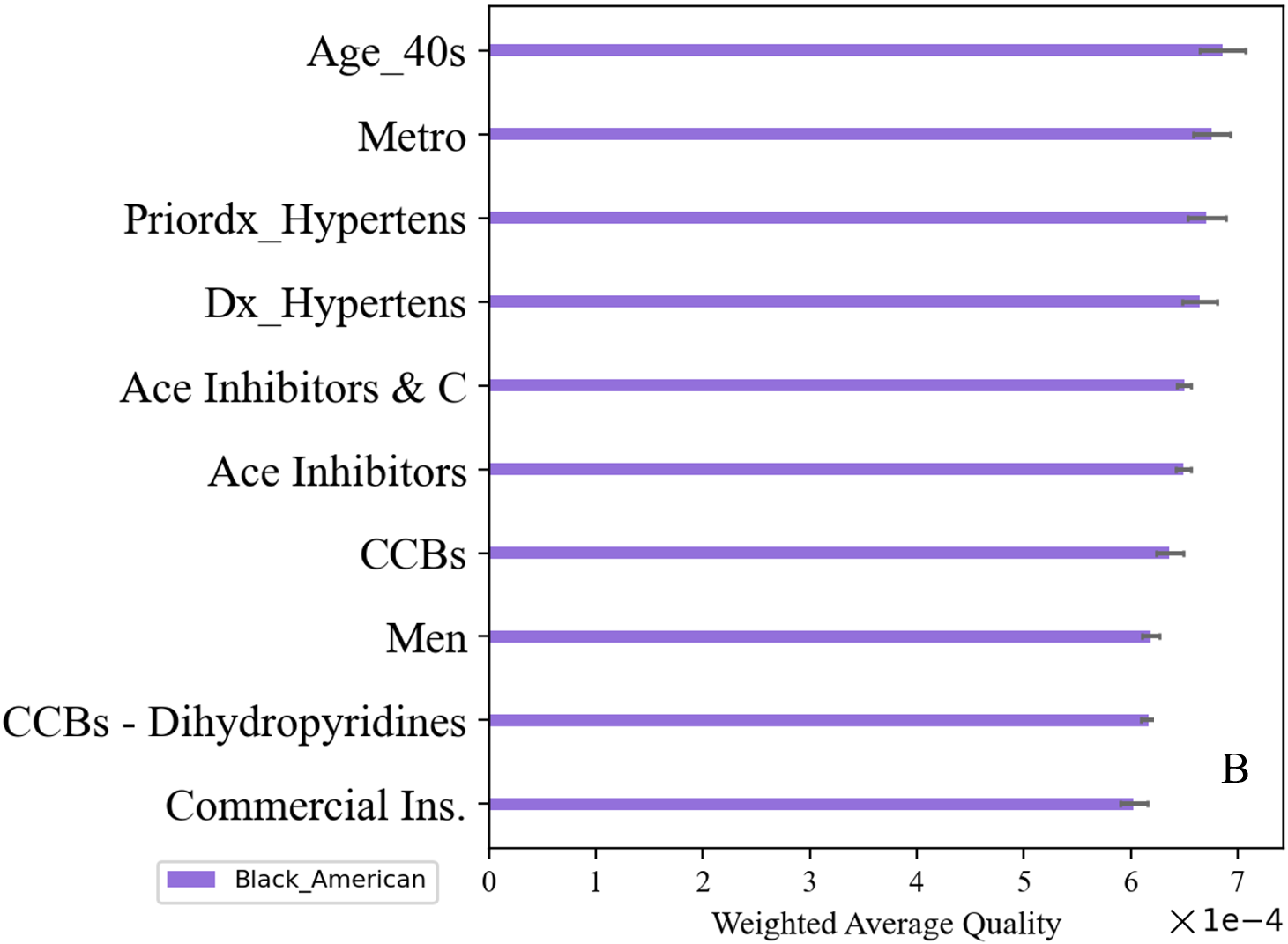}\hfill
  \includegraphics[width=.25\linewidth]{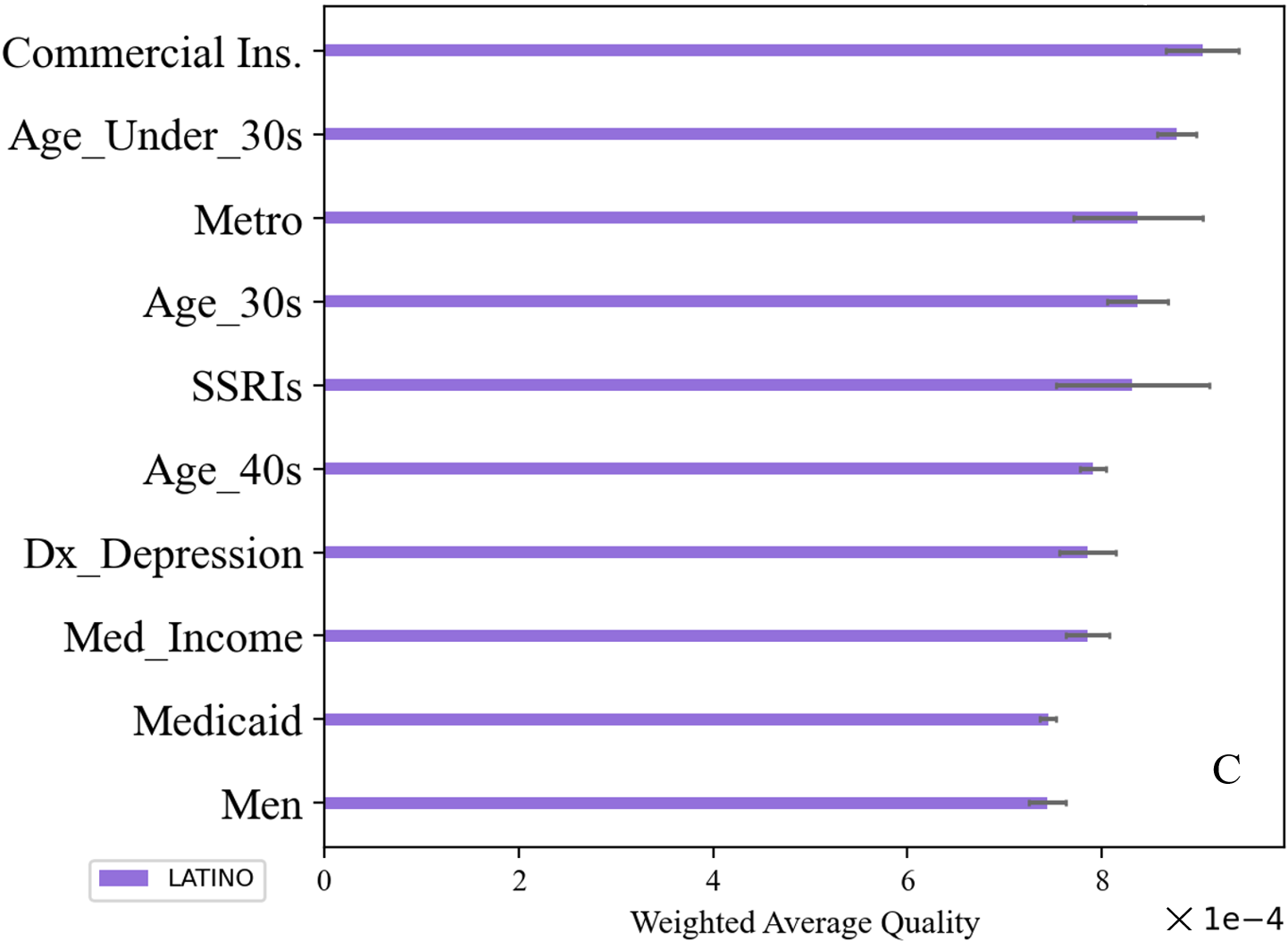}\hfill
  \includegraphics[width=.25\linewidth]{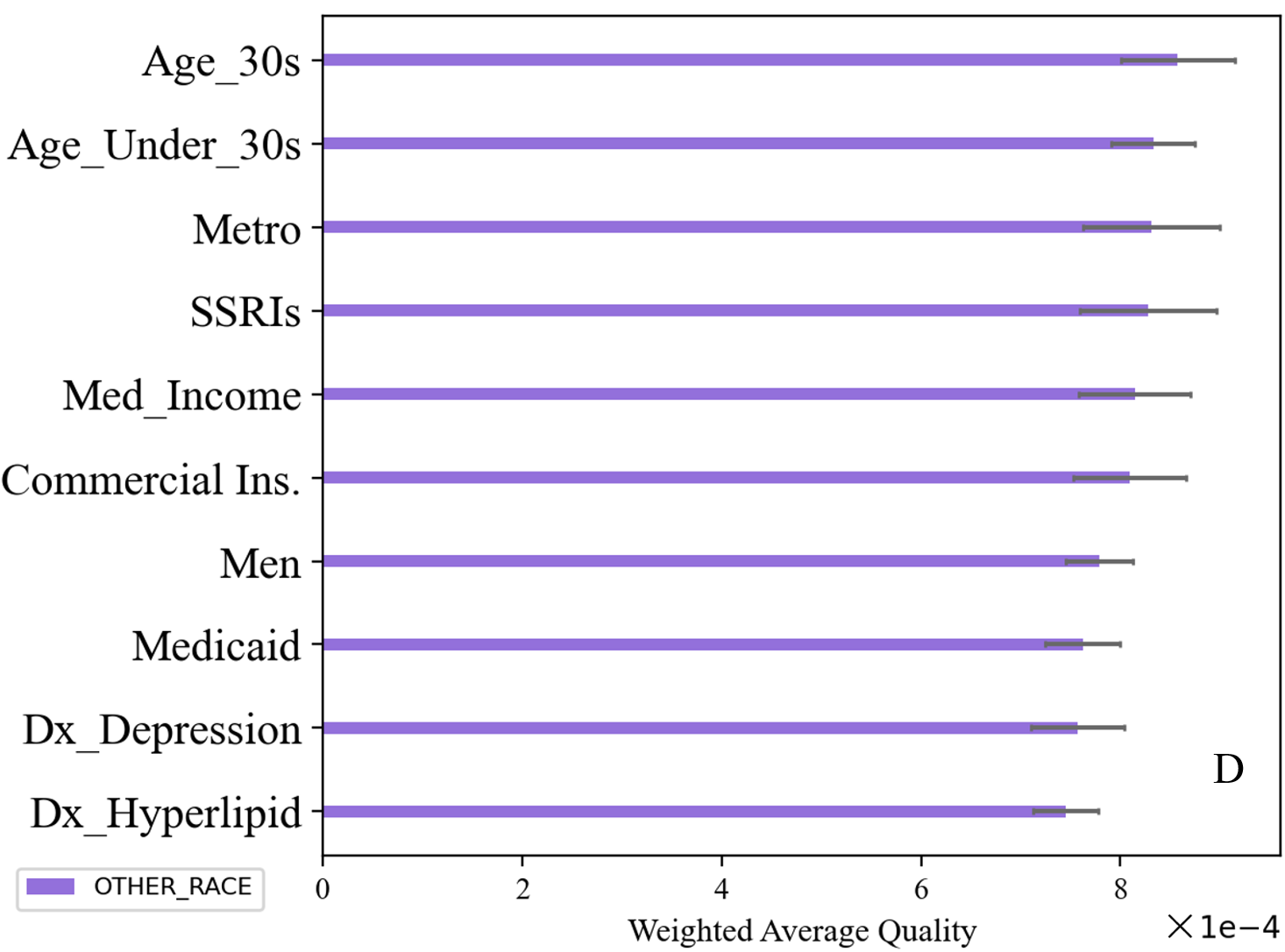}\\[-1ex]
  \caption{Race}
  \end{subfigure}\\[-1ex]%\par\medskip
 
  \begin{subfigure}{\linewidth}
      \centering
  \includegraphics[width=.25\linewidth]{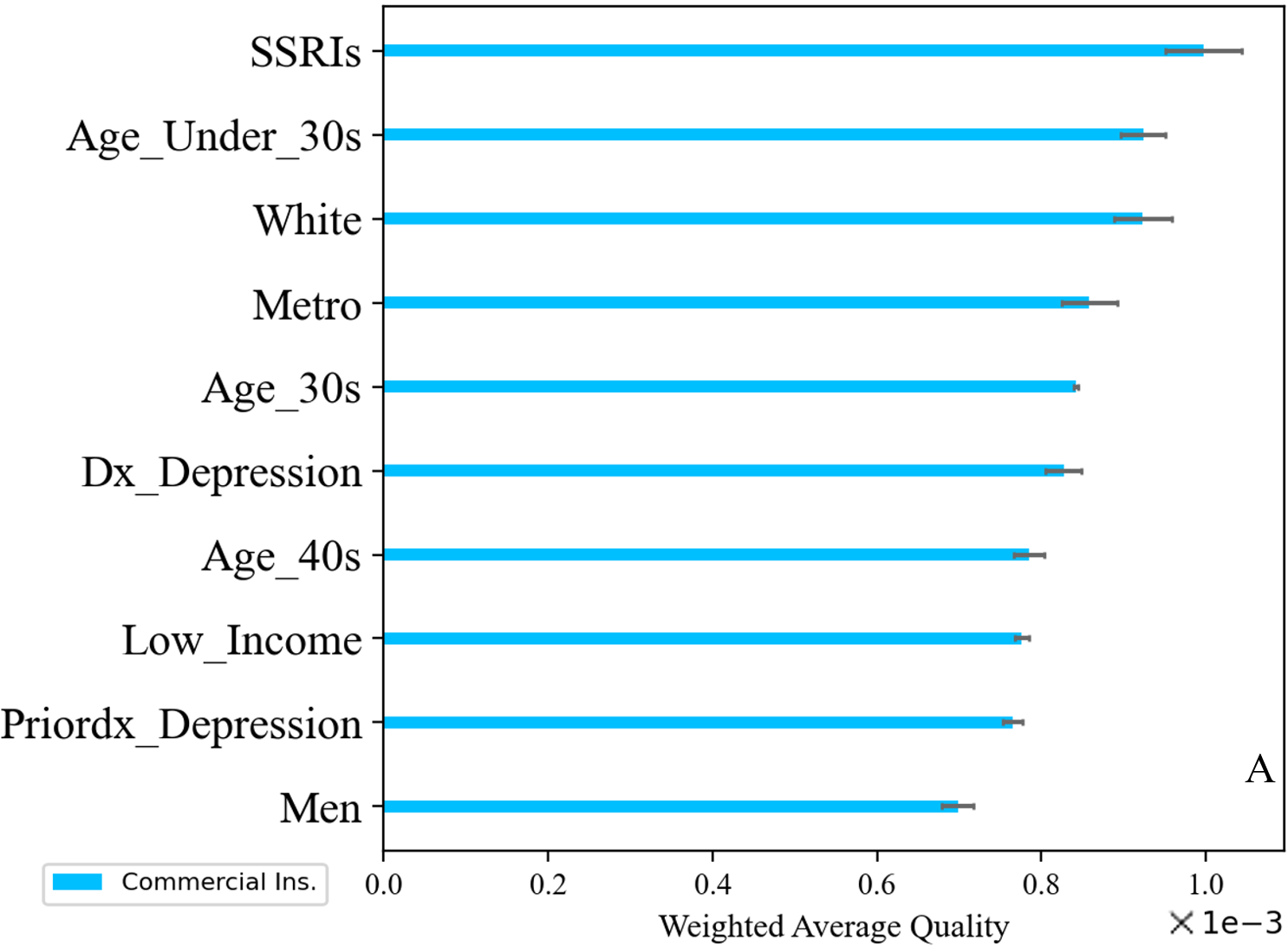}\hfill
  \includegraphics[width=.25\linewidth]{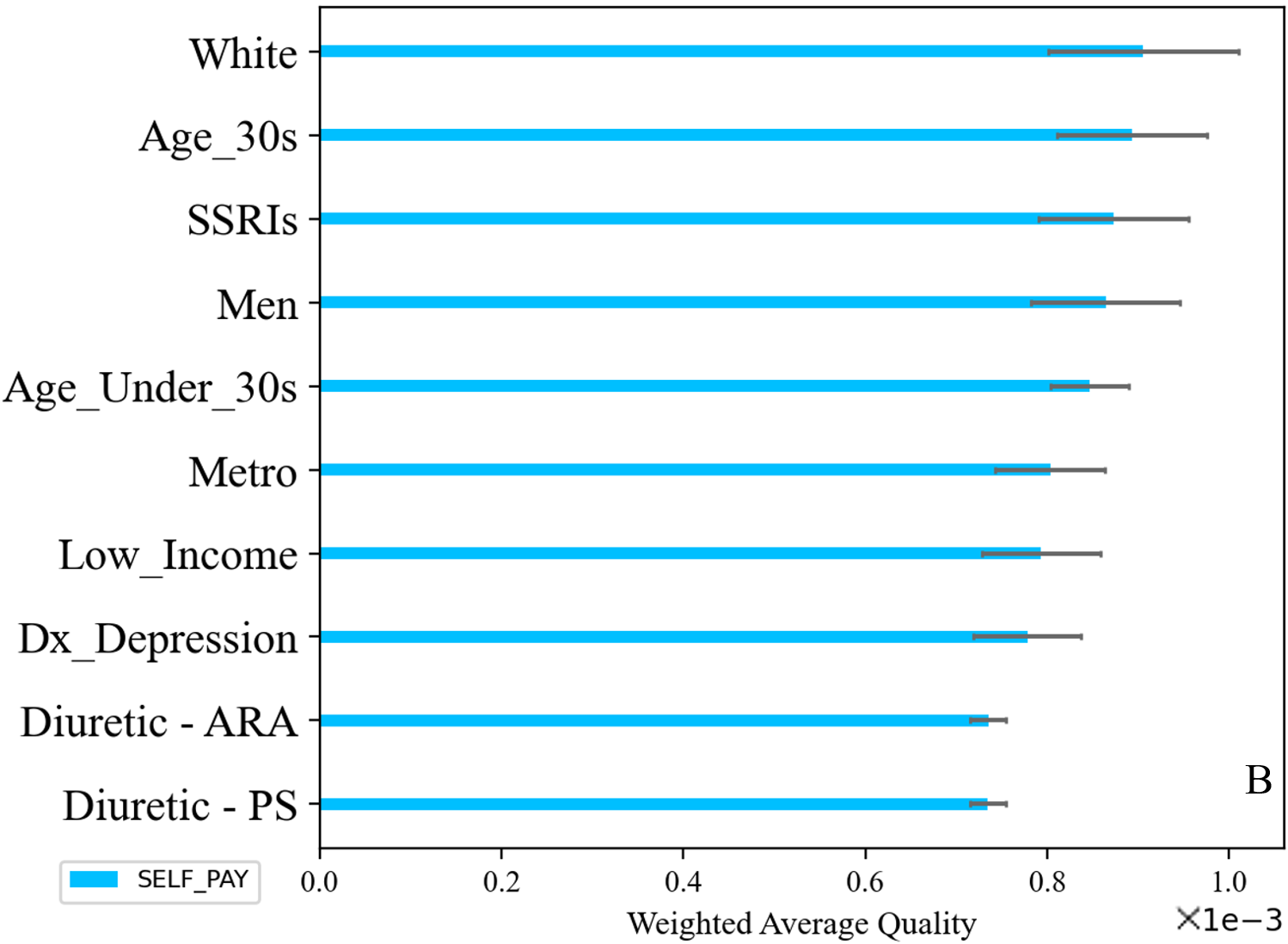}\hfill
  \includegraphics[width=.25\linewidth]{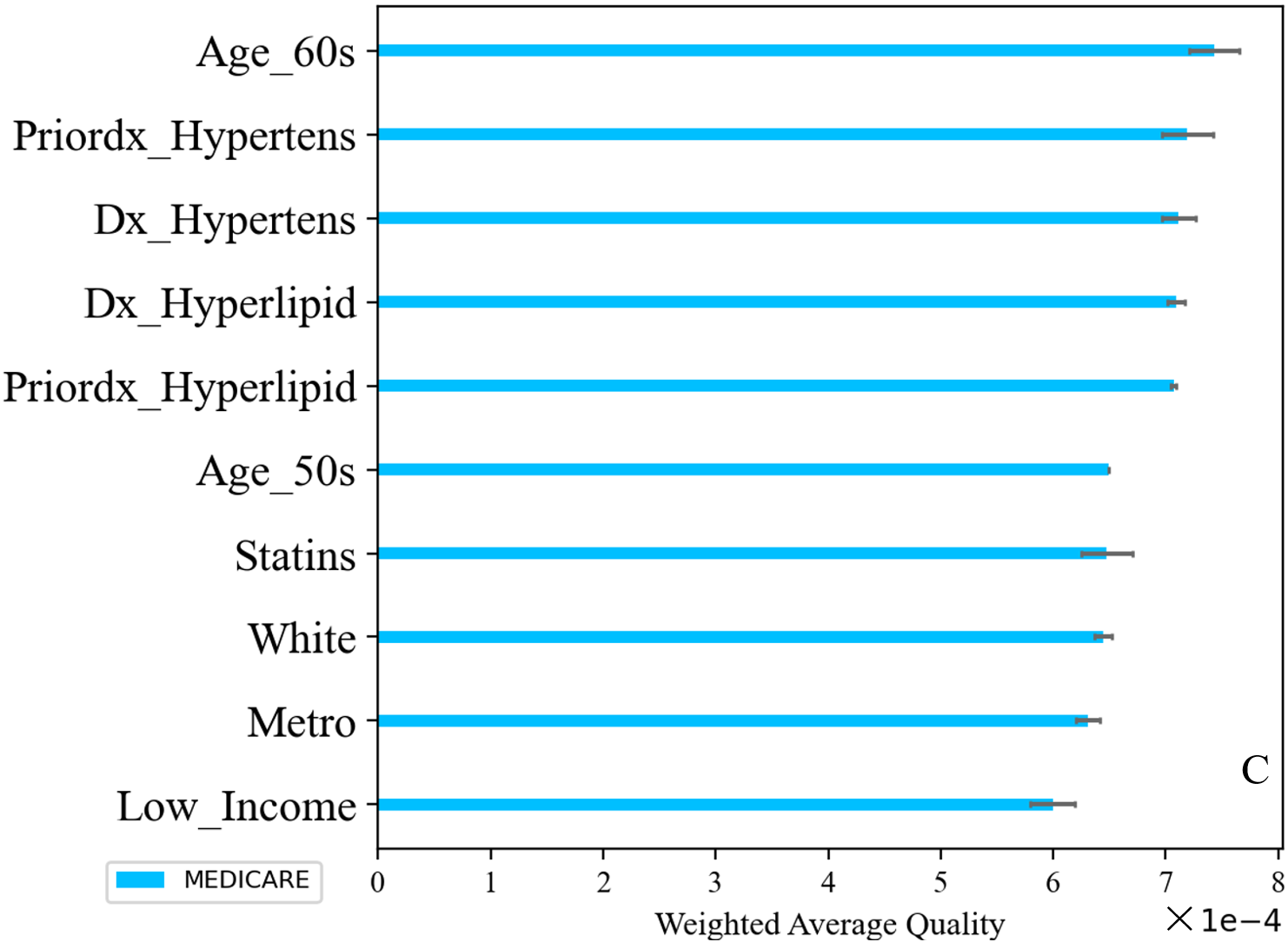}\hfill
  \includegraphics[width=.25\linewidth]{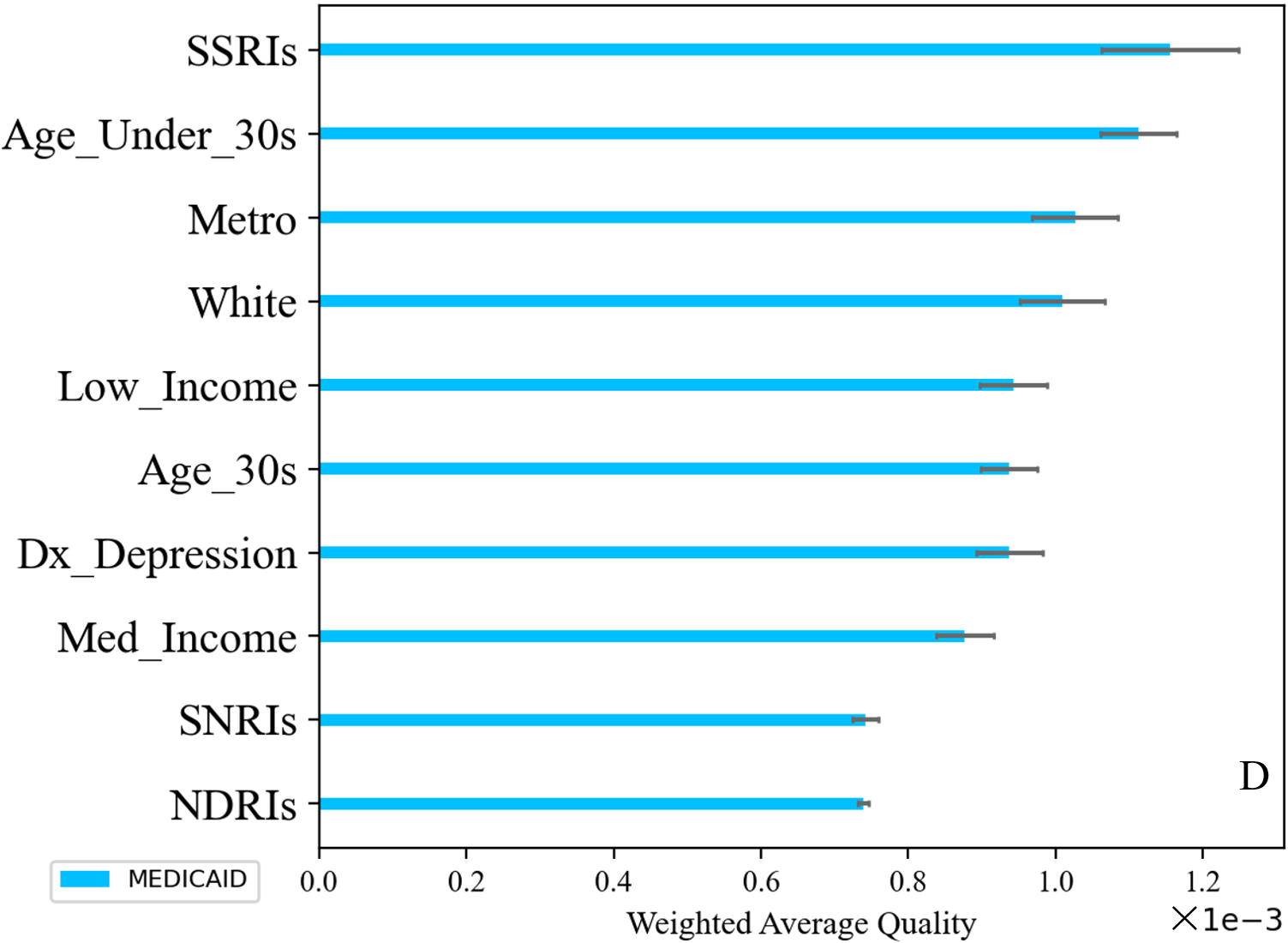}\\[-1ex]
  \caption{Insurance}
  \end{subfigure}\\[-1ex]%\par\medskip
  
  \begin{subfigure}{\linewidth}
    \centering
  \includegraphics[width=.25\linewidth]{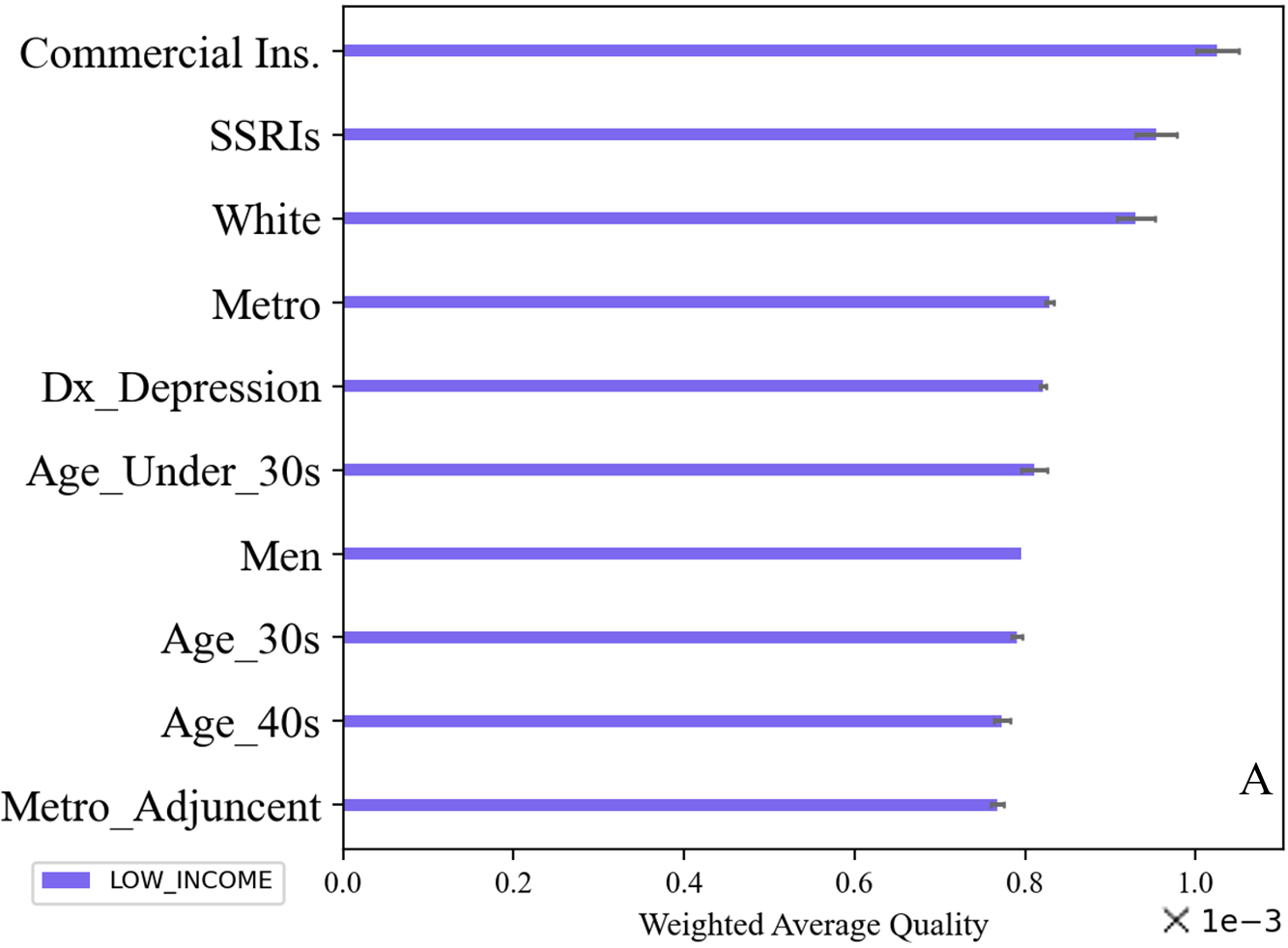}
  \includegraphics[width=.25\linewidth]{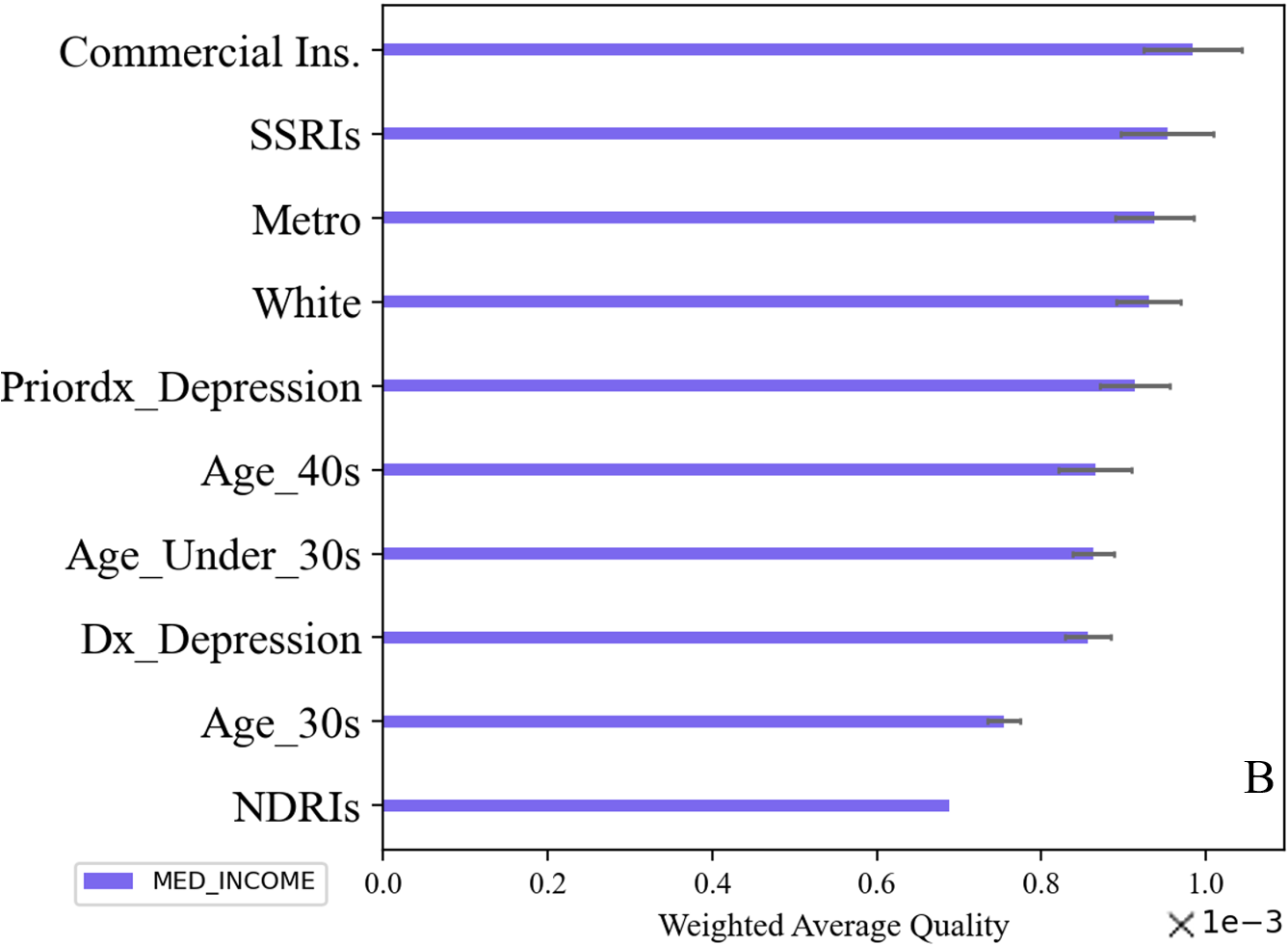}
  \includegraphics[width=.25\linewidth]{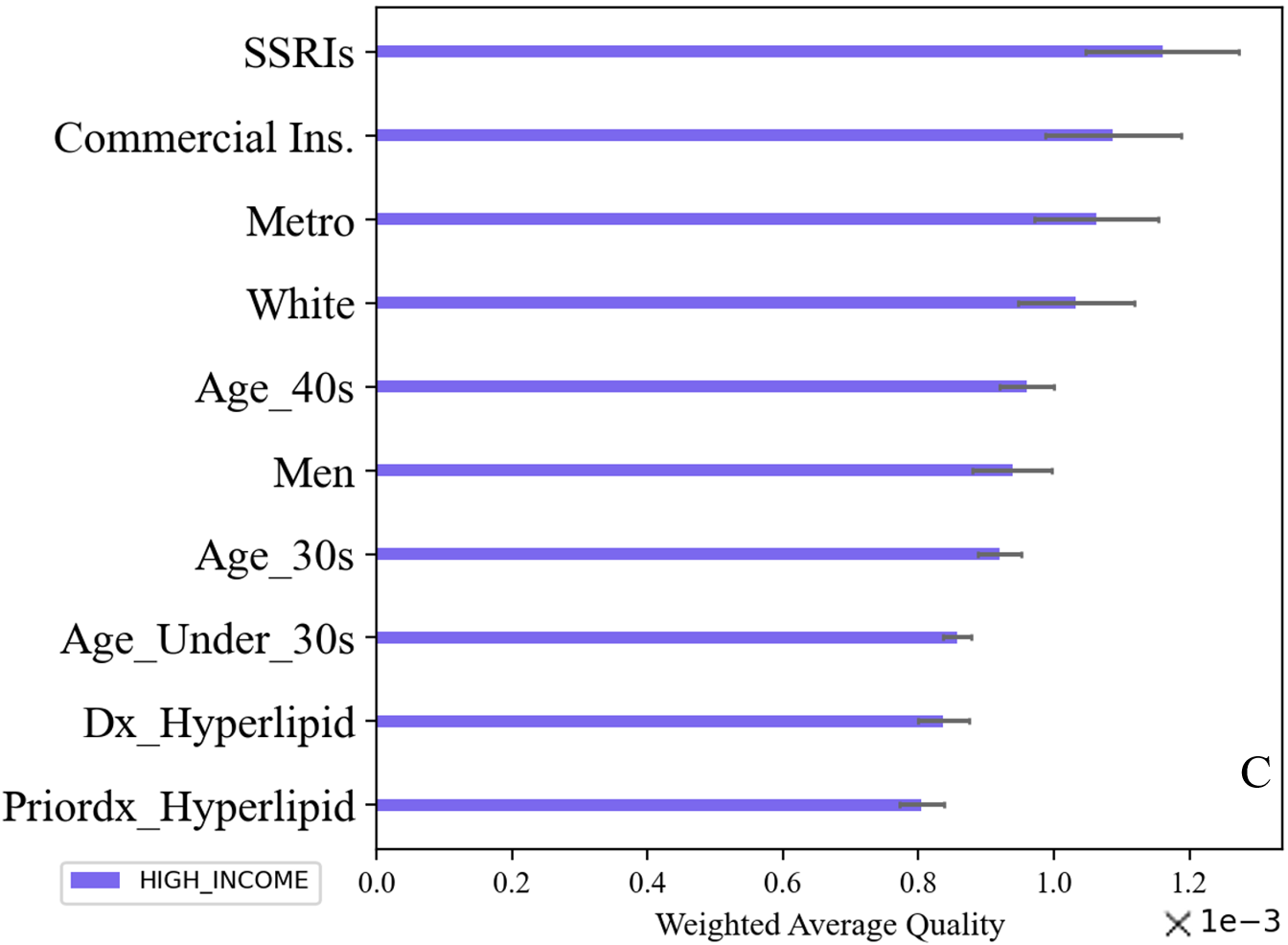}\\[-1ex]
  \caption{Income}
  \end{subfigure}\\[-1ex]%\par\medskip
  
  \begin{subfigure}{\linewidth}
      \centering
  \includegraphics[width=.25\linewidth]{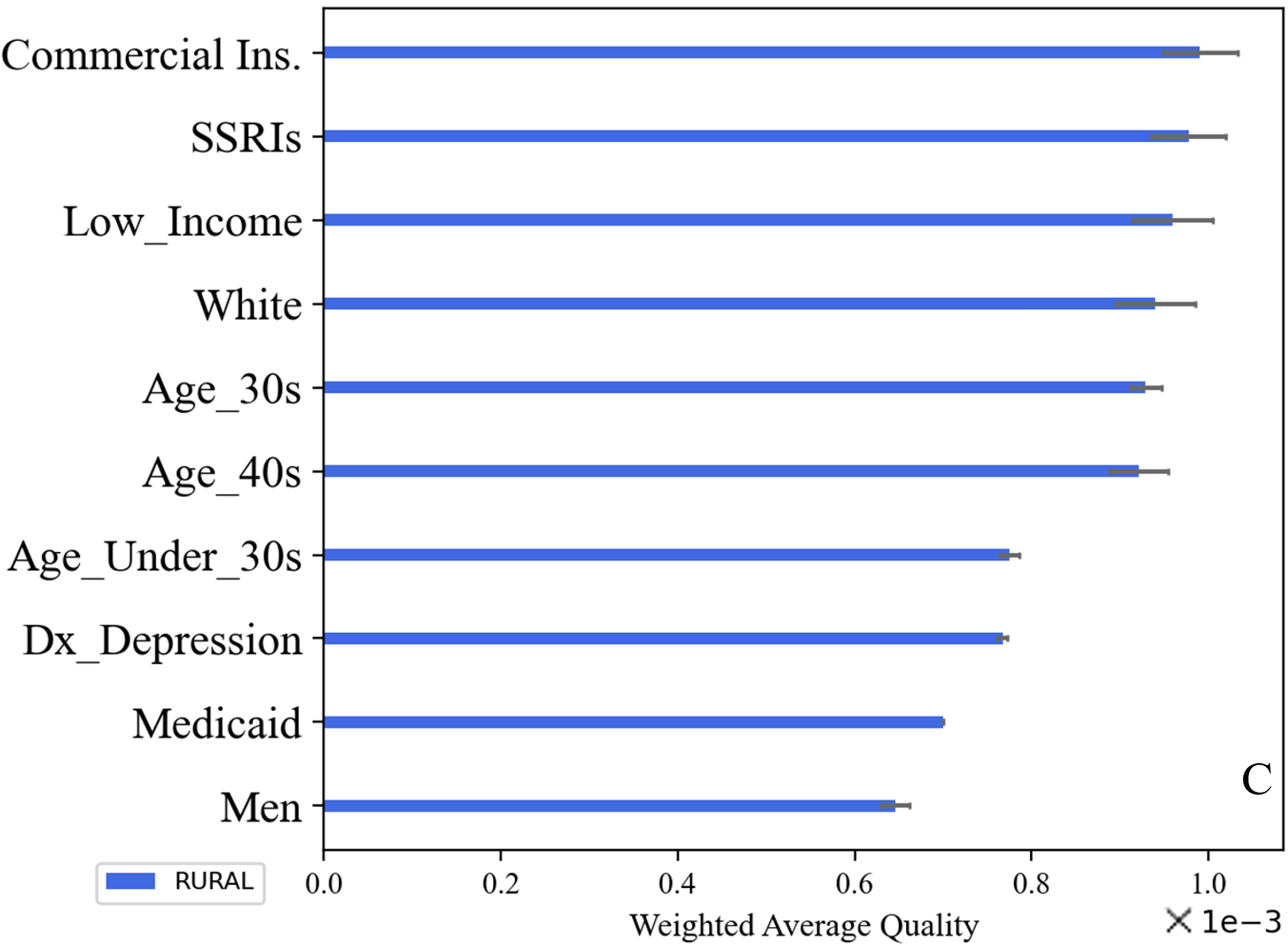}
  \includegraphics[width=.25\linewidth]{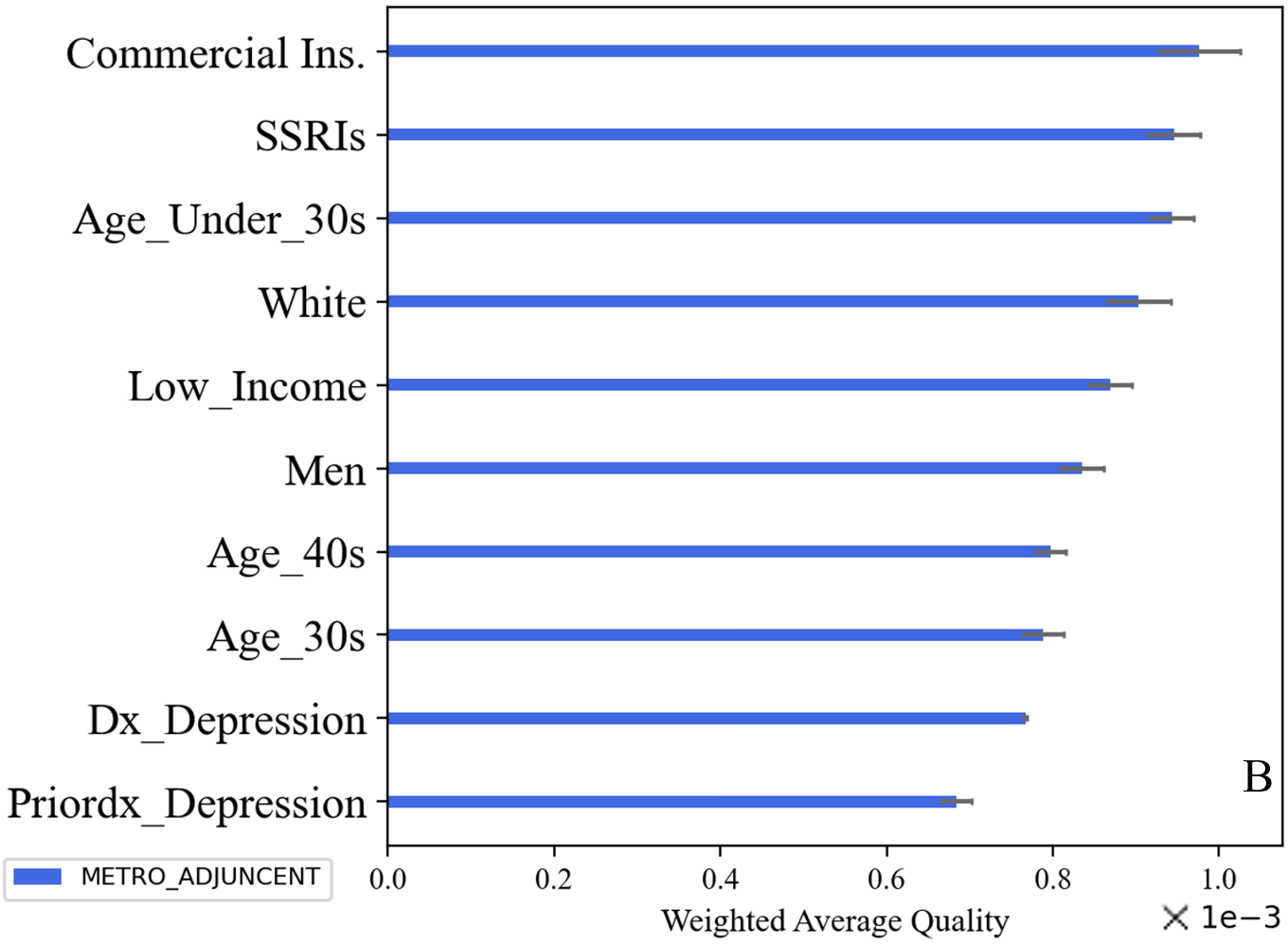}
  \includegraphics[width=.25\linewidth]{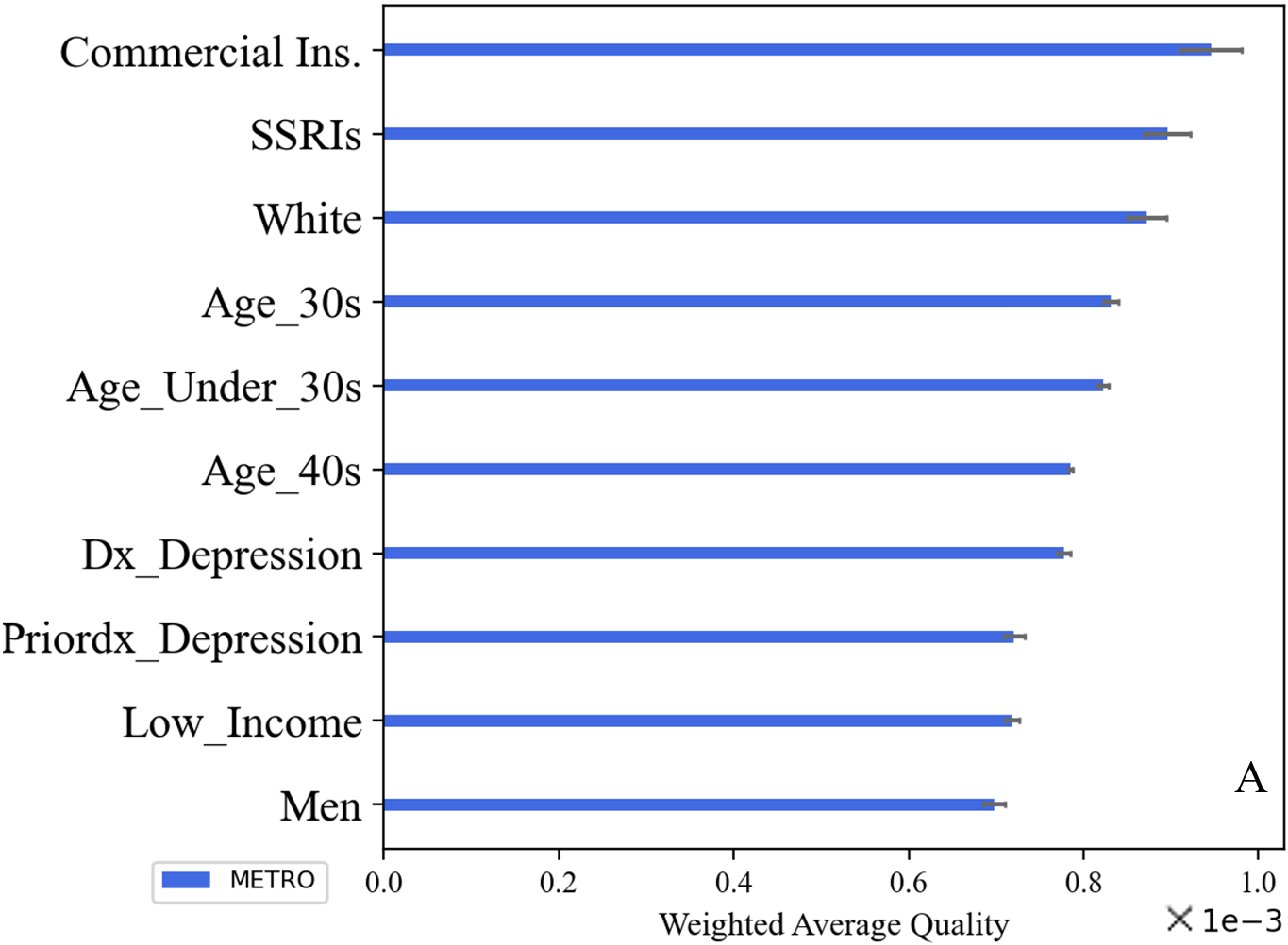}\\[-1ex]
  \caption{Neighborhood}
  \end{subfigure}\\[-1ex]
  
   \begin{subfigure}{\linewidth}
       \centering
  \includegraphics[width=.25\linewidth]{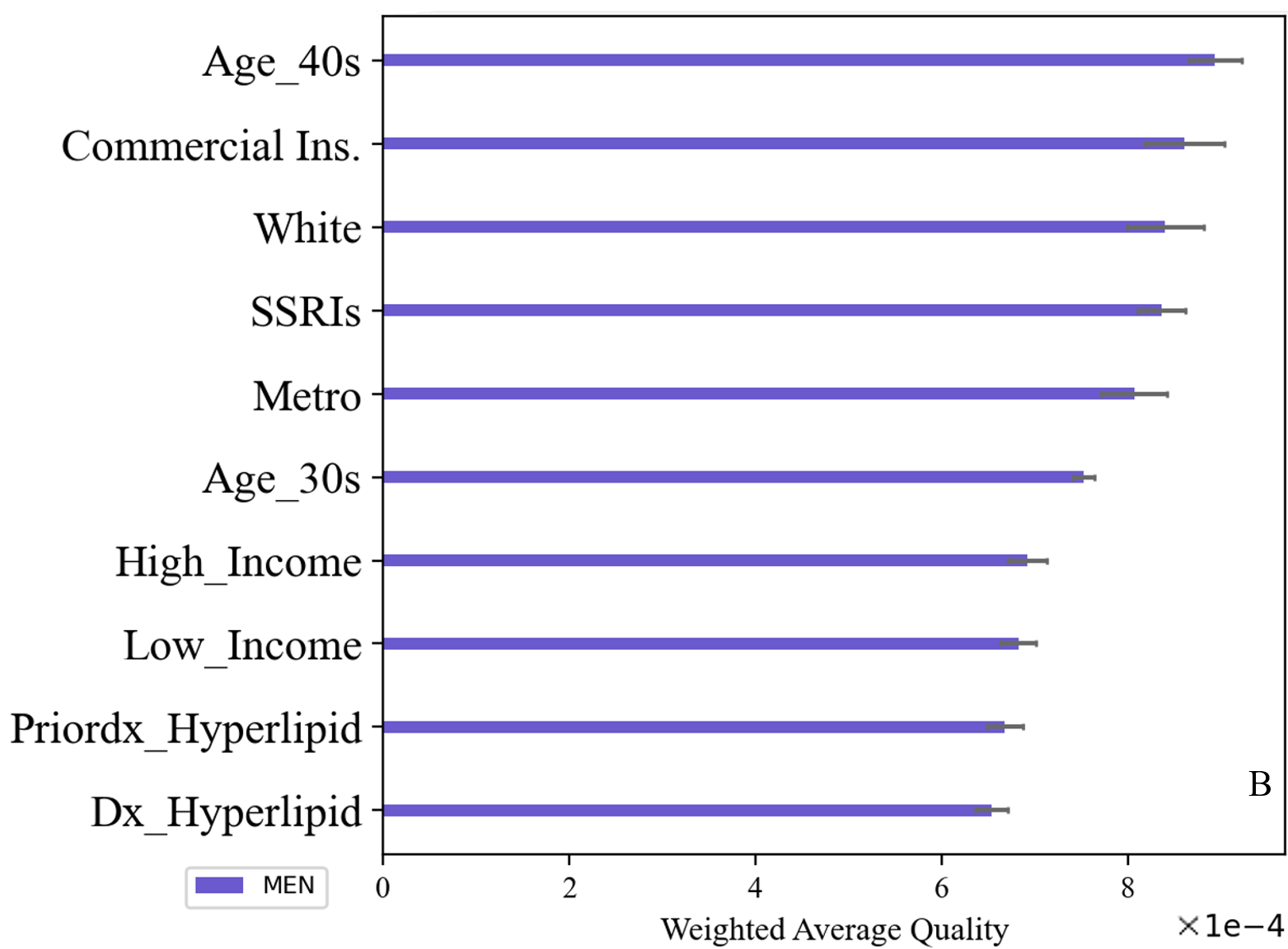}
  \includegraphics[width=.25\linewidth]{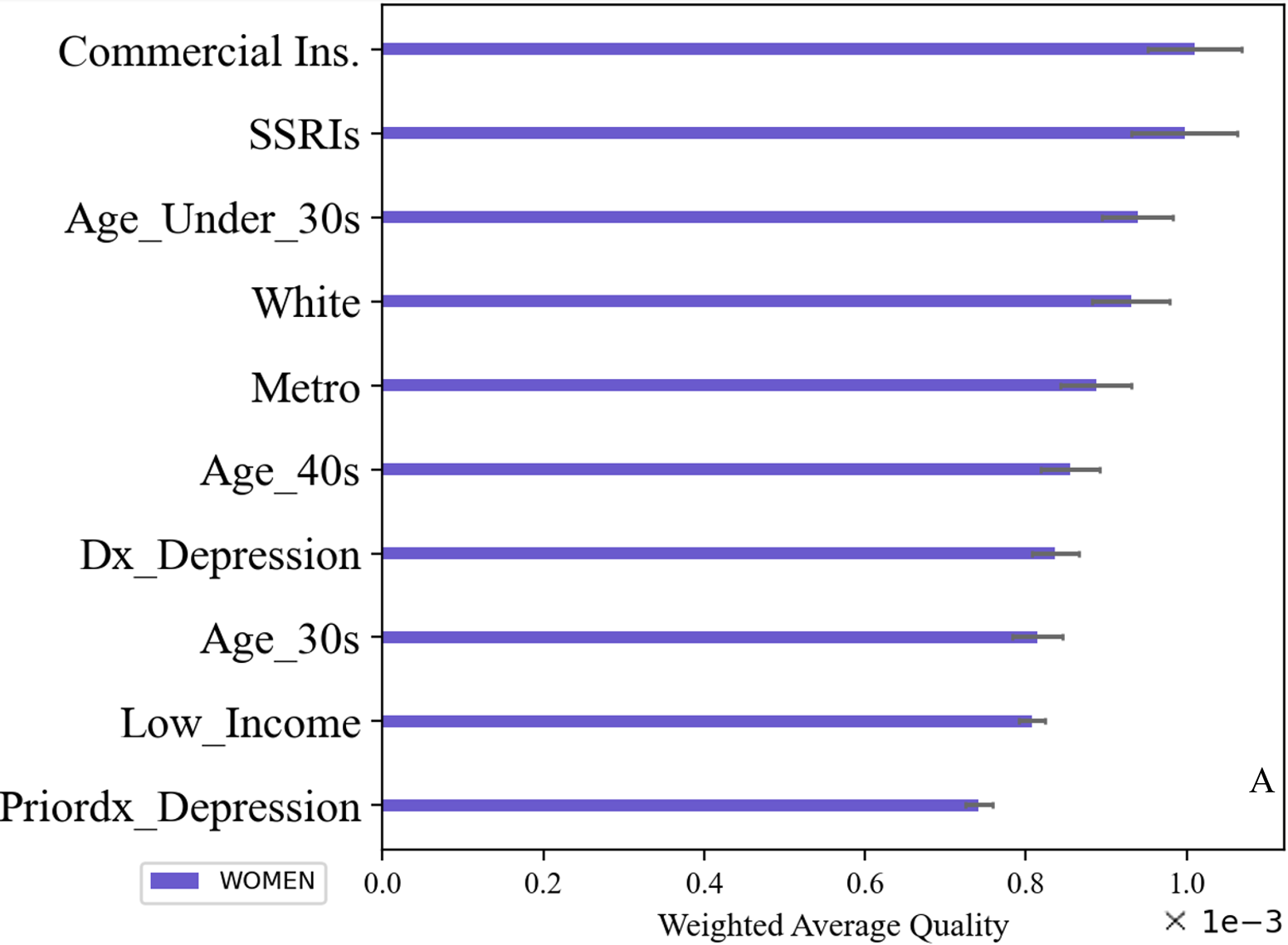}\\[-1ex]
  \caption{Gender}
  \end{subfigure}\\[-1ex]
  %\par\medskip
  \begin{subfigure}{\linewidth}
  \centering
  \includegraphics[width=.25\linewidth]{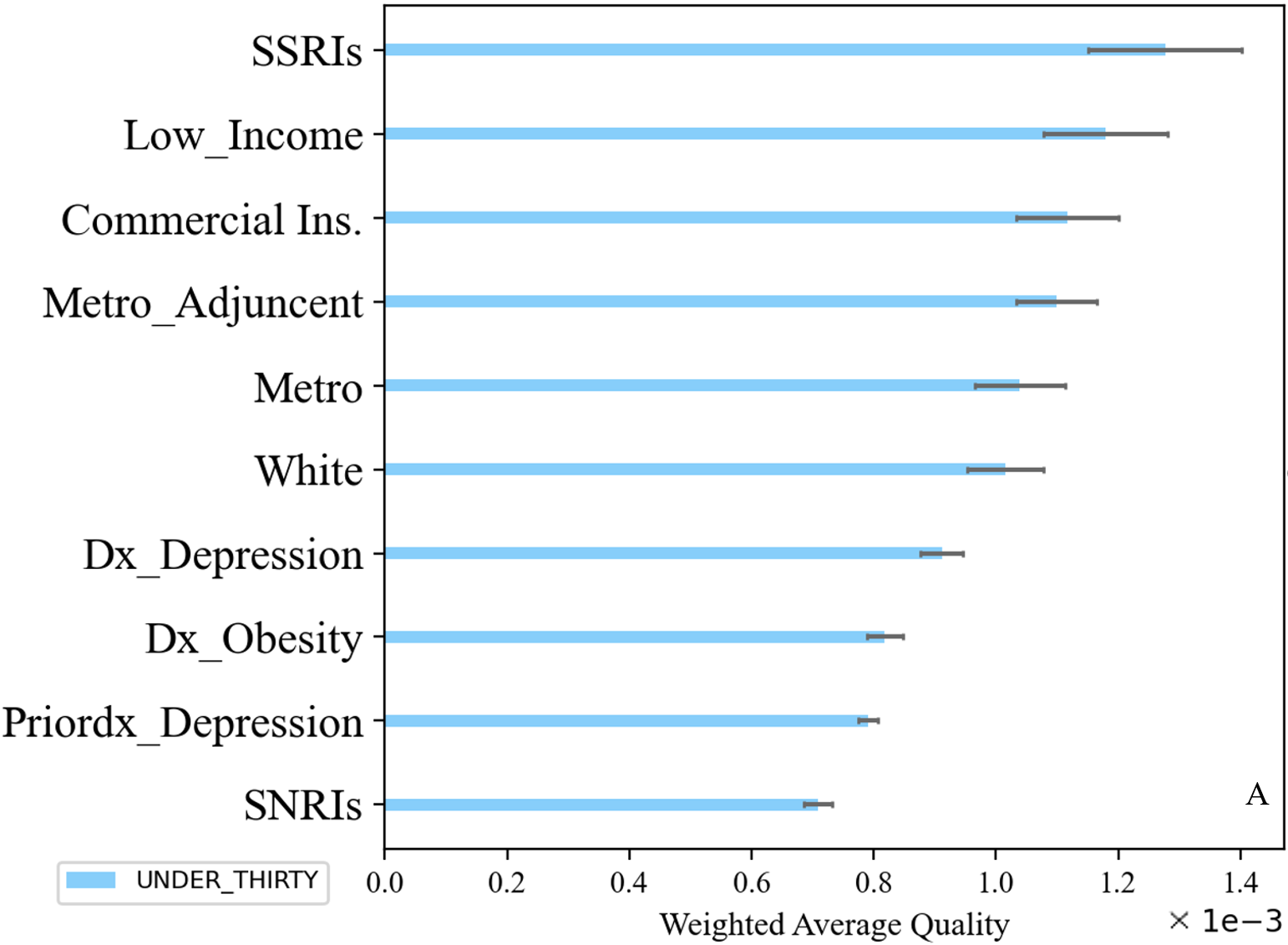}\hfill
  \includegraphics[width=.25\linewidth]{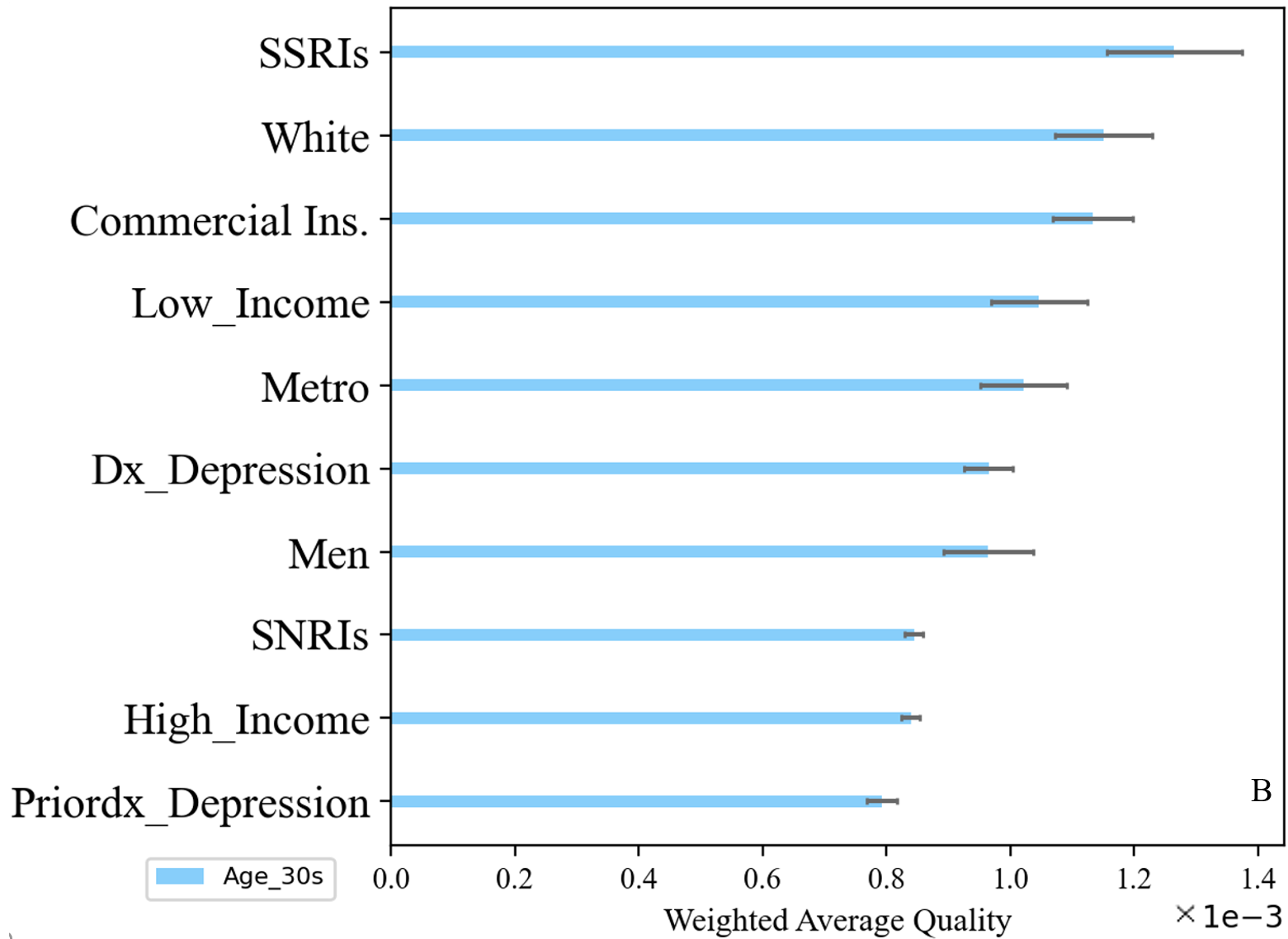}\hfill
  \includegraphics[width=.25\linewidth]{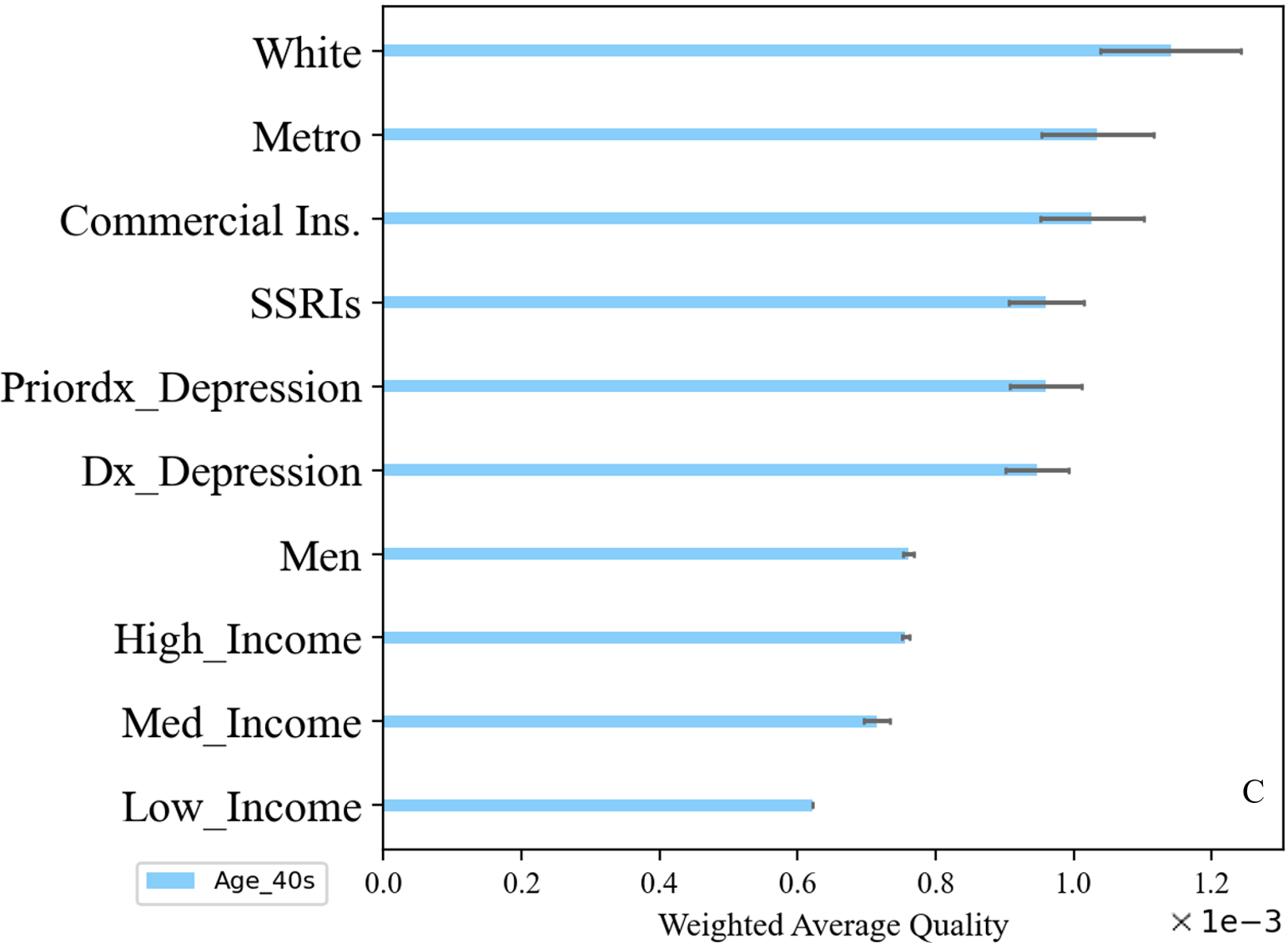}\hfill
  \includegraphics[width=.25\linewidth]{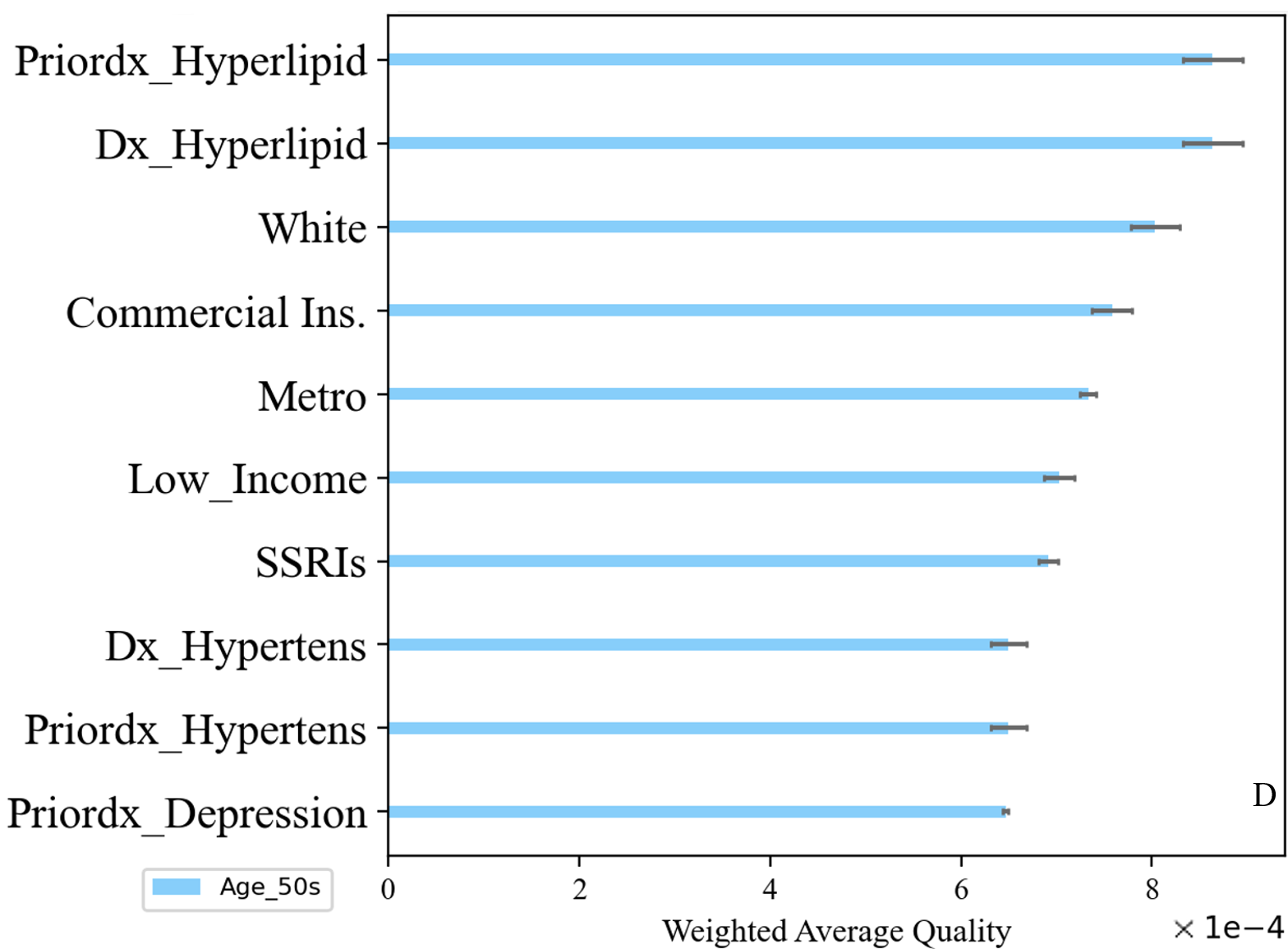}\hfill
  \\
  \includegraphics[width=.25\linewidth]{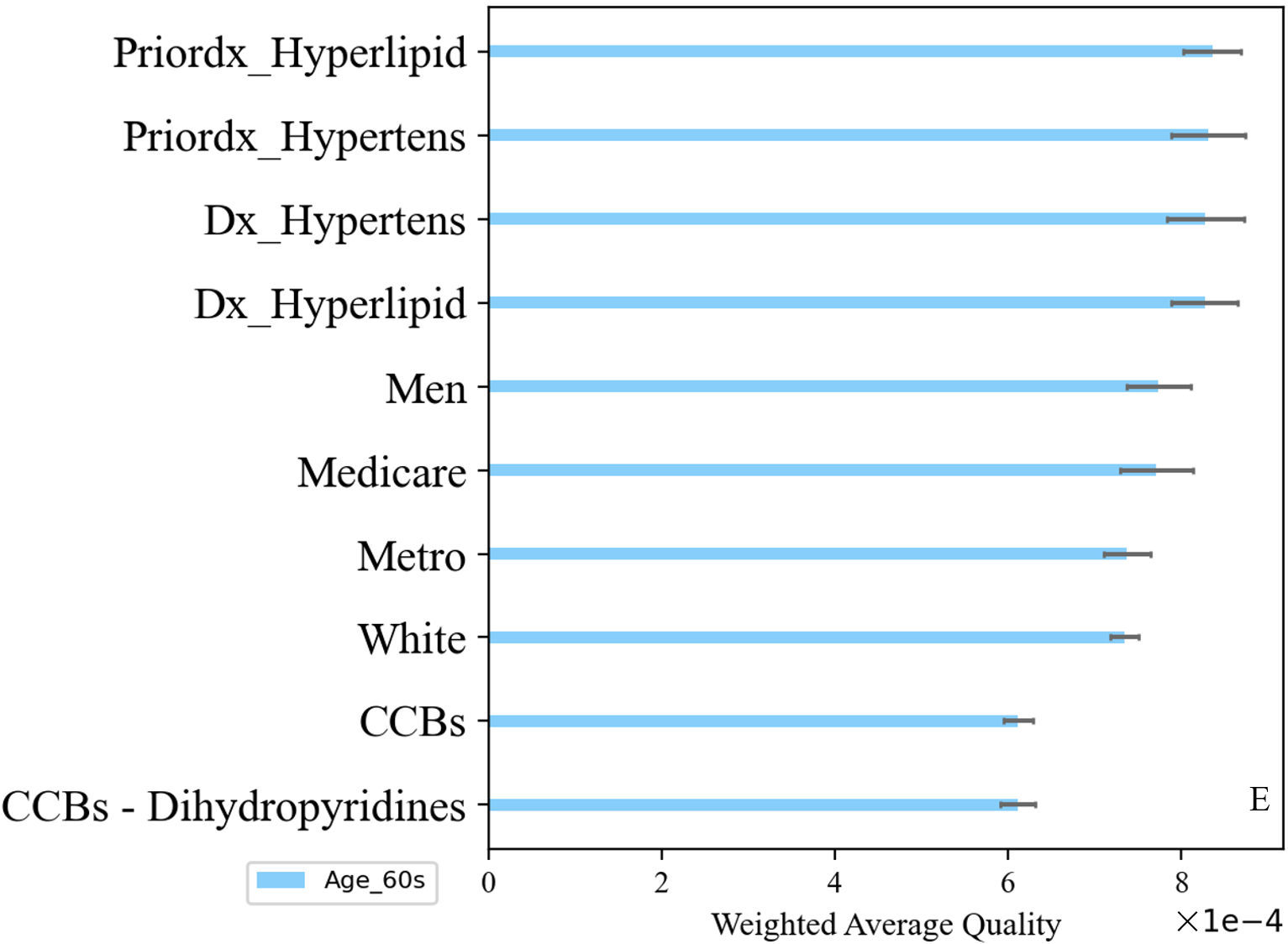}
  \includegraphics[width=.25\linewidth]{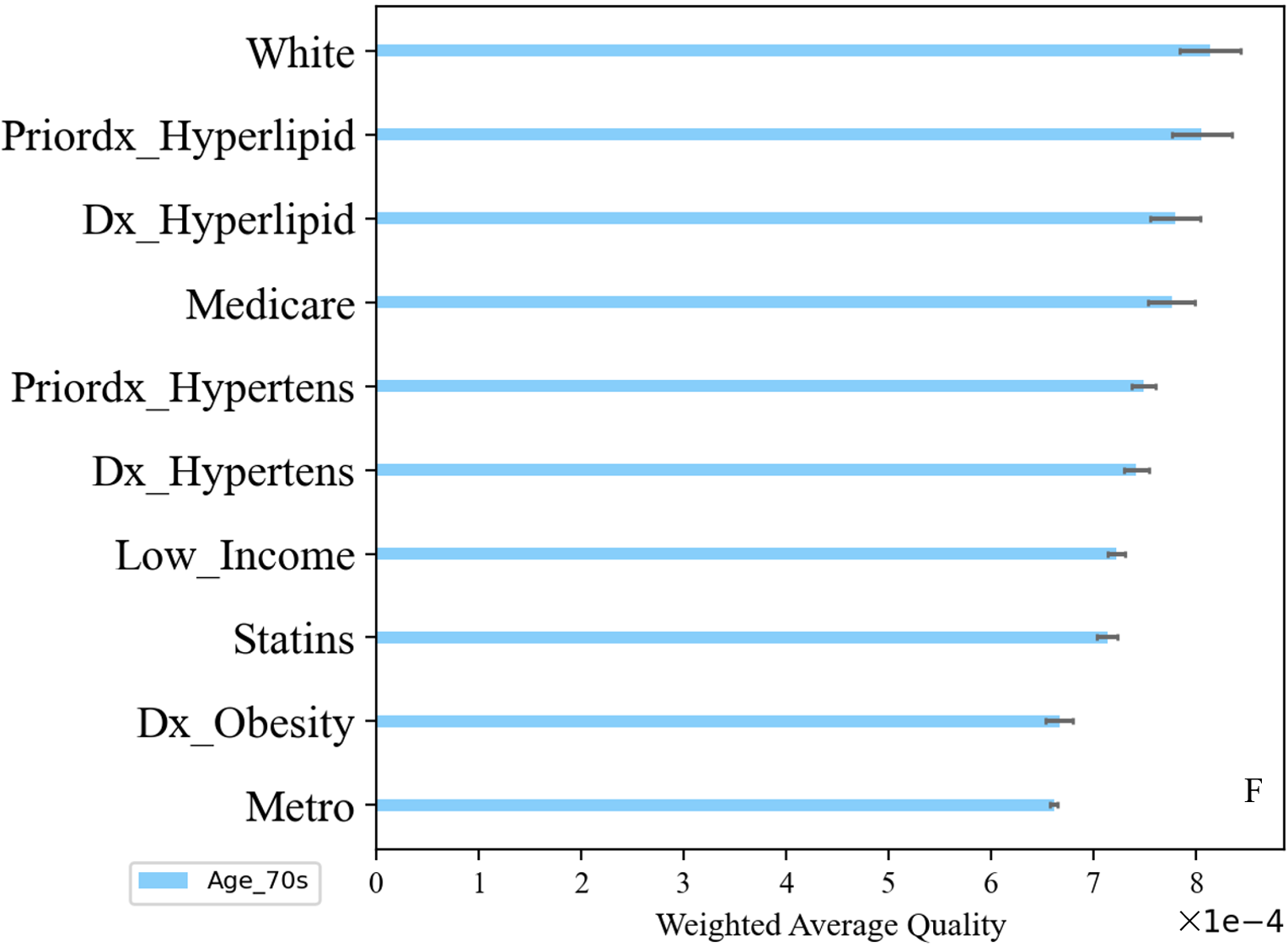}\\[-1ex]
  \caption{Age}
  \end{subfigure}\\[-1ex]
  \caption{Top 10 features with highest weighted average quality across 22 different strata indicated by: 1) race, 2) insurance, 3) income, 4) neighborhood, 5) gender, and 6) age. }
  \label{Fig:4}
\end{figure}

In our experiments, we first study the differences in the identified dangerous weight gain patterns across the entire study population, and then compare those versus the patterns obtained from any of the six strata indicated by the six demographic and socioeconomic characteristics. We refer to the 22 subgroups identified by these six characteristics as “strata” (instead of subgroups), to avoid confusion with the term subgroups used in the SD method). The notion of being a “top feature” in our study combines accuracy (features that most accurately predict weight gain) and support (features that apply to a large portion of samples). Choosing only accuracy may lead to identifying patterns that apply to few numbers of individuals and choosing only support may lead to identifying patterns that are not accurate enough. We run our experiments using three maximum lengths for the rules (referring to the {\it X} side of {\it X → Y} rules, as \textit{Y} is fixed in all cases) with values of three, four, and five; and three beam search widths (the number of generated rules in iterations) of 2,000, 5,000, and 10,000. Each bar in the results presented in this section (Fig \ref{Fig:2} to Fig \ref{Fig:4}) corresponds to the average value across the nine experimental settings that we have used in our main experiments (three different width of beam search by three different numbers of generated rules). Similarly, the error bars in the figures show the variation (95\% confidence interval) across the experimental settings. We use 5.0e-4 (or 0.0005), as the {\it WRAcc} threshold, which we empirically identified in our experiments. Besides the nine settings used here, we present more in-depth sensitivity analyses in Appendix \ref{secA1}. Also, in Appendix \ref{secA2}, we present the results related to the similar experiments to those reported here, but by further combining the 128 medication variables into five categories of antihypertension, antihyperlipidemic, antidepressant, antiobesity, and antidiabetic medications.
When studying the whole cohort (Fig. \ref{Fig:2}), having a commercial type of insurance is found to be the top factor predicting dangerous weight gains. This is in line with several studies reporting the overall importance of social determinant of health (SDOH) in determining the risks of developing obesity \citep{yusuf2020social,bryant2015social}. The second top feature is taking selective serotonin reuptake inhibitors (SSRIs), which are generally used as antidepression medications, and gaining weight is known to be among their major side effects \citep{ferguson2001ssri,cascade2009real,patten2009major,dannon2007naturalistic}. The third top feature is being white. While obesity is known to be more prevalent among non-Hispanic blacks \citep{fryar2020prevalence}, identifying the white race as a top feature in predicting dangerous weight gains may indicate complementary patterns. We note that our method (through {\it WRAcc}) already accounts for imbalanced nature of samples. For instance, while white race is the dominant race in our samples, appearing on top of the important features list is not solely due to its higher frequency. Living in metro areas is identified as the fourth feature, which is in line with the studies reporting more sedentary lifestyle and consuming less healthy foods in metropolitan areas \citep{bodor2010association,ewing2003relationship}. A similar pattern to the third feature (race=white) can be also observed for the fifth feature (gender=man), where obesity is known to be more prevalent among women than men in the US  \citep{fryar2020prevalence,fryar2018mean}. Being in the 40s, under 30s, and 30s are placed in sixth, seventh, and ninth features, aligned with the studies reporting that the middle aged groups (40s and 50s) suffer more from obesity \citep{fryar2020prevalence,fryar2018mean}. Depression is ranked eighth (the Dx prefix indicates the diagnosis of a disease). Like SSRIs, the positive association between obesity and depression is reported by many studies \citep{roberts2003prospective,luppino2010overweight,dong2004relationship,faith2002obesity,preiss2013systematic}. The tenth feature is having low income, which is another major SDOH, widely known as a major risk factor in developed countries (a phenomena sometime called ``reverse gradient'') \citep{bentley2018recent}.

\begin{figure}
  \begin{subfigure}{\linewidth}
  \includegraphics[width=.5\linewidth]{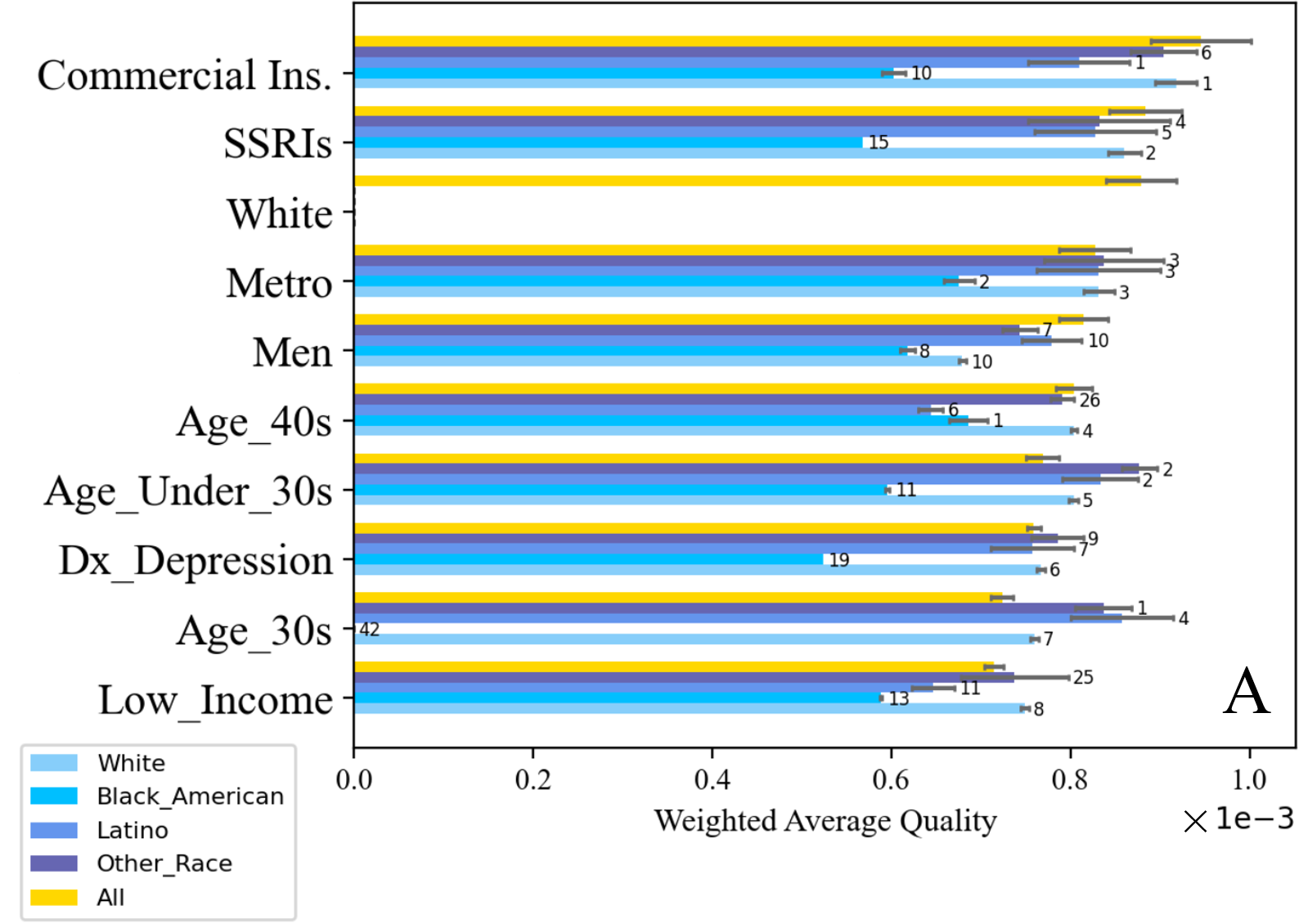}\hfill
  \includegraphics[width=.5\linewidth]{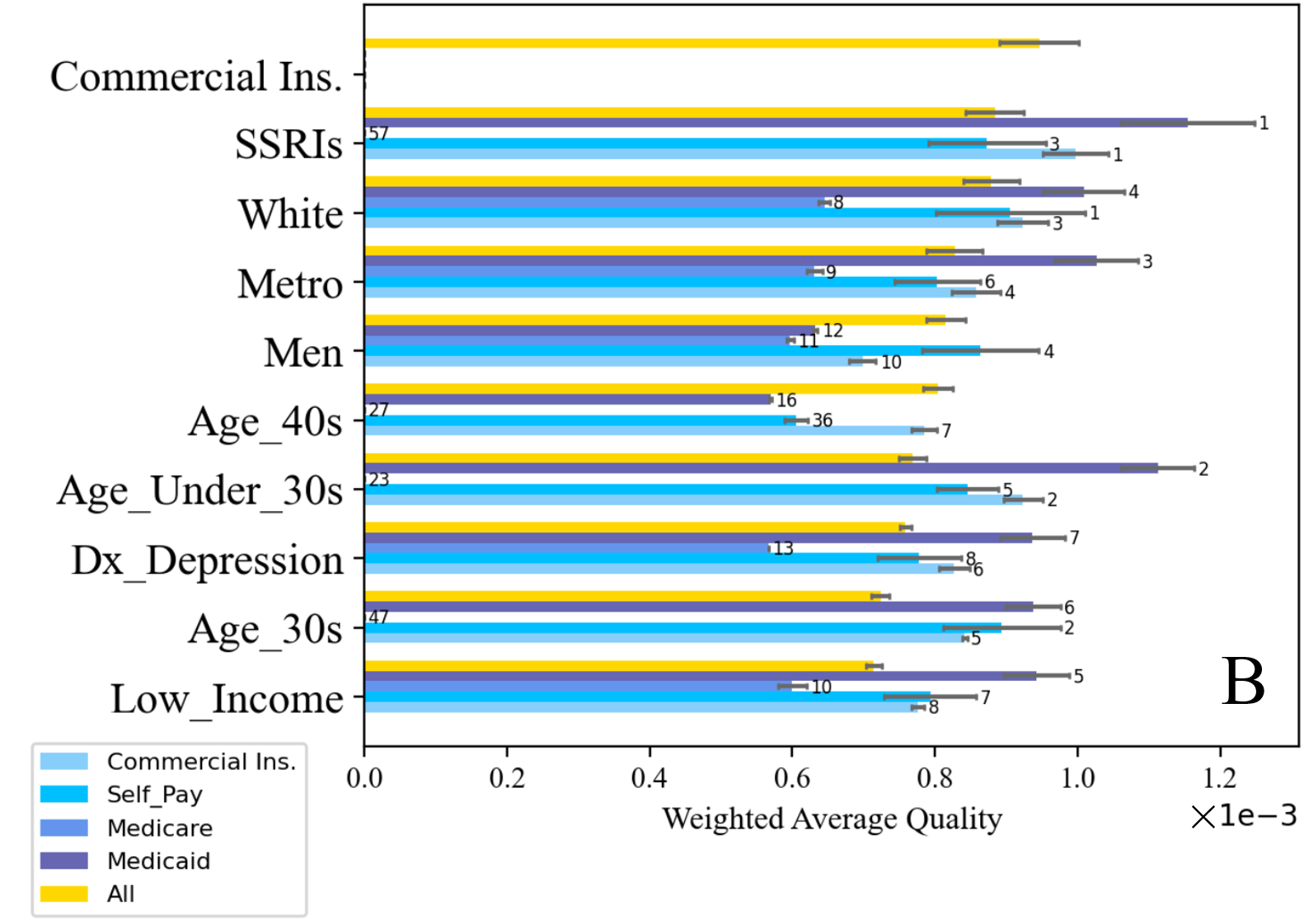}
  
  \end{subfigure}\par\medskip
  \begin{subfigure}{\linewidth}
  \includegraphics[width=.5\linewidth]{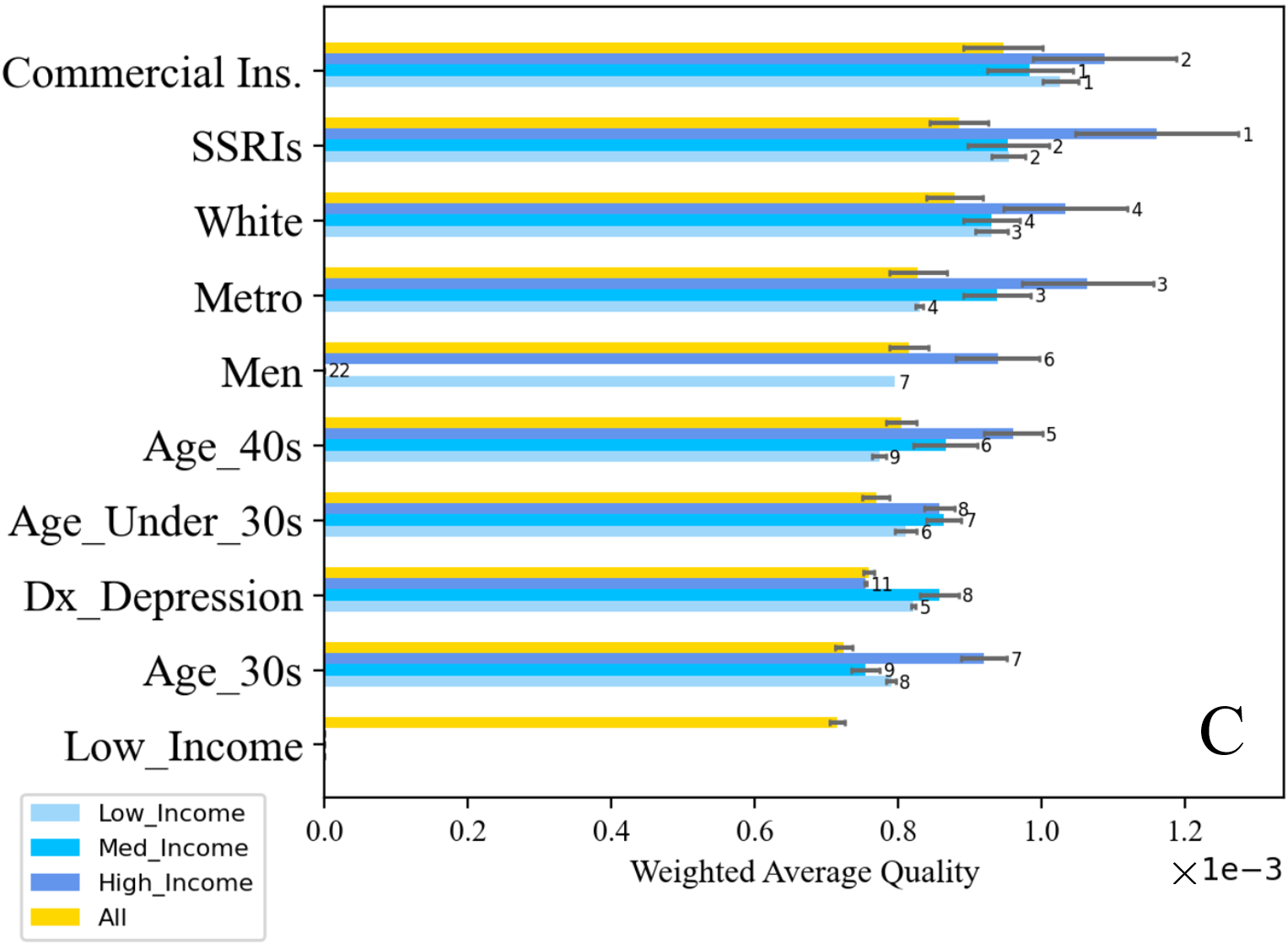}\hfill
  \includegraphics[width=.5\linewidth]{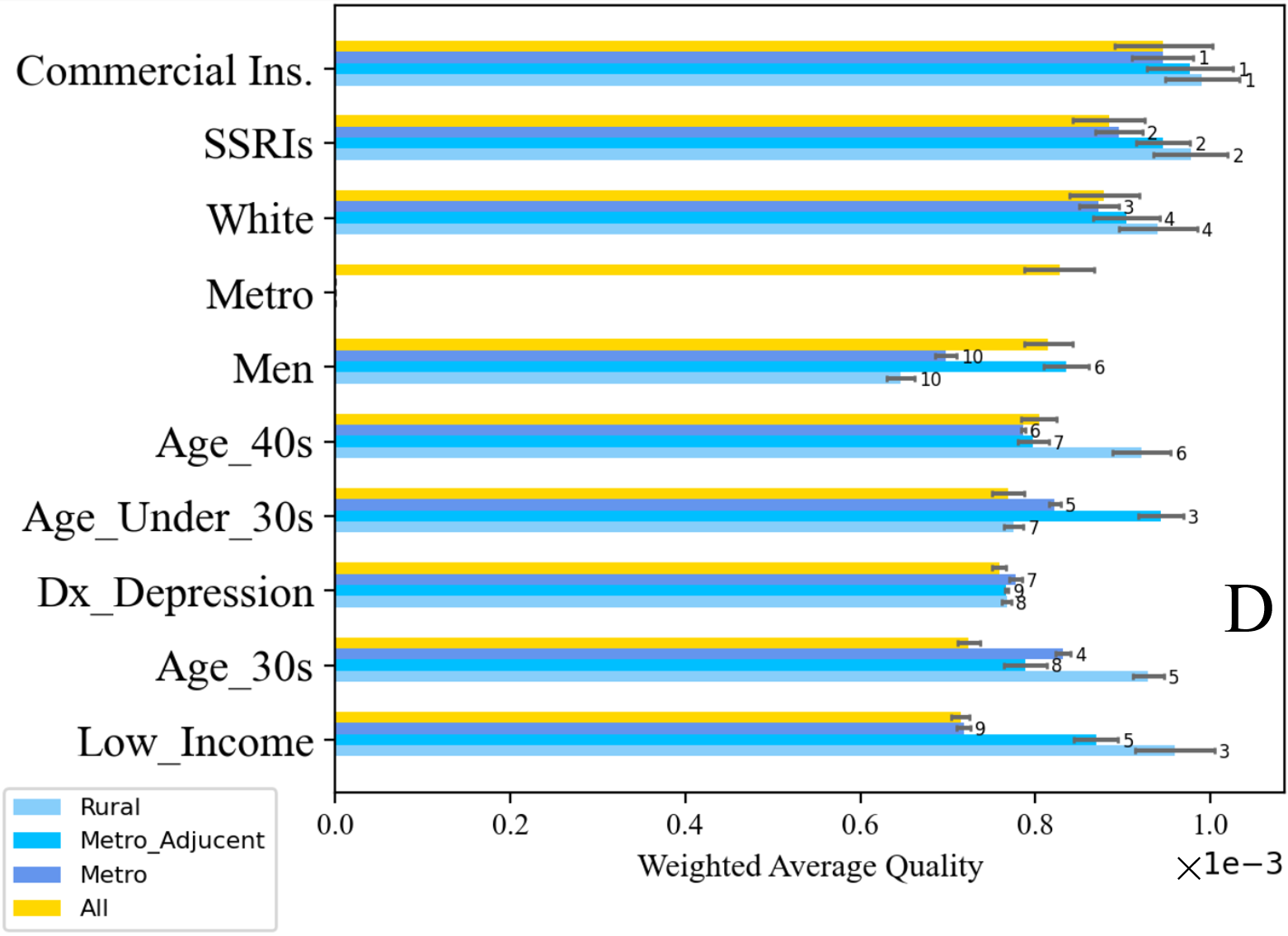}
  
  \end{subfigure}\par\medskip
  \begin{subfigure}{\linewidth}
  \includegraphics[width=.5\linewidth]{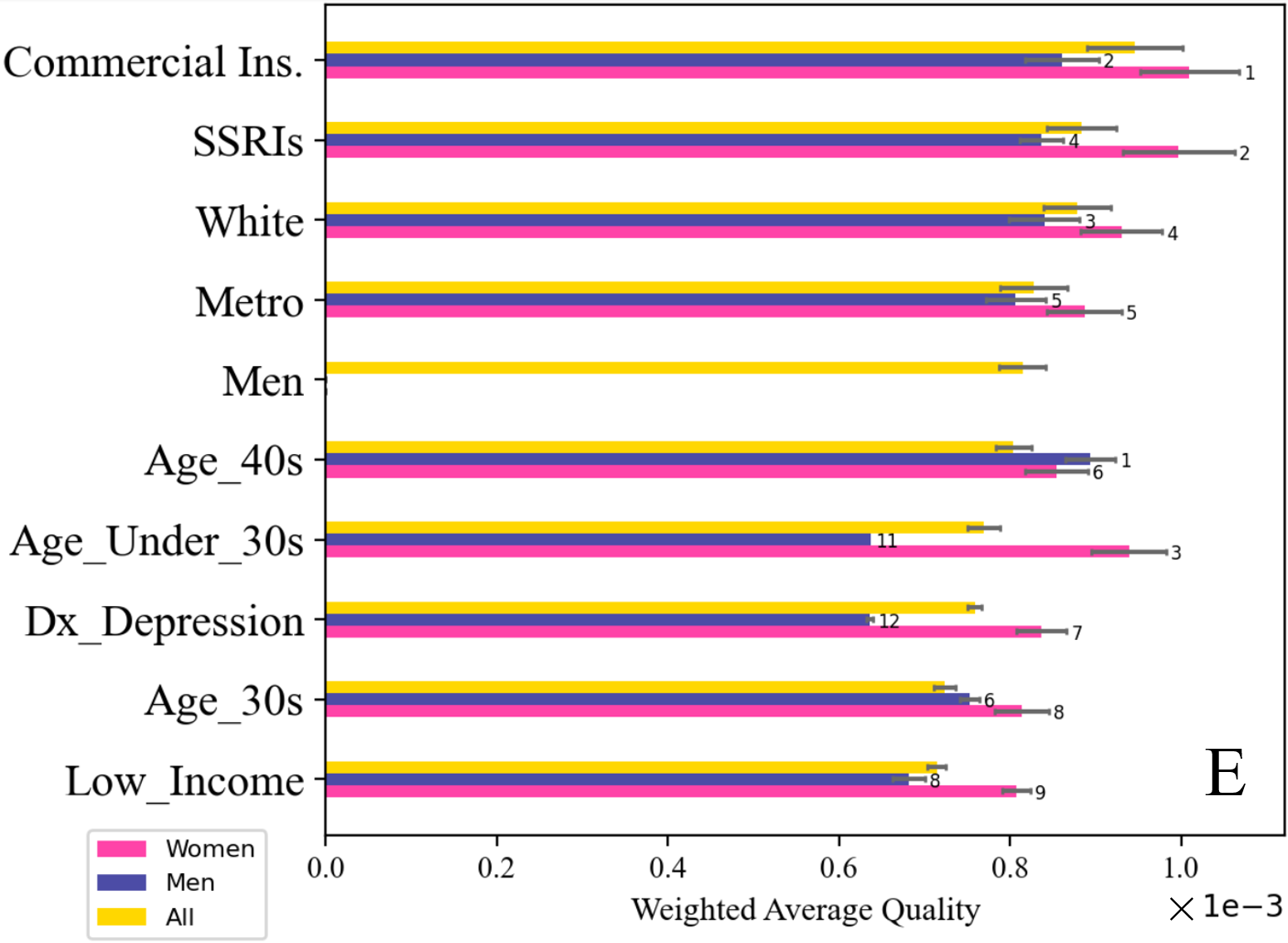}\hfill
  \includegraphics[width=.5\linewidth]{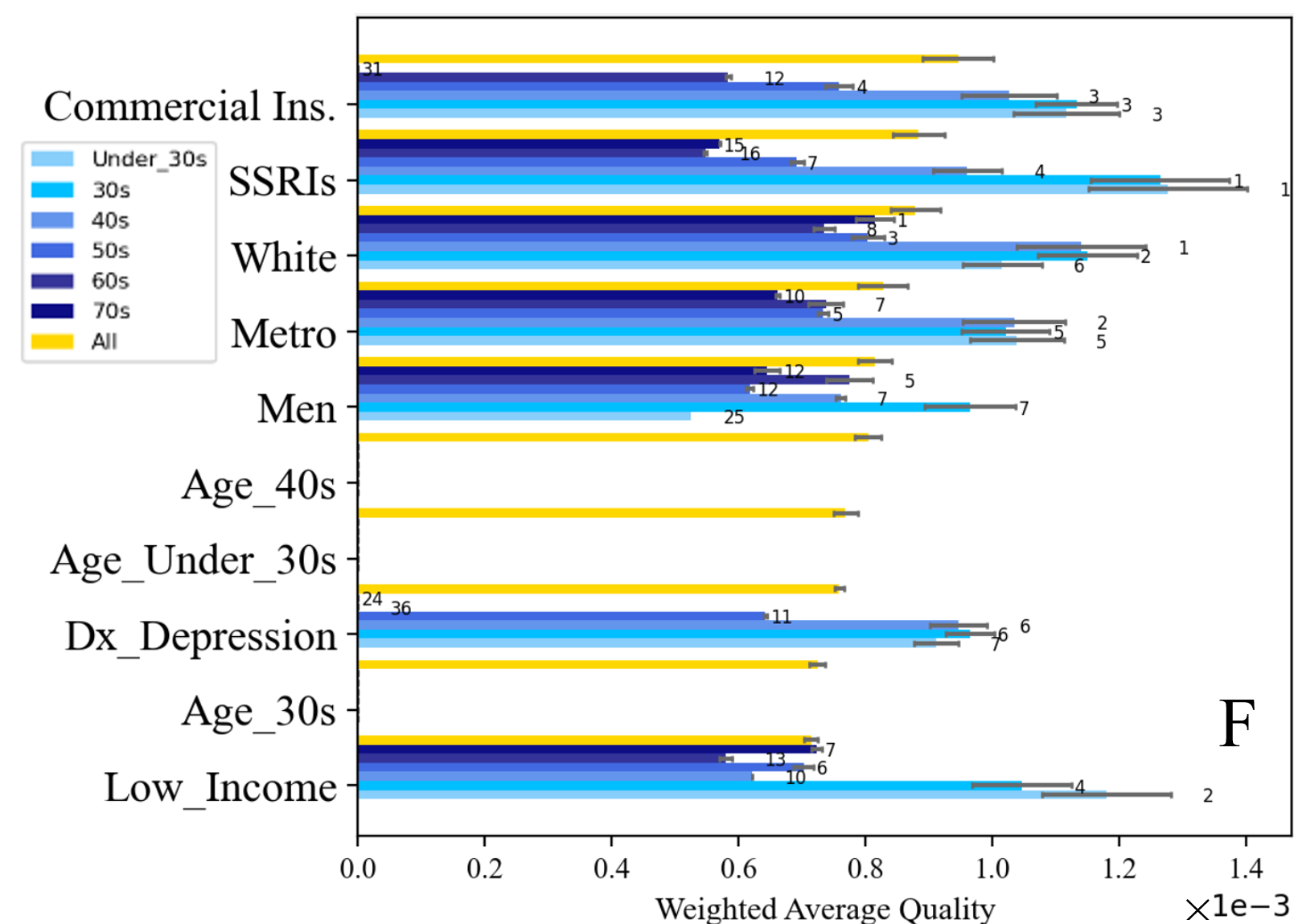}
  
  \end{subfigure}
  \caption{Comparing where the top 10 features shown in Fig. \ref{Fig:2}  (shown by the yellow bars here) stand  in 22 strata determined by six demographics and socioeconomic factors: A) race, B) insurance type, C) income level, D) neighborhood type, E) gender, and F) age. }
  \label{Fig:3}
\end{figure}

Following the top features across the whole cohort in Fig. \ref{Fig:2}, the six charts in Fig. \ref{Fig:3} show where these ``same top 10 features'' stand across the 22 separate strata. While most patterns observed in Fig. \ref{Fig:3} remain unchanged compared to the whole cohort as shown in Fig. \ref{Fig:2}, many new patterns can be also observed. In following, we cover some of the more prominent patterns. In Fig. \ref{Fig:3}.A, ranking of the top features in Fig. \ref{Fig:2} are shown accross different races. Here, it can be seen that taking SSRIs and depression have higher ranks in Black-Americans. In Fig \ref{Fig:3}.B (stratified by insurance type), one can observe that SSRIs, Under-30s, 30s, and 40s have values equal to zero in the Medicare strata. This is reasonable, as Medicare is only offered to 65+ yr individuals. In Fig \ref{Fig:3}.C (stratified by income levels), in the middle-income category the Men feature has a zero value, which can be compared to the studies showing that low- and high-income men have higher risks of obesity \citep{bentley2018recent,ogden2017prevalence}. Fig \ref{Fig:3}.E (stratified by gender) shows that in men, depression is not among the top 10 features and SSRIs are ranked fourth. Fig \ref{Fig:3}.F (stratified by age), shows that a commercial insurance (which was found to be the most important in entire cohort analysis) loses its importance as the age goes up to the extent that it is not among the top 10 factors in the category of Age\_70s. Additionally, the importance of the features related to depression, such as SSRIs and Dx\_Depression, decreases as the age increases. Specifically, Dx\_Depression feature is not an important factor in age categories of 60s and 70s. 

\renewcommand{\thesubfigure}{\arabic{subfigure}}

Similar to our first experiment, we also identify the top features in any of the 22 strata  as identified by the six different demographic and socioeconomic characteristics  (Fig. \ref{Fig:4}). Note that this time not the fixed 10 features determined from the whole cohort are used.  Fig \ref{Fig:4}.1 shows the important features across different races. Among noticeable patterns, in Fig \ref{Fig:4}.1.B (Black-American stratum), hypertension-related features (including PriorDx\_Hypertens, Dx\_Hypertens) and Hyperlipedemia medications (Ace Inhibitors, Ace Inhibitors and combination), CCB (Calcium Channel Blockers), and CCBs- Dihydropyridines are important factors in this cohort. Hypertension prevalence is higher among minority groups in comparison to whites \citep{fei2017racial}, and the association of hyperlipidemia and obesity have been reported in other studies \citep{sullivan2008impact,bozkurt2016contributory,robbins2009association,klop2013dyslipidemia}, as well as a higher prevalence in Black-Americans in comparison to the whites \citep{morris2009hyperlipidemia}. Hyperlipidemia is a condition in which patient has high levels of lipids (bad cholesterol and triglycerides) in their blood \citep{havel1995management}. The lipid abnormalities include higher serum triglycerides, VLDL, apolipoprotein B, and non-HDL-C levels \citep{feingold2020obesity}. In Fig \ref{Fig:4}.1.C (Latino stratum)  Med-income and Medicaid insurance are showing up among the top 10 important factors. 
Fig \ref{Fig:4}.2, shows the top features across different insurance types. Here, SSRIs, Dx\_Depression, and PriorDx\_Depression are not important in Medicare insurance category, unlike the other insurance types. Instead, 60s, 50s, Priordx\_Hyperlipid, Dx\_Hypelipid, Priordx\_Hypertens, and Dx\_Hypertens are ranked high among patients having the Medicare insurance type.  An interesting pattern with respect to Fig \ref{Fig:4}.2.B (self-pay stratum) is the larger uncertainty ranges  compared to other figures (as shown by the error bars), possibly due to the greater heterogeneity in the characteristics of self-pay patients. 
Fig \ref{Fig:4}.3 (income-level strata) shows that the features related to hyperlipidemia (such as Priordx\_Hyperlipid, and Dx\_Hyperlipid are among the top features in the high-income stratum. In Fig \ref{Fig:4}.4 (neighborhood strata), having the Medicaid insurance is identified as a more important factor for the patients in rural areas. Fig \ref{Fig:4}.5 (gender strata), shows that Dx\_Depression and PriorDx\_Depression are among the important factors in women and SSRIs in the both strata. This pattern is conforming with recent studies reporting a positive association of depression with obesity in general population and a higher association among women \citep{de2010depression,simon2008association,li2017gender}. The results also indicate Priordx\_Hyperlipid and Dx\_Hyperlipid are among important features in the men strata which is consistent with other studies \citep{deng2012prevalence,brown2000body,kawada2002body}. In Fig \ref{Fig:4}.6 (age strata), a noticeable pattern relates to decreasing the rank of taking antidepression medications, as the age increases, which relates to the reports showing higher prevalence of depression among younger people specifically those in 24-30 \citep{cuijpers2020psychotherapy}. Having a commercial insurance is not among the top for the 60s and 70s strata. On the other side, PriorDx\_Hyperlipid, Dx\_Hyperlipid, and antihyperlipidemic class of medications (such as statins and antihyperlipidemic – HMG CoA Reductase Inhibitors) are important in older patients (those in their 50s, 60s, and 70s). The ranking of hyperlipidemia and antihyperlipidemic medication also increases with the patient’s age, aligned with the higher prevalence of hyperlipidemia in older adults \citep{pencina2014application,navar2015hyperlipidemia}. Other noticeable features are the antihypertension CCBs (Calcium Channel Blockers) and dihydropyridines (also used as antihypertension) are ranked high in the age 60s stratum. Medicare is also an important feature among those in their 60s and 70s. Dx\_Obesity appears as top feature only in the under 30s and 70s strata stratum. More details about the specified feature across different categories are available in Appendix \ref{secA3}.

Lastly, we note some limitations of our study. We only study the weight gain patterns within a two-year frame based on the considerations about not choosing a too short period and not excluding too many samples. The two-year period may still offer a reasonable gap for engaging in timely interventions before seeing the adverse outcomes of excessive weight gain. Our approach does not consider all the concerning patterns of dangerous weight gain such as transitioning from the normal to overweight. Still, our work captures some of the most concerning weight patterns. The way we define our positive class also captures weight cycling patterns (also known as the yo-yo effect) \cite{Brownell94}. Our results are based on the analysis in one dataset. Despite this, our study is among the largest of this kind. We also tried to address many of the concerns of this type by running significant sensitivity analyses demonstrating the robustness of our results.

\section{Conclusion}\label{sec3}
In this study, aiming to identify the most important factors that lead to dangerous weight gains, we generated a series of {\it X → Y} ({\it if-then}) rules predicting such patterns using a very large longitudinal dataset. To do this, we presented a new subgroup discovery method inspired by existing data mining methods. In our method, we first rank the generated rules using the {\it WRAcc} (weighted relative accuracy) value for each rule, and then determine the most important features on the left-side that can predict the right side. We identify these important features by calculating the weighted average {\it WRAcc} of the top {\it n} rules that each feature appears in. Through a series of sensitivity analysis experiments, we show that our results are robust to the parameter choices. Applying this method to our weight gains problem reveal various patterns in our large dataset and across 22 different strata (determined by different demographic and socioeconomic factors). Specifically, we show how the most important features predicting dangerous weight gains differ across individuals with different sex, age, race/ethnicity, insurance type, residence type, and income-level. Our findings may inform more effective interventions of obesity. Moreover, our application agnostic rule-discovery method can be applied to other similar problems. 

\FloatBarrier
\bibliography{sn-article}

\hfill \break
\hfill \break
\textbf{Mina Samizadeh} is a Ph.D. student at University of Delaware. She works on the application of Machine Learning in solving health related problems.
\begin{figure}[H]
  \includegraphics[width=.5\linewidth, height=.25\textheight]{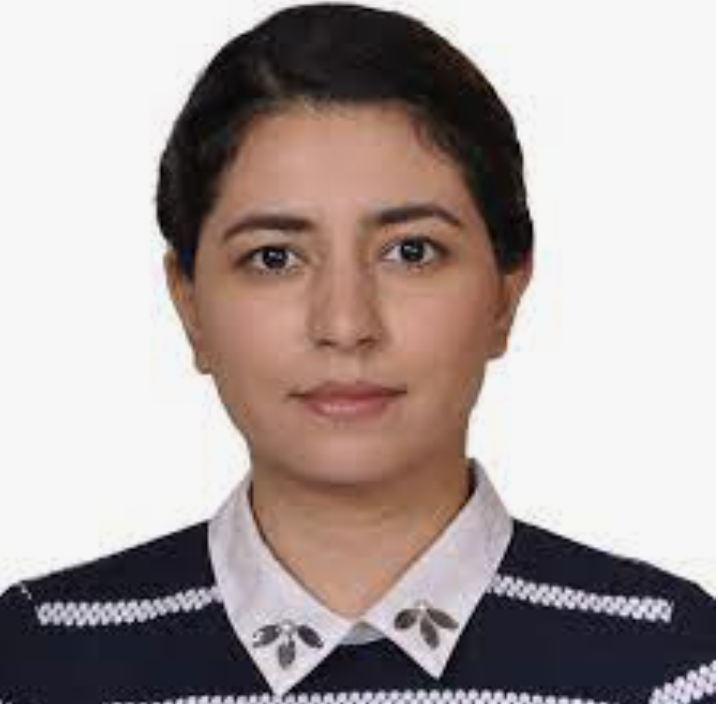}
  \end{figure}

\textbf{Jessica Jones-Smith} is an associate professor at University of Washington who investigates socioeconomic causes and correlates of obesity risk in both high- and low/middle-income countries. 
\begin{figure}[H]
  \includegraphics[width=.5\linewidth, height=.25\textheight]{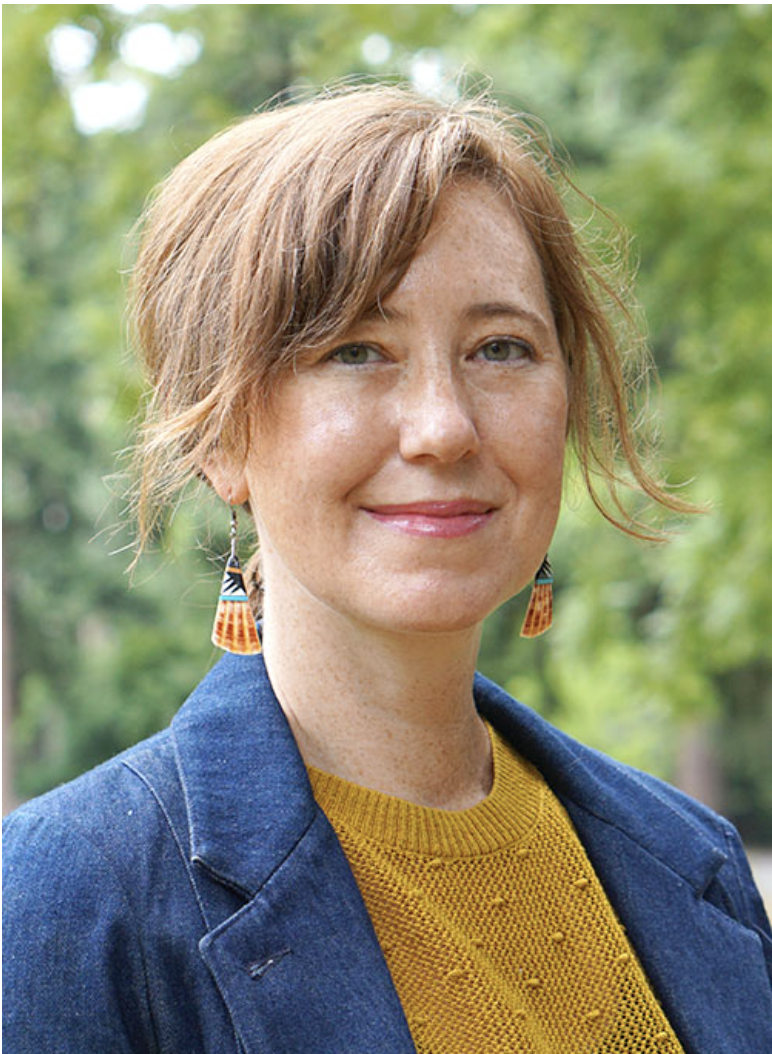}
  \end{figure}

\textbf{Bethany Sheridan} was leading athenahealth’s Research and Insights team, a group of researchers and data scientists dedicated to expanding knowledge on clinical operations, care patterns, and public health trends from athenahealth’s unique data asset. 
\begin{figure}[H]
  \includegraphics[width=.5\linewidth, height=.25\textheight]{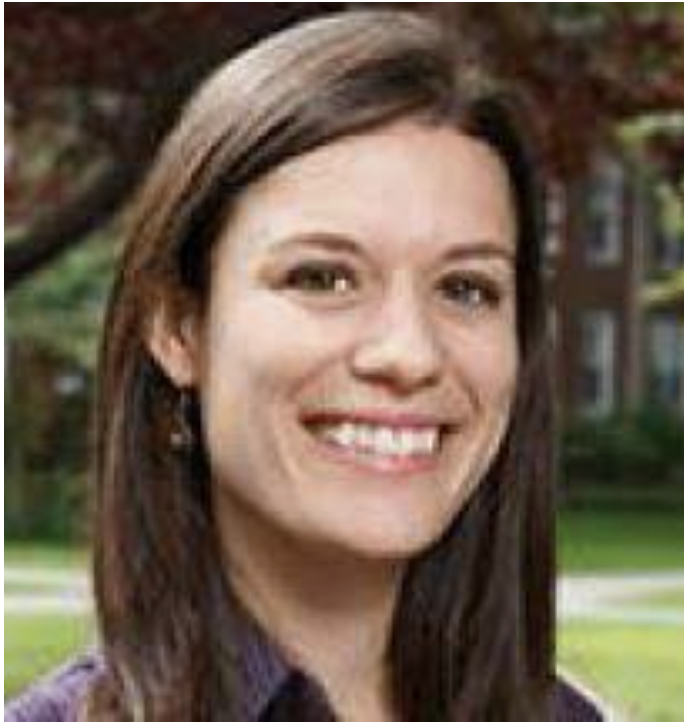}
  \end{figure}

\textbf{Rahmat Beheshti} is an assistant professor at University of Delaware. He works in the area of Applied Machine Learning and Health Data Science and directs the healthy lAife lab at UD. 
\begin{figure}[H]
  \includegraphics[width=.5\linewidth, height=.25\textheight]{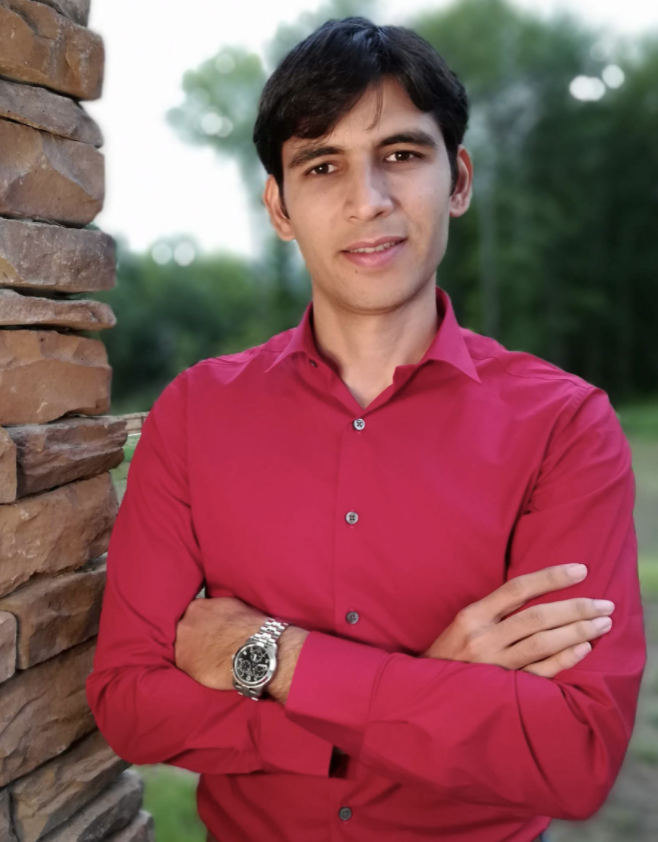}
  \end{figure}
\begin{appendices}
\renewcommand{\thesubfigure}{\arabic{subfigure}}
\section{}\label{secA1}
\begin{figure}[H]
  \begin{subfigure}[H]{\linewidth}
  \includegraphics[width=.45\linewidth, height=.18\textheight]{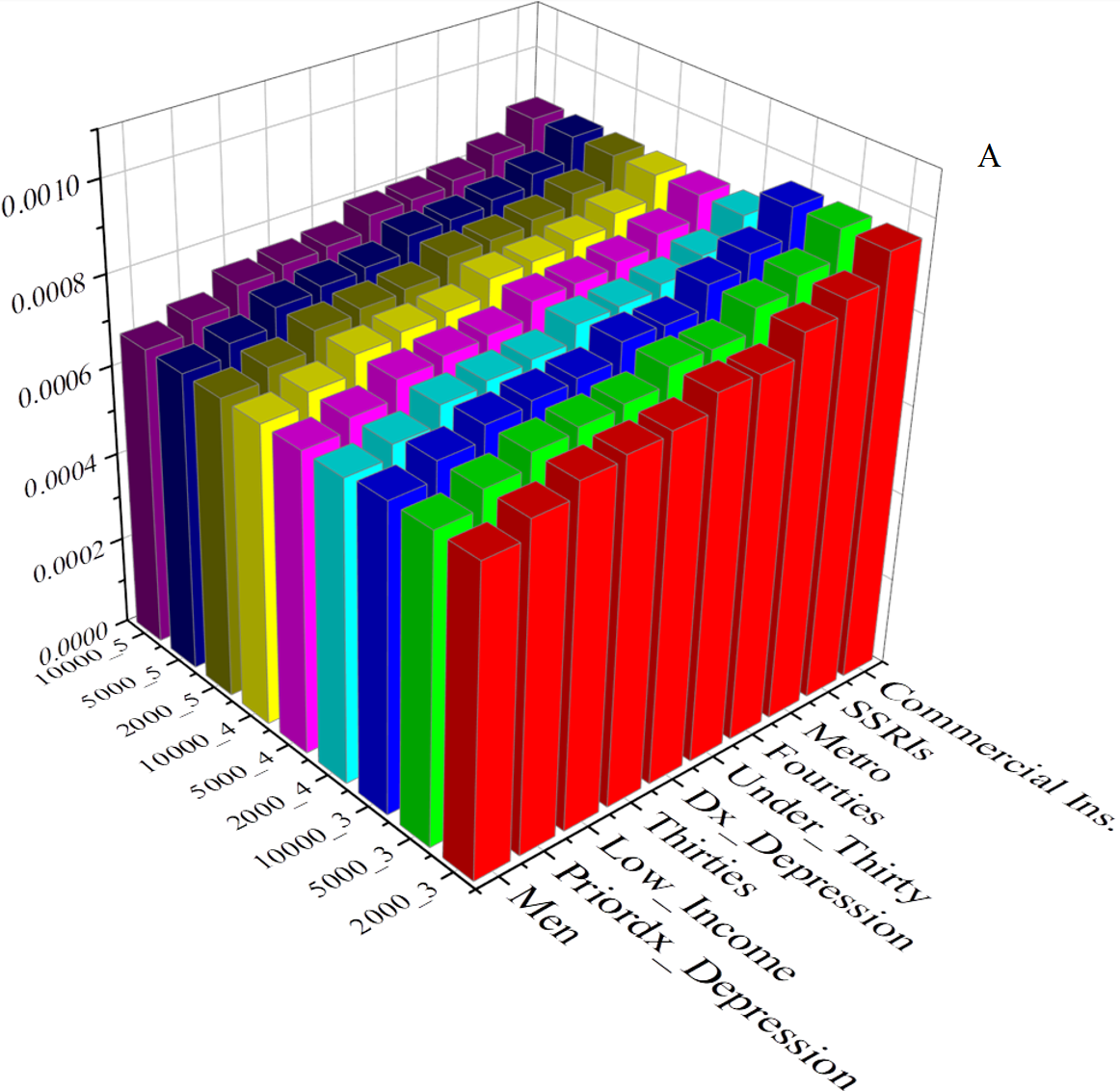} \hfill
  \includegraphics[width=.45\linewidth, height=.18\textheight]{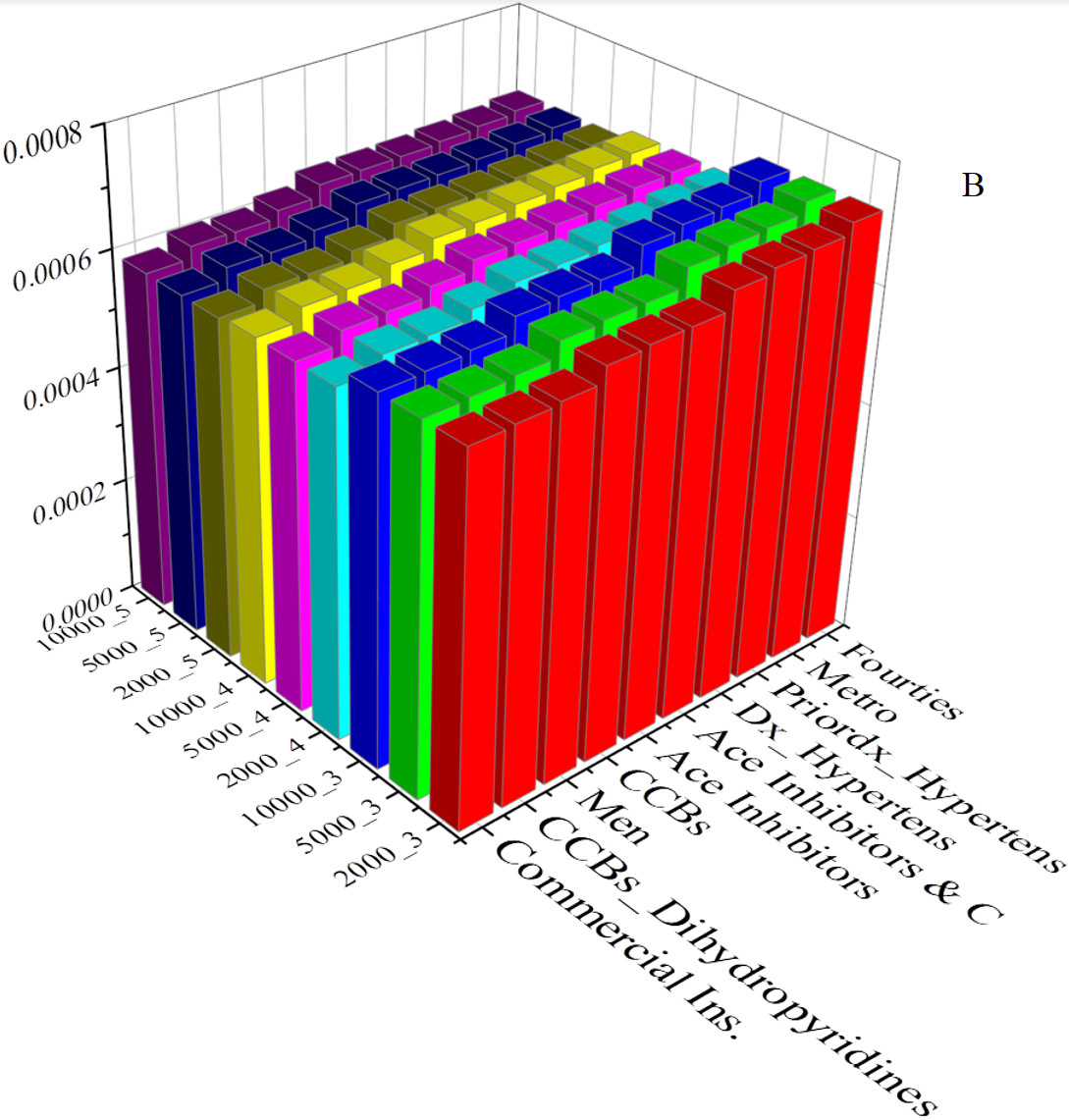}\hfill
  \includegraphics[width=.45\linewidth, height=.18\textheight]{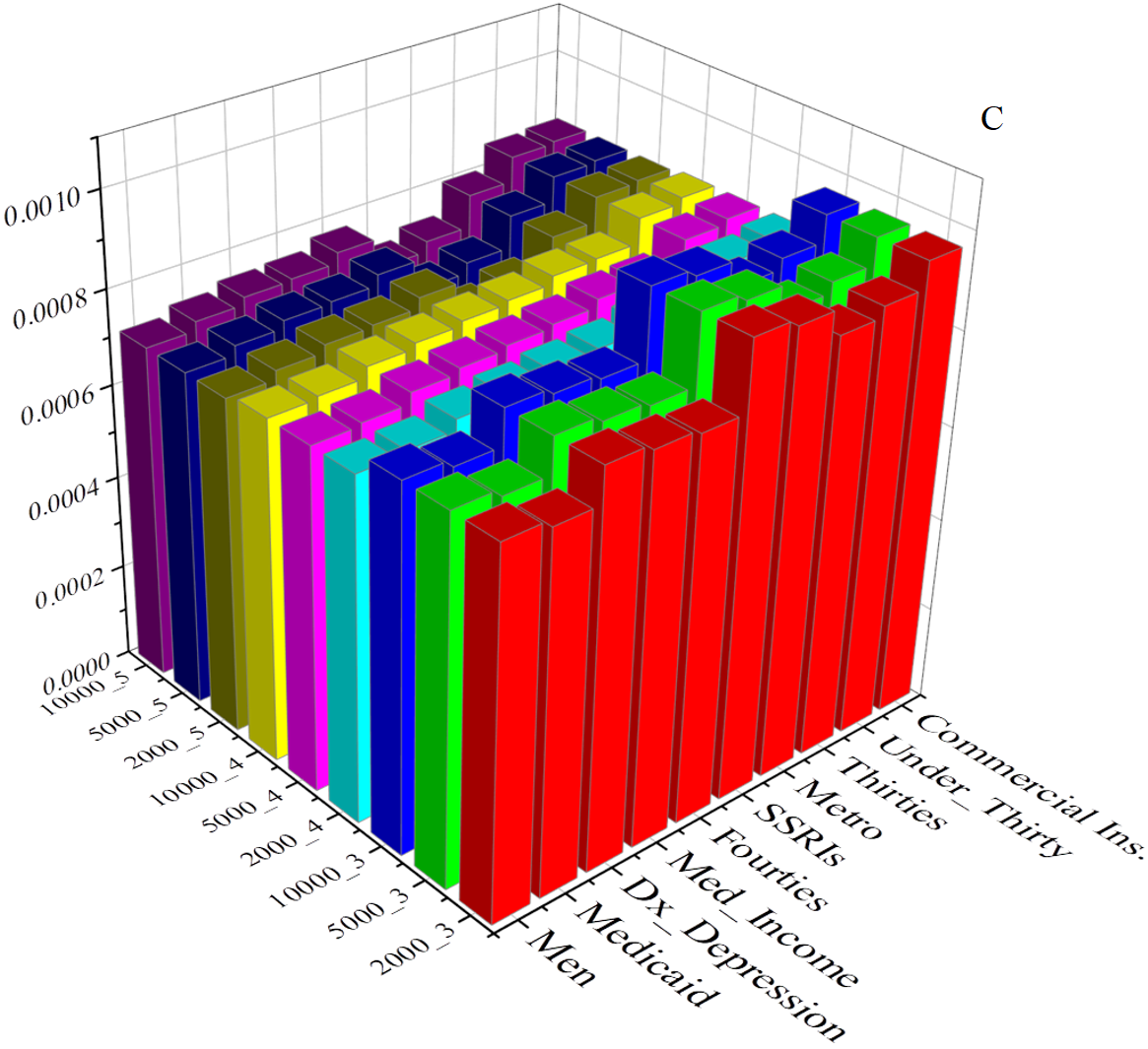}\hfill
  \includegraphics[width=.45\linewidth, height=.18\textheight]{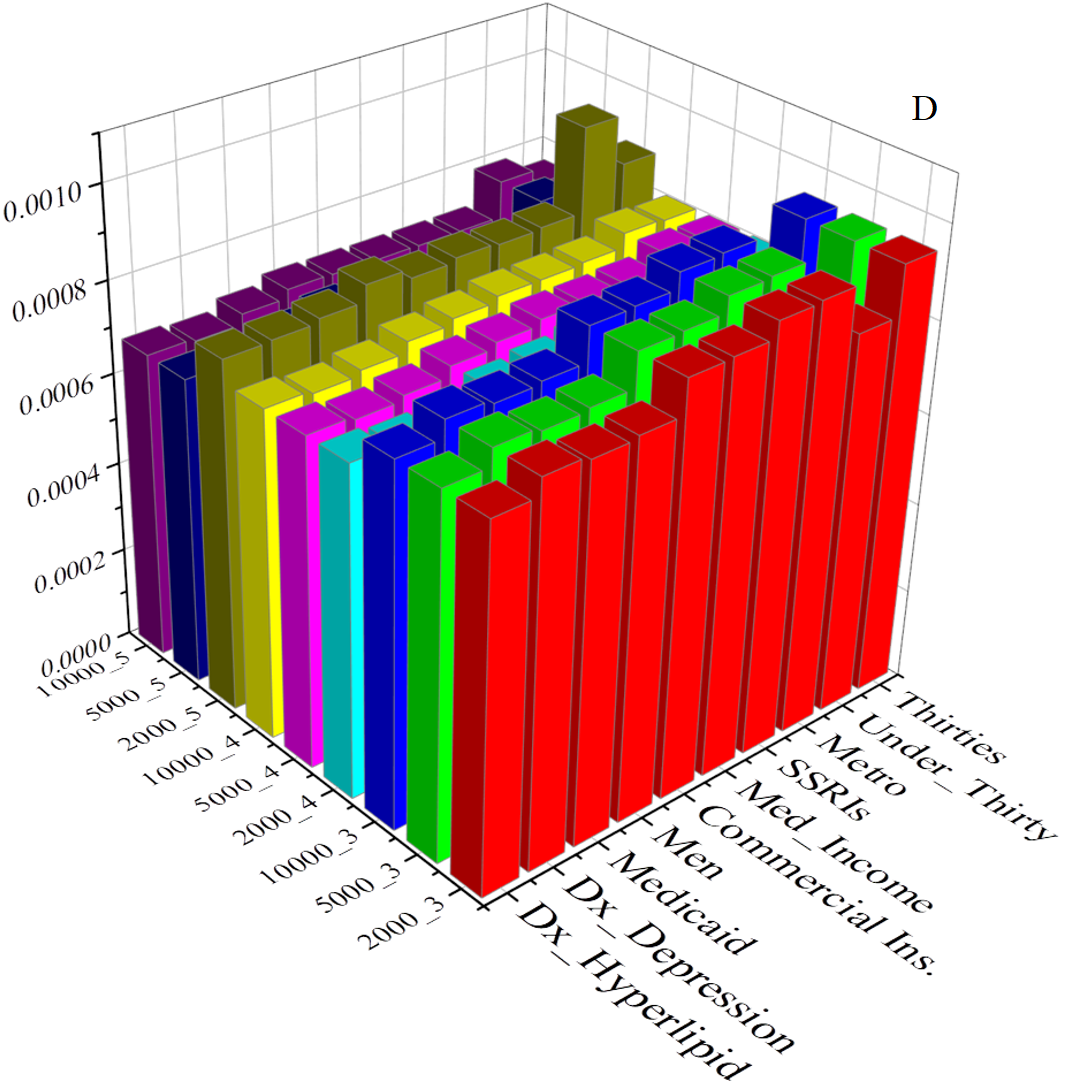}
  \caption{Race}
  \end{subfigure}\par\medskip
\end{figure}%
\begin{figure}[ht]\ContinuedFloat  
  \begin{subfigure}{\linewidth}
  \includegraphics[width=.5\linewidth, height=.25\textheight]{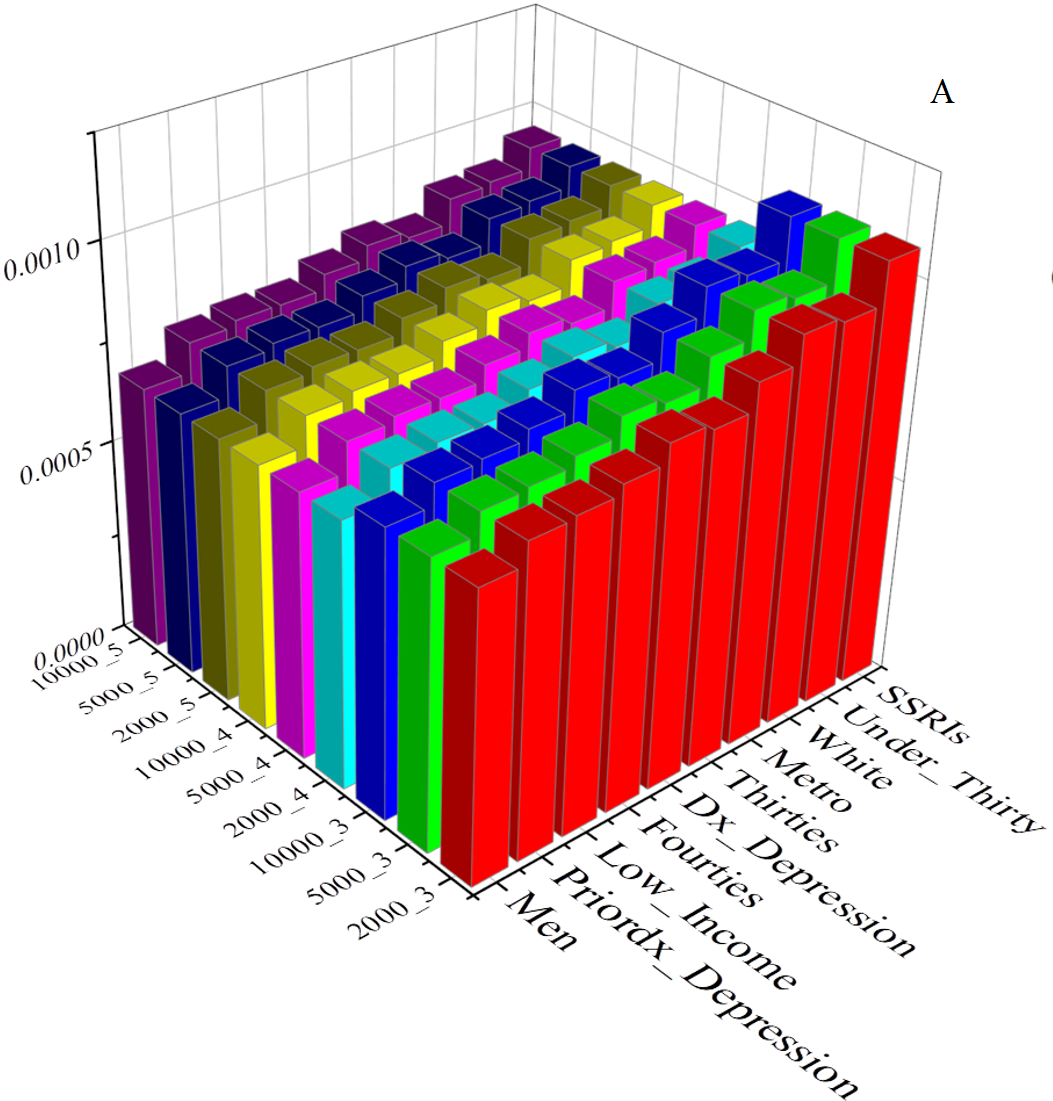}\hfill
  \includegraphics[width=.5\linewidth, height=.25\textheight]{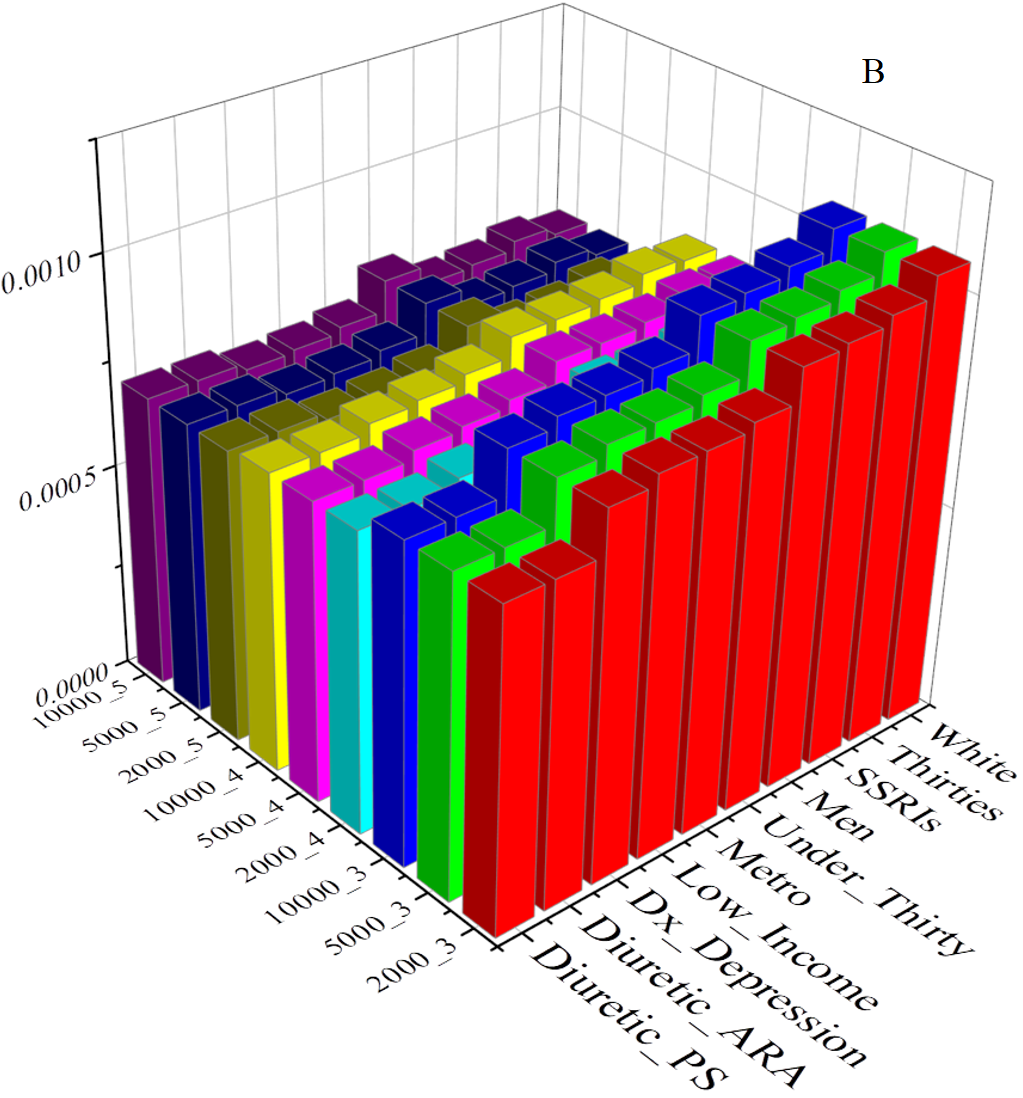}\hfill
  \includegraphics[width=.5\linewidth, height=.25\textheight]{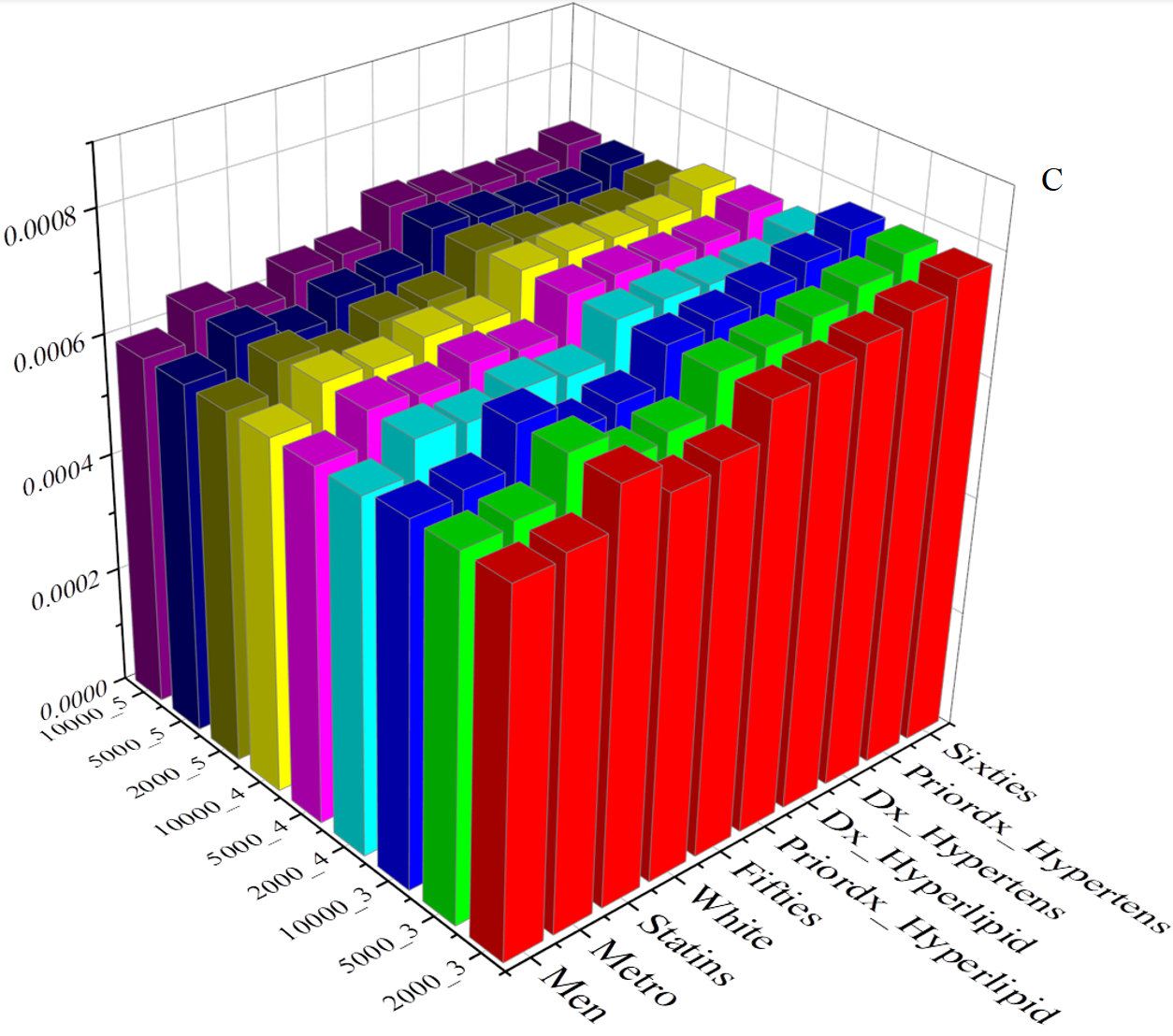}\hfill
  \includegraphics[width=.5\linewidth, height=.25\textheight]{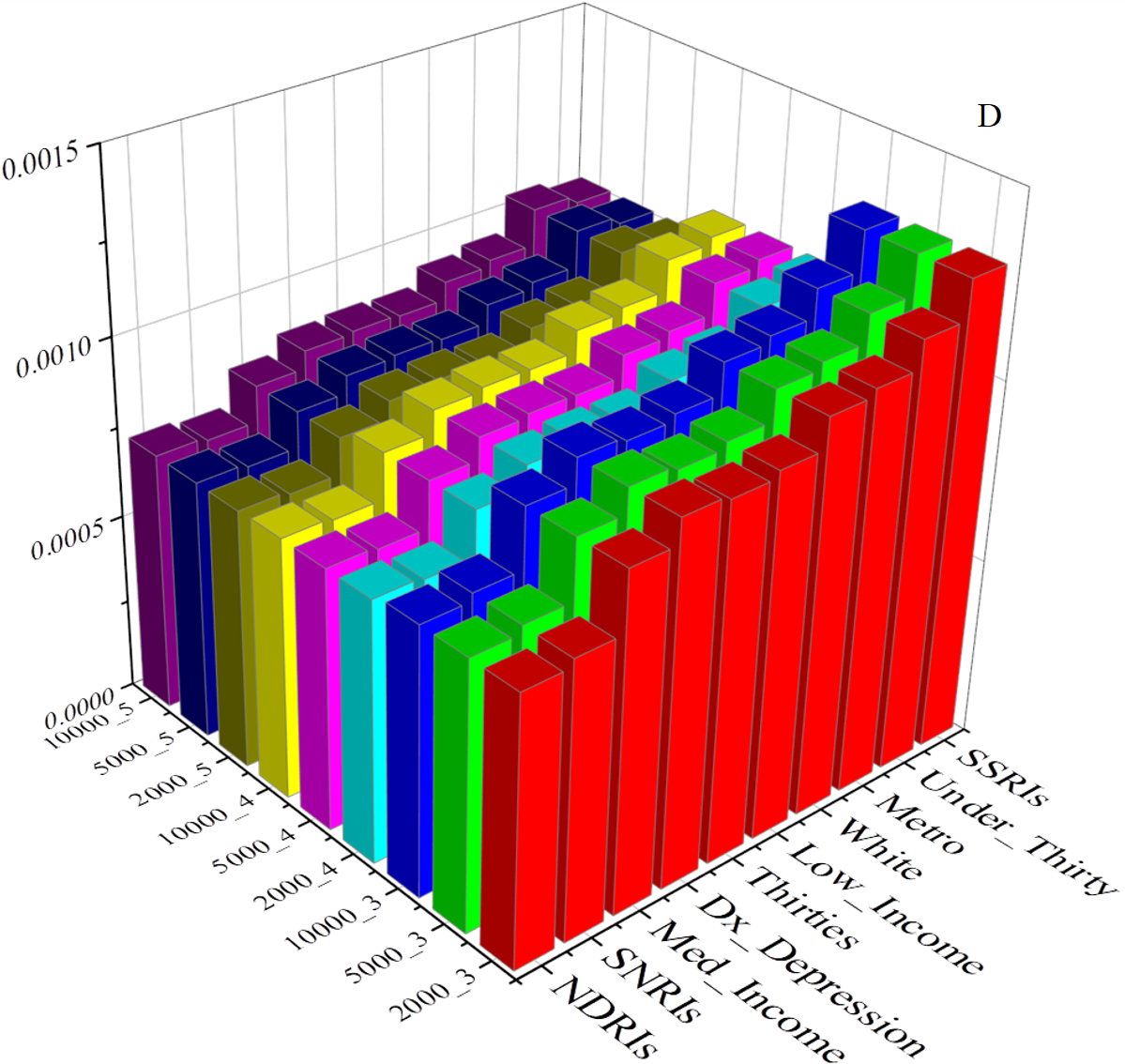}
  \caption{Insurance}
  \end{subfigure}\par\medskip

  \begin{subfigure}{\linewidth}
  \includegraphics[width=.5\linewidth, height=.25\textheight]{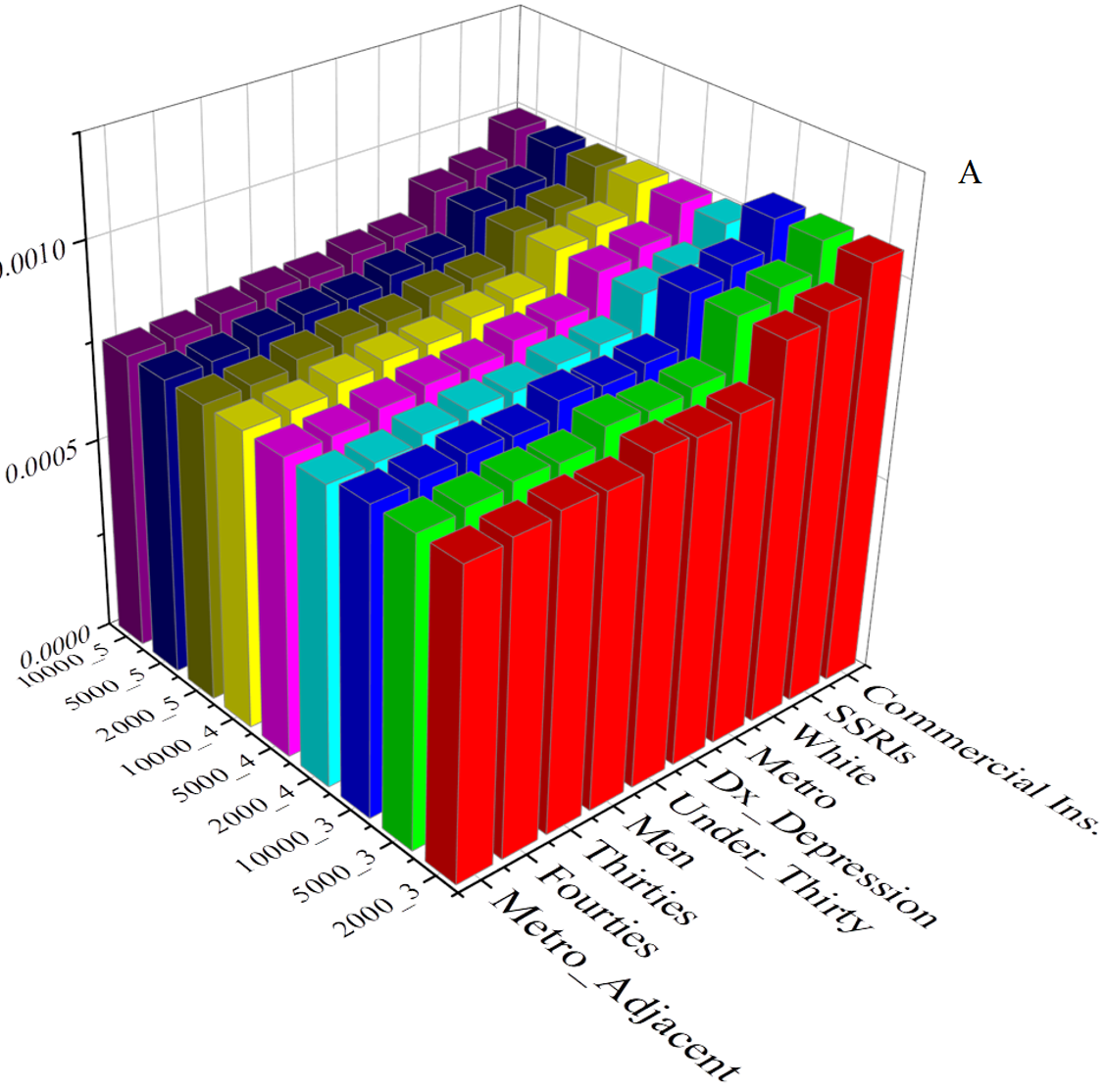}\hfill
  \includegraphics[width=.5\linewidth, height=.25\textheight]{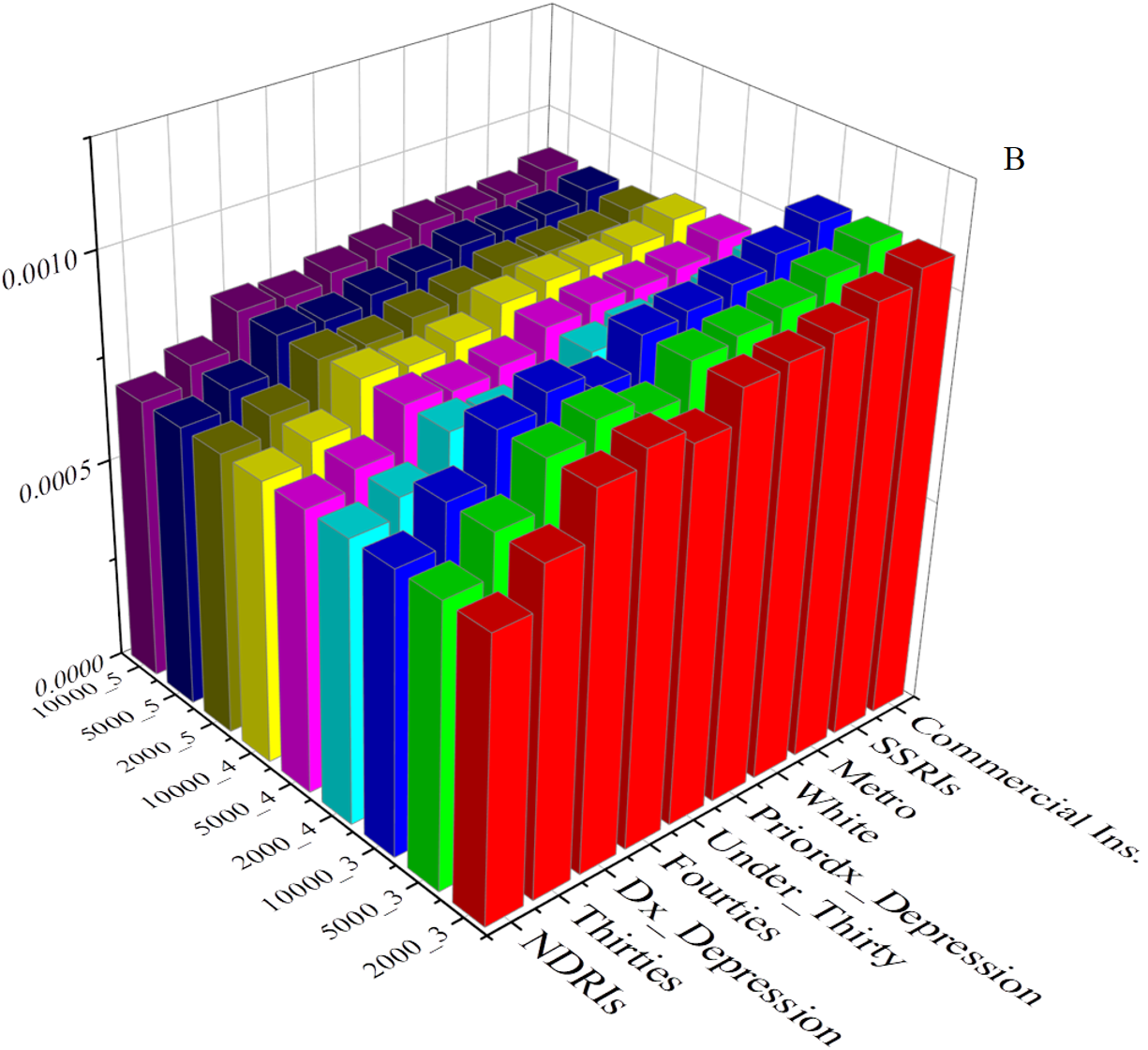}\hfill
  \includegraphics[width=.5\linewidth, height=.25\textheight]{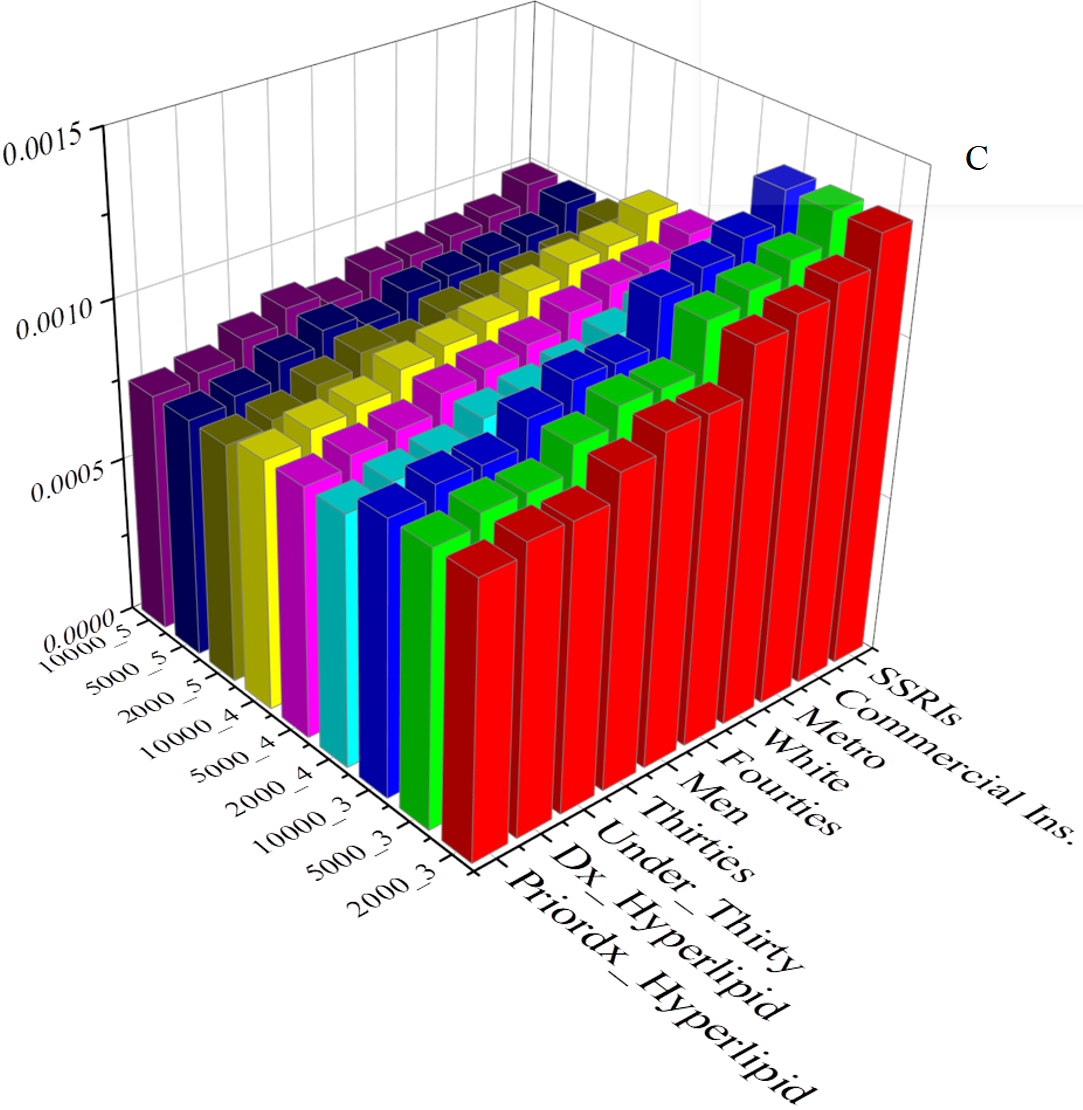}
  \caption{Income}
  \end{subfigure}\par\medskip
 \end{figure}%
\begin{figure}[ht]\ContinuedFloat 
  \begin{subfigure}{\linewidth}
  \includegraphics[width=.5\linewidth, height=.25\textheight]{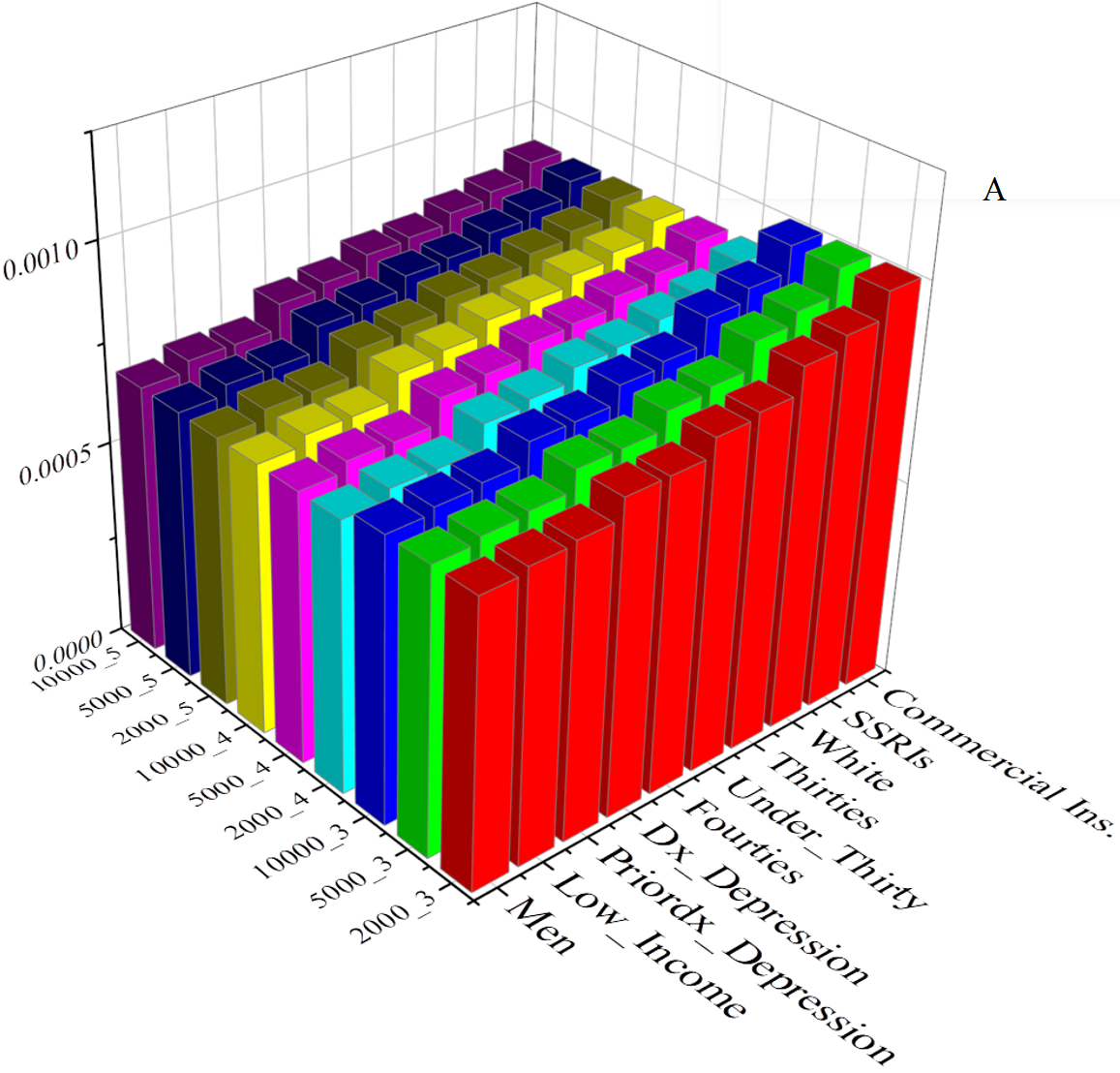}\hfill
  \includegraphics[width=.5\linewidth, height=.25\textheight]{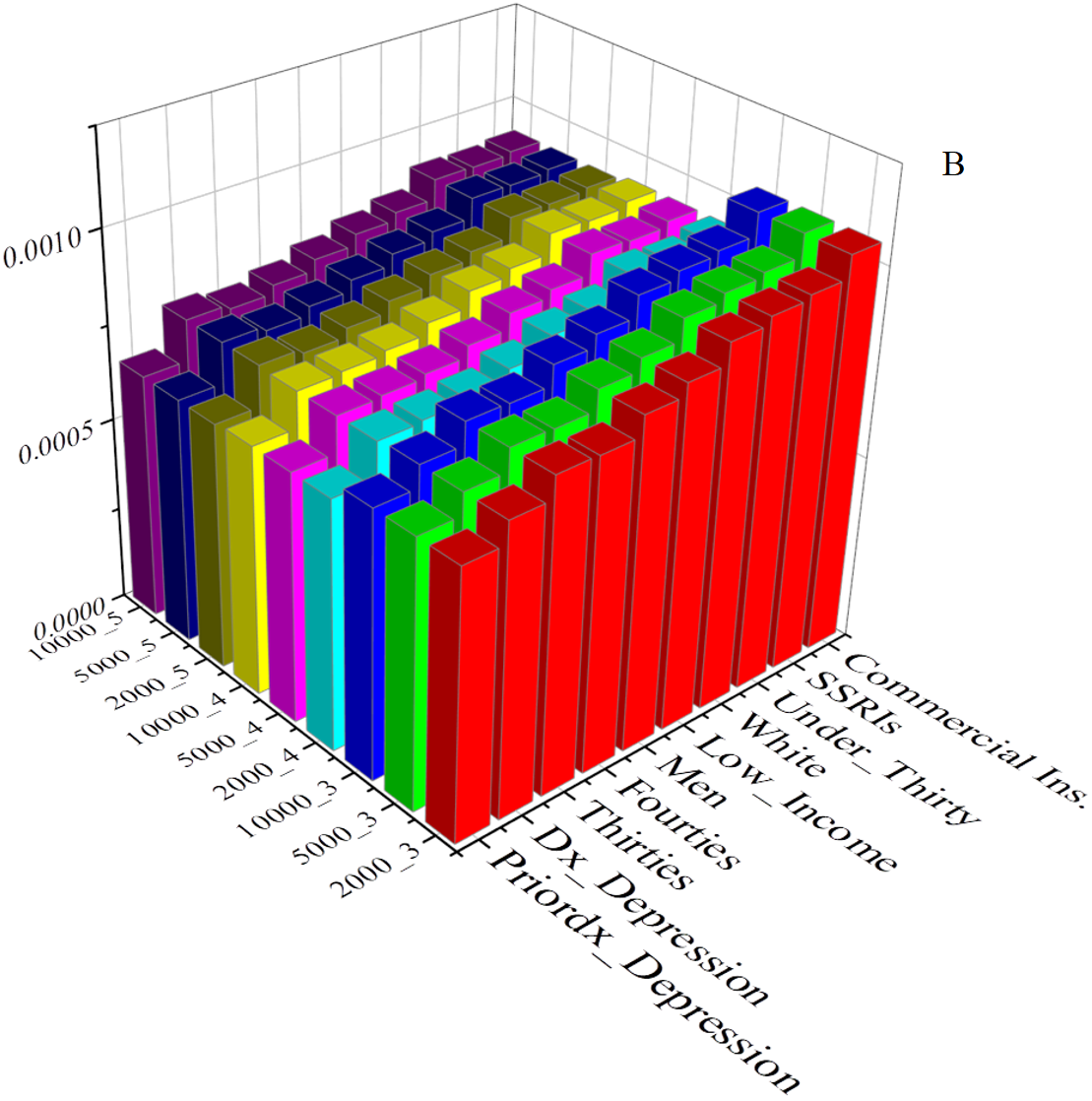}\hfill
  \includegraphics[width=.5\linewidth, height=.25\textheight]{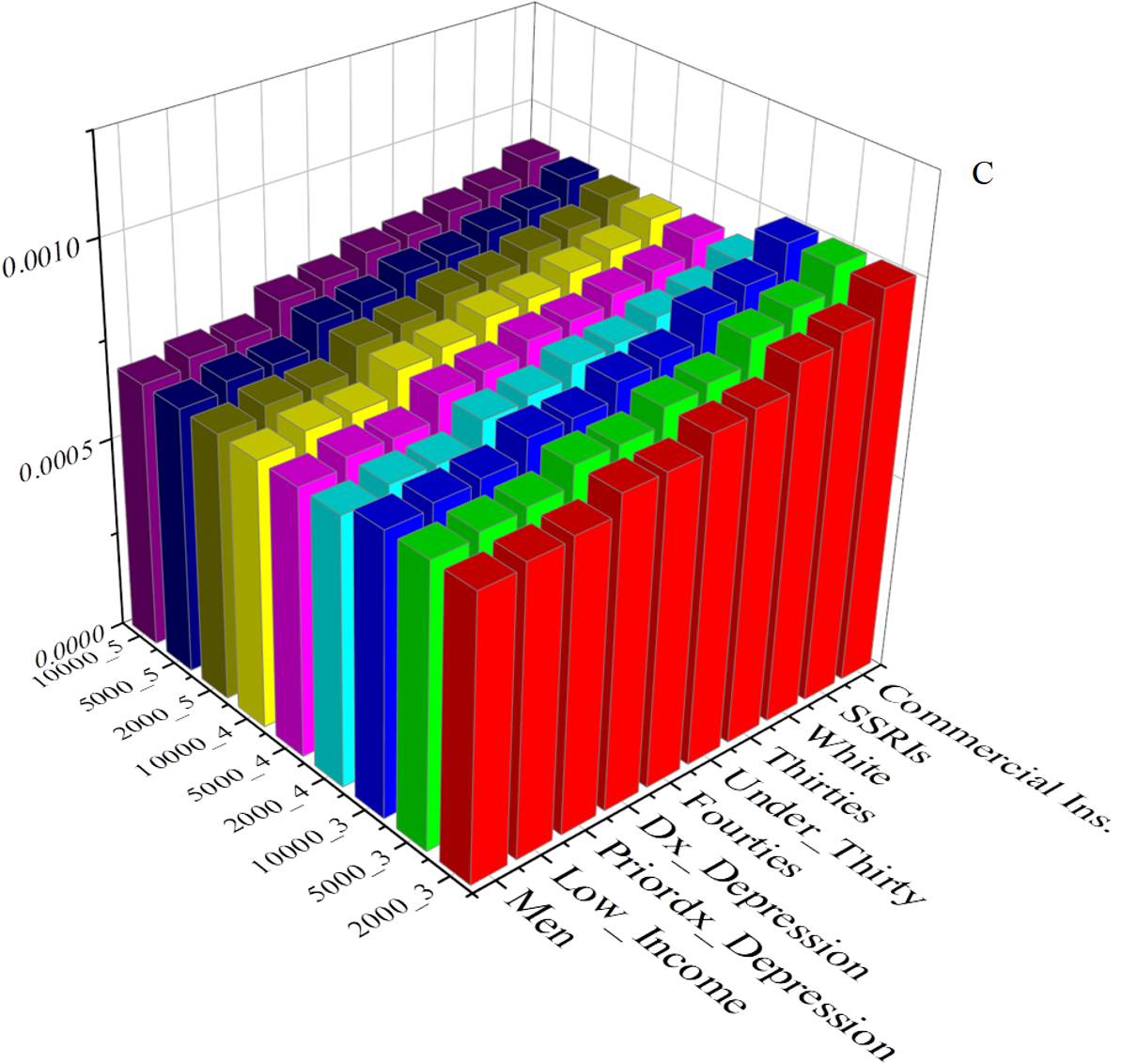}
  \caption{Neighborhood}
  \end{subfigure}\par\medskip 
  \begin{subfigure}{\linewidth}
  \includegraphics[width=.5\linewidth, height=.25\textheight]{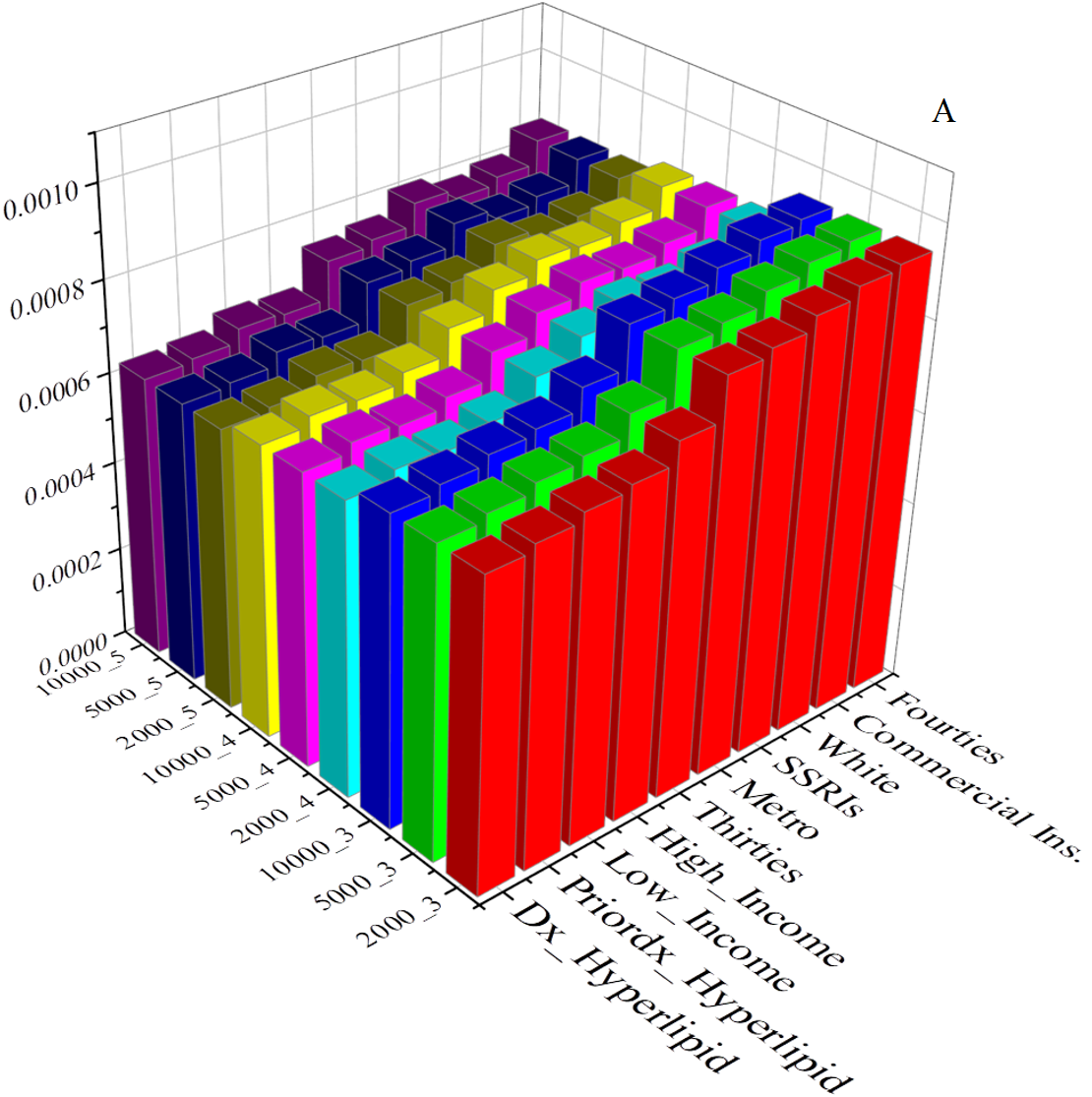}\hfill
  \includegraphics[width=.5\linewidth, height=.25\textheight]{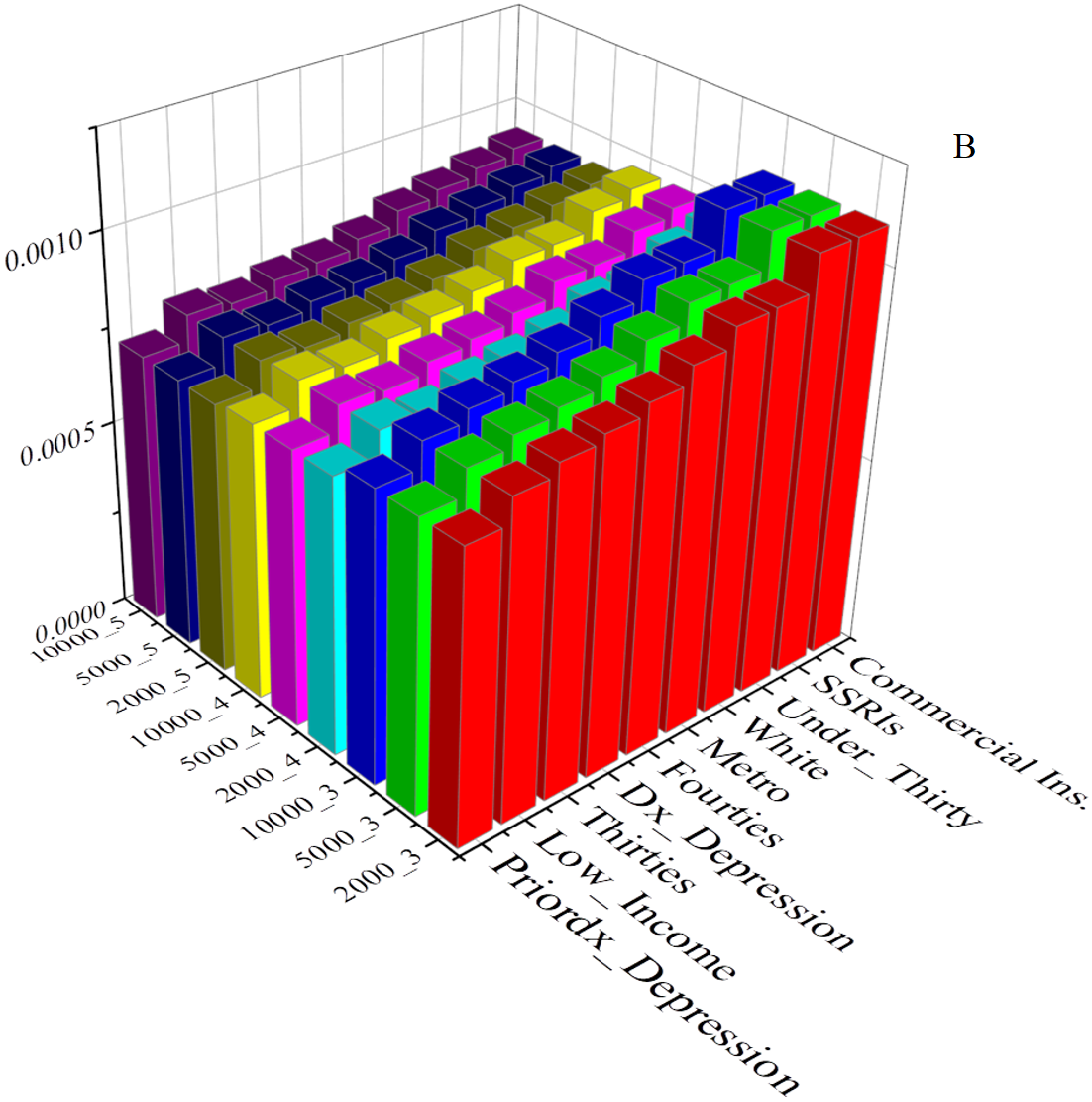}
  \caption{Gender}
  \end{subfigure}\par\medskip
\end{figure}%
\begin{figure}[ht]\ContinuedFloat  
  \begin{subfigure}{\linewidth}
  \includegraphics[width=.5\linewidth]{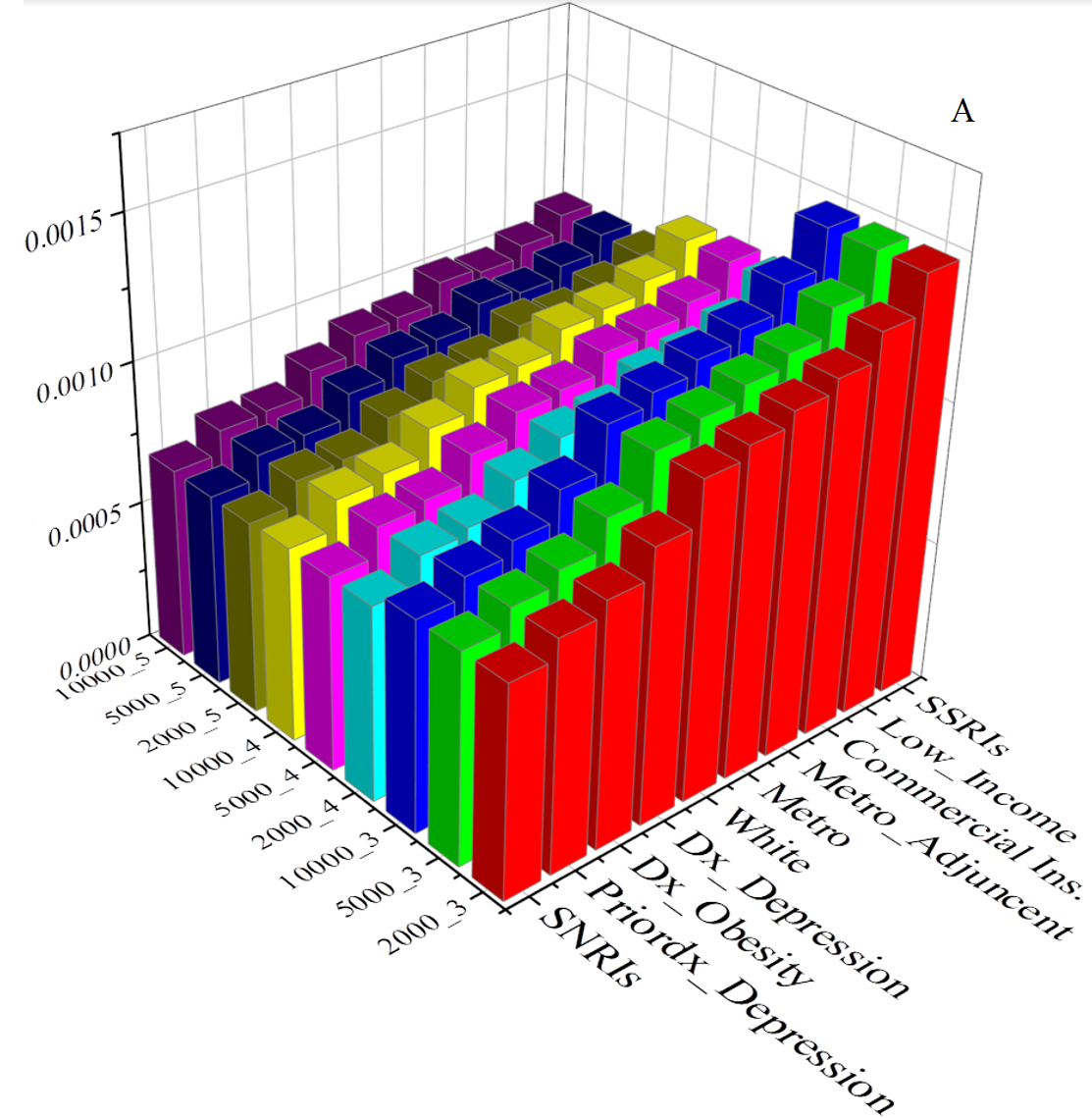}\hfill
  \includegraphics[width=.5\linewidth]{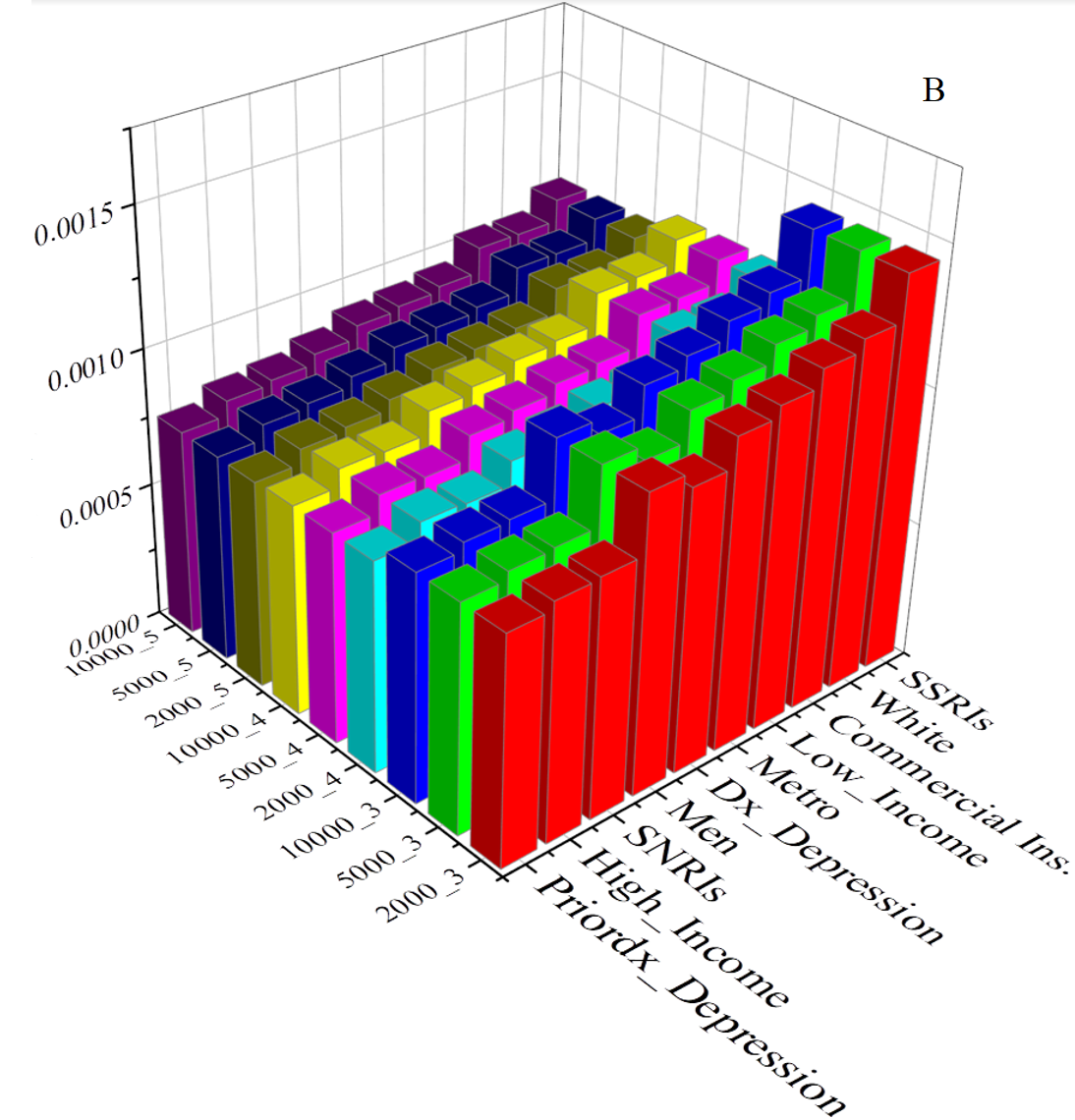}\hfill
  \includegraphics[width=.5\linewidth]{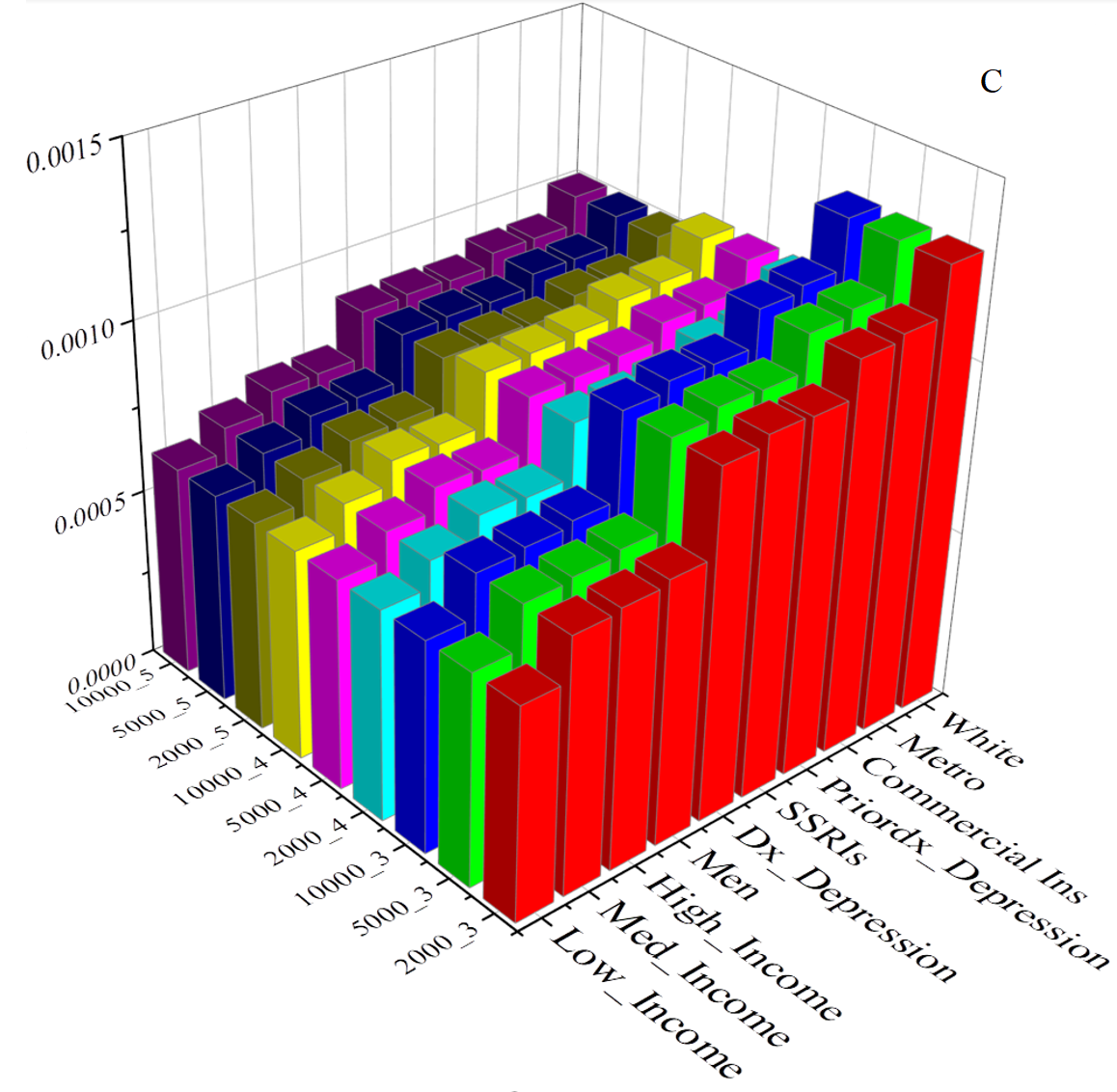}\hfill
  \includegraphics[width=.5\linewidth]{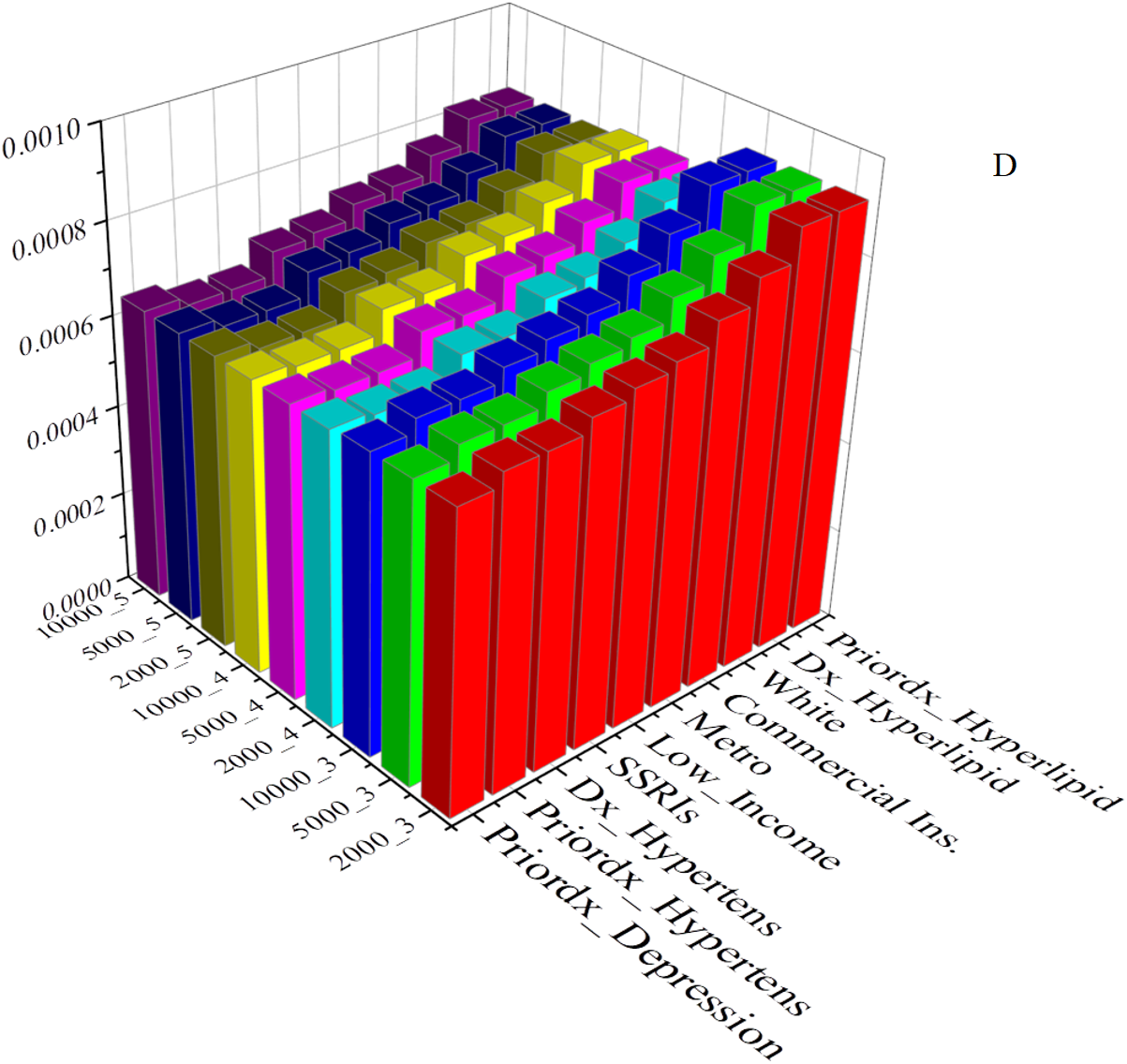}\hfill
  \includegraphics[width=.5\linewidth]{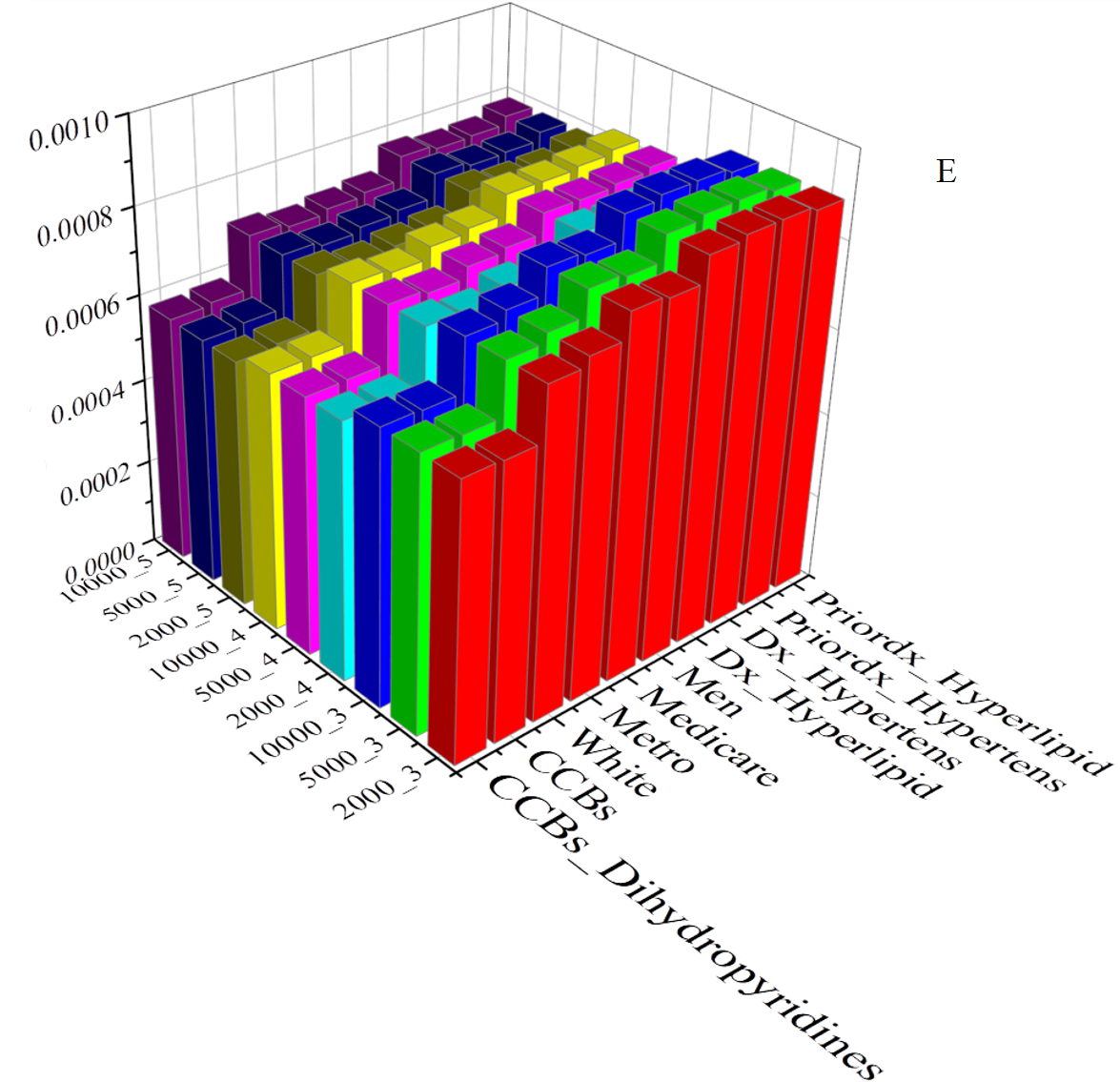}\hfill
  \includegraphics[width=.5\linewidth]{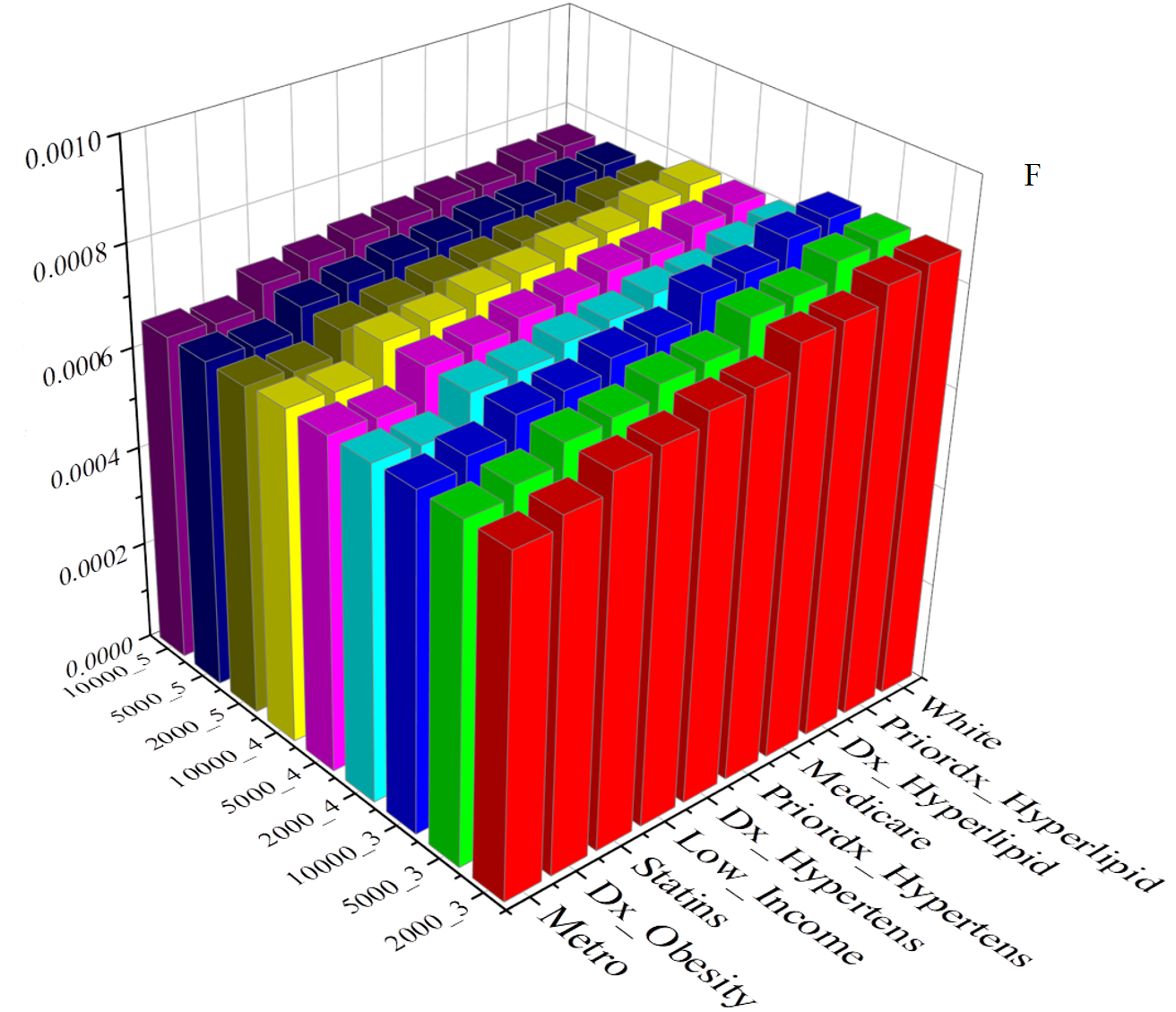}
  \caption{Age}
  \end{subfigure}\\[-1ex]
  \caption{Sensitivity Analysis of Weighted Average Quality of important features in low-level (medication-level) data contributing to obesity across 6 different groups of 1) Race, 2) Insurance, 3) Income, 4) Neighborhood, 5) Gender, and 6) Age in 9 different experimental settings. The vertical (Y) axis demonstrates WA, the X axis demonstrate the setting of experiment (width of the beam search\_max length of the rules), and the Z axis shows the name of features}
  \label{Fig:5}
\end{figure}
\FloatBarrier

\section{}\label{secA2}

\begin{figure}[H]
  \begin{subfigure}{\linewidth}
  \includegraphics[width=.5\linewidth, height=.25\textheight]{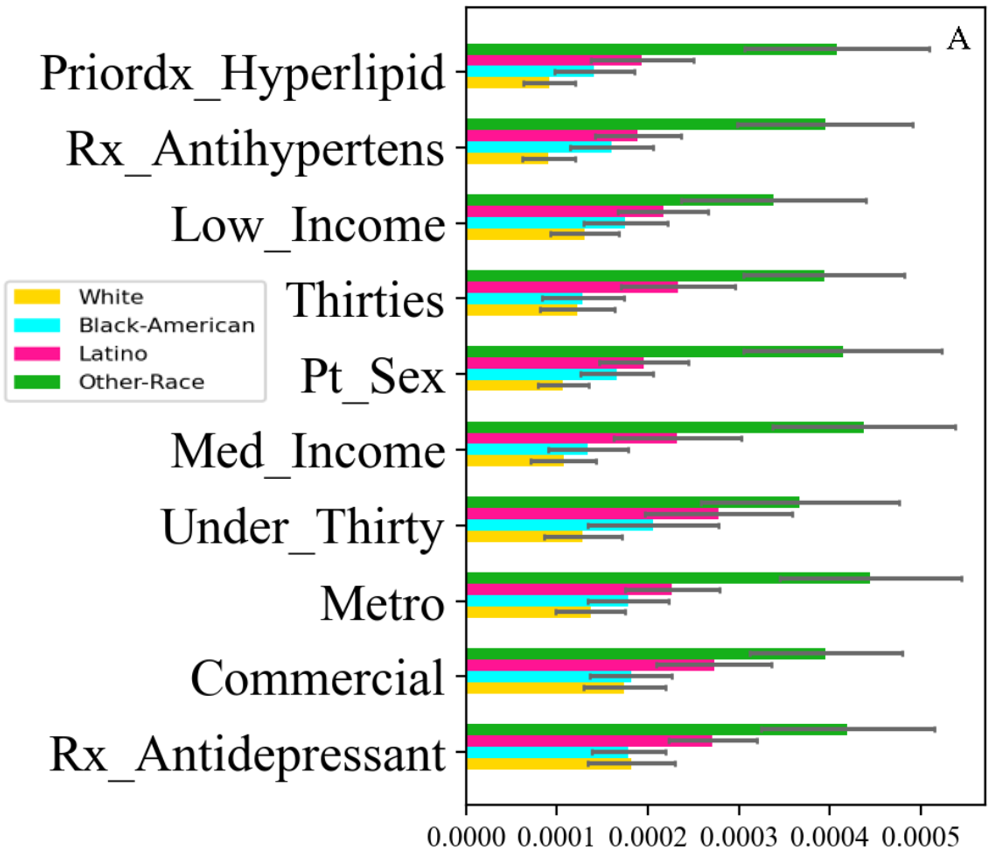}\hfill
  \includegraphics[width=.5\linewidth, height=.25\textheight]{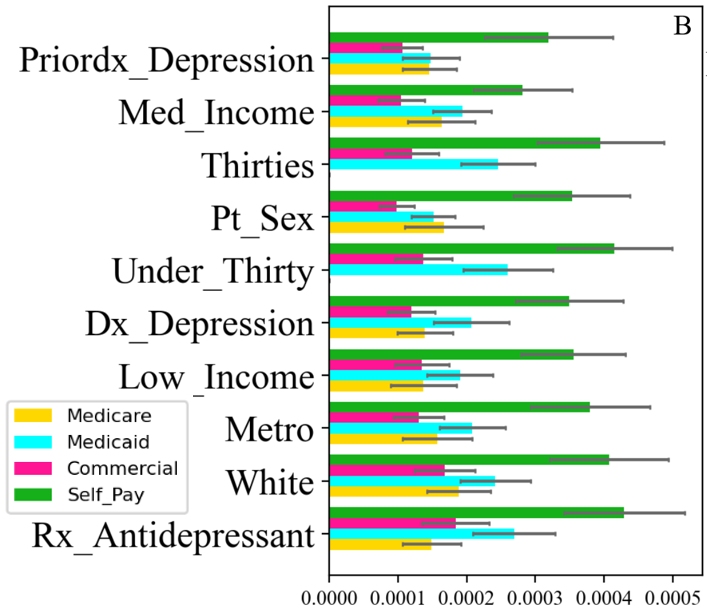}
  
  \end{subfigure}\par\medskip
  \begin{subfigure}{\linewidth}
  \includegraphics[width=.5\linewidth, height=.25\textheight]{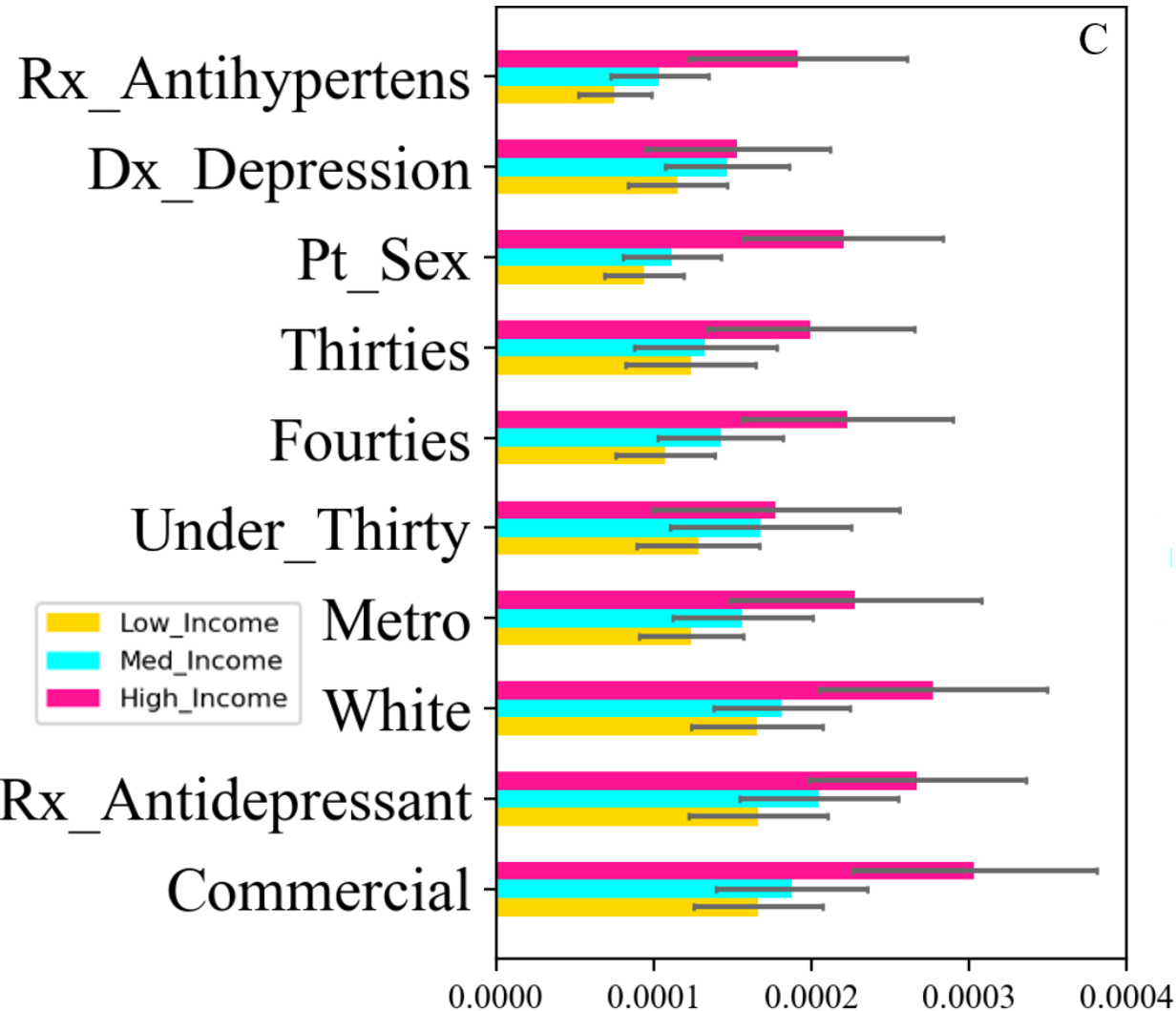}\hfill
  \includegraphics[width=.5\linewidth, height=.25\textheight]{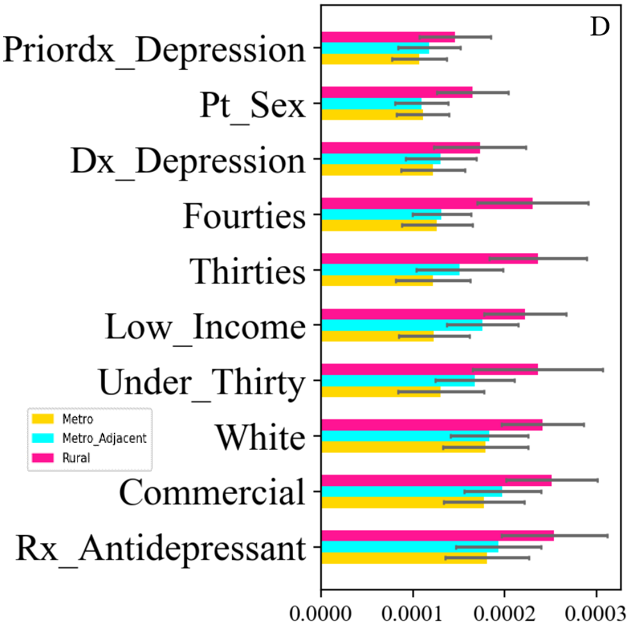}
  
  \end{subfigure}\par\medskip
  \begin{subfigure}{\linewidth}
  \includegraphics[width=.5\linewidth, height=.25\textheight]{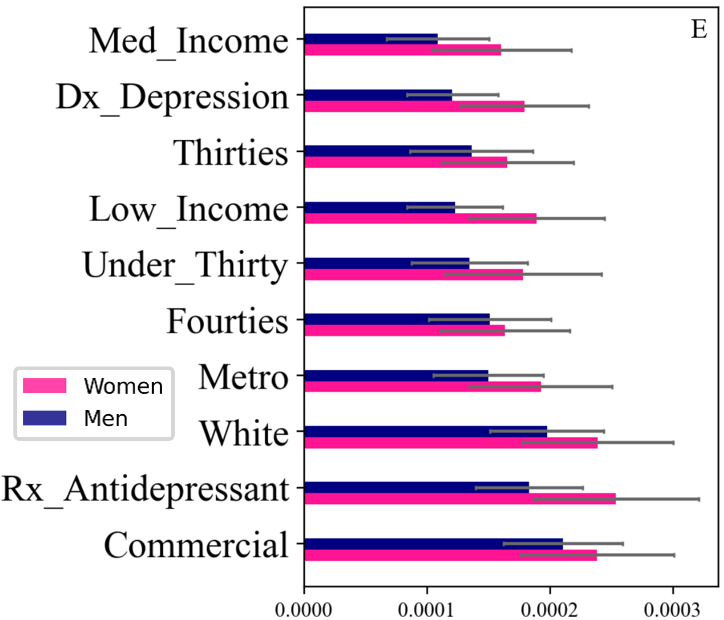}\hfill
  \includegraphics[width=.5\linewidth, height=.25\textheight]{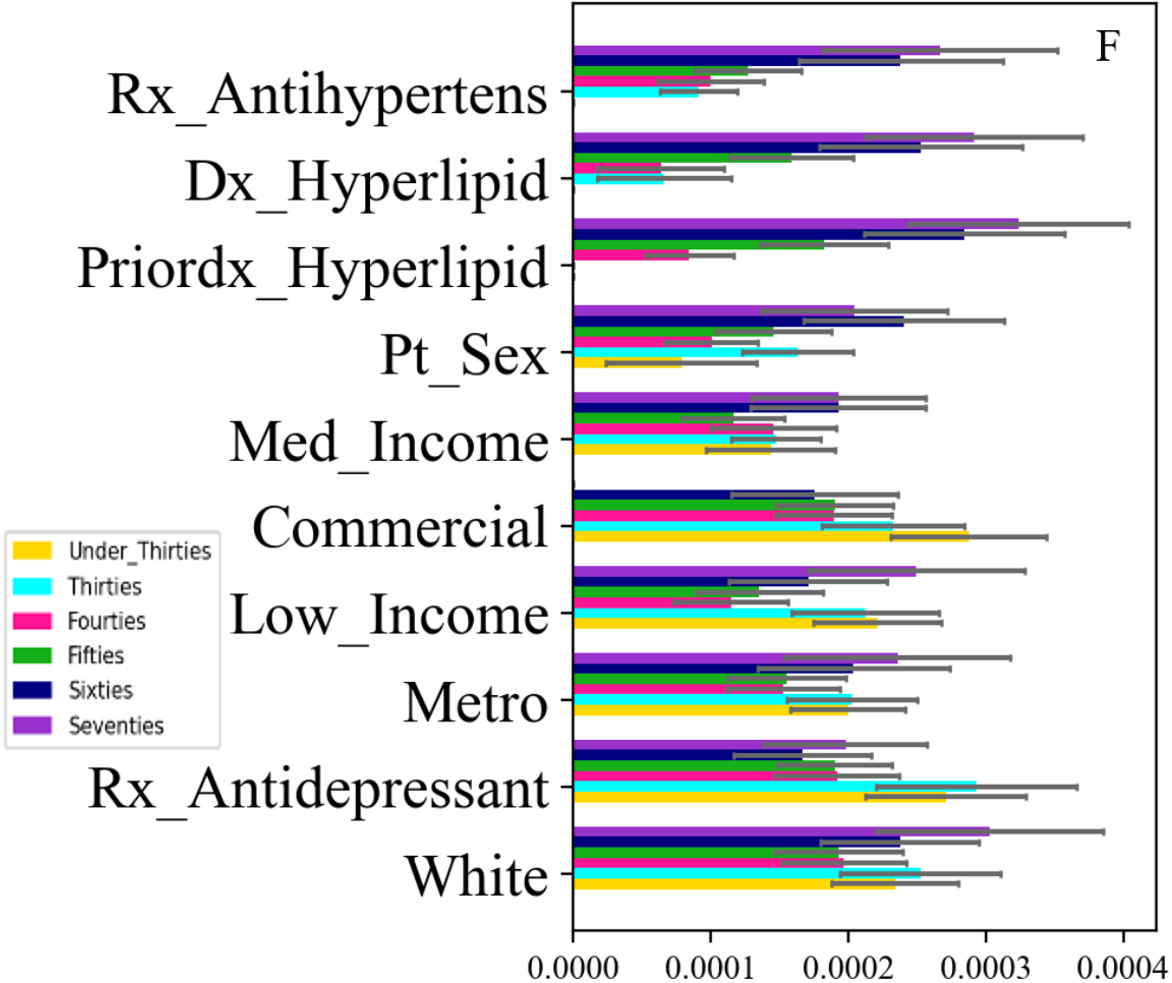}
  
  \end{subfigure}
  \caption{Comparison between average weighted quality of visit level (high-level) features with the highest summation across different categories of different groups consisting of: A) Race, B) Insurance, C) Income, D) Neighborhood, E) gender, F) Age.}
  \label{Fig:6}
\end{figure}

\FloatBarrier

\section{}\label{secA3}

\begin{figure}[H]
  \begin{subfigure}{\linewidth}
  \includegraphics[width=.5\linewidth, height=.25\textheight]{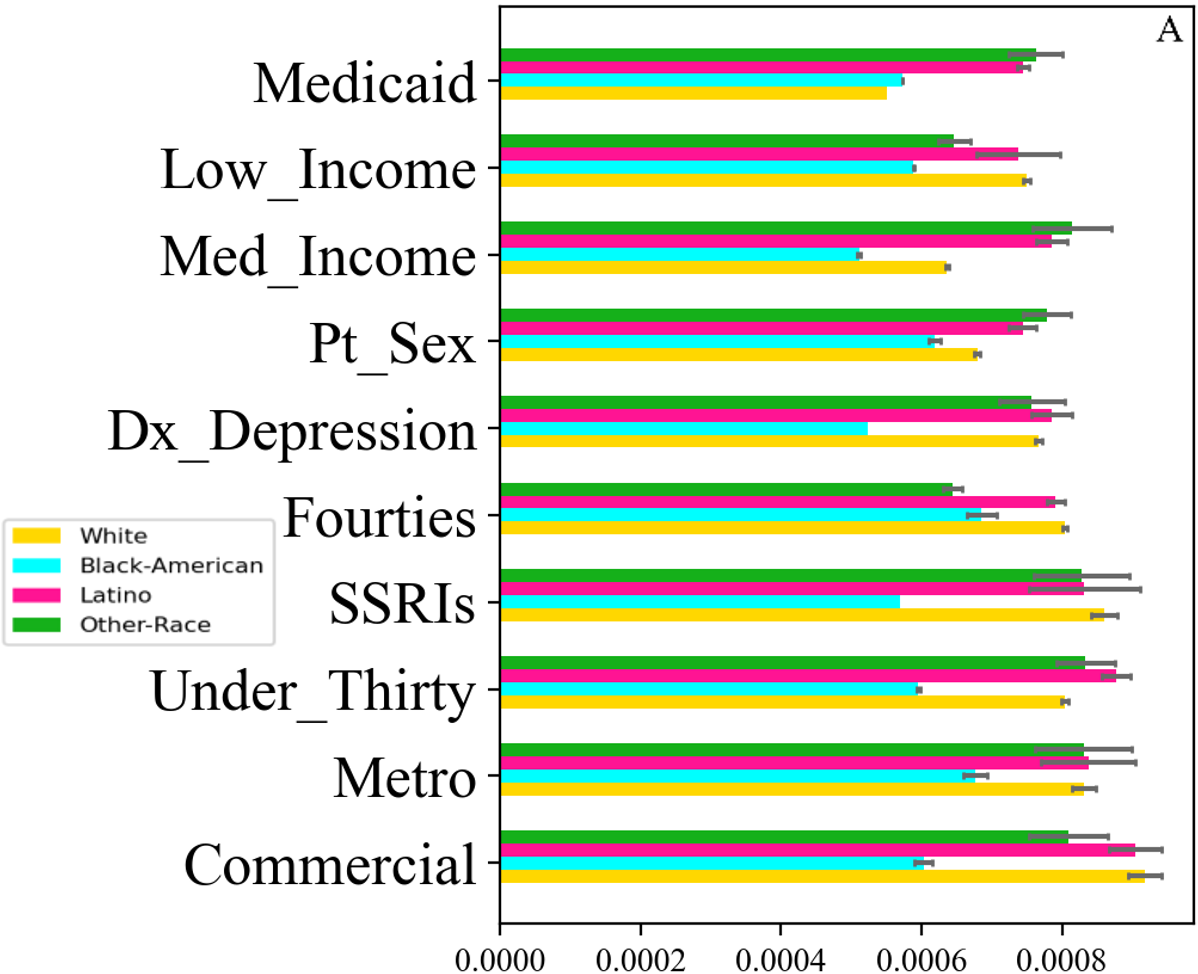}\hfill
  \includegraphics[width=.5\linewidth, height=.25\textheight]{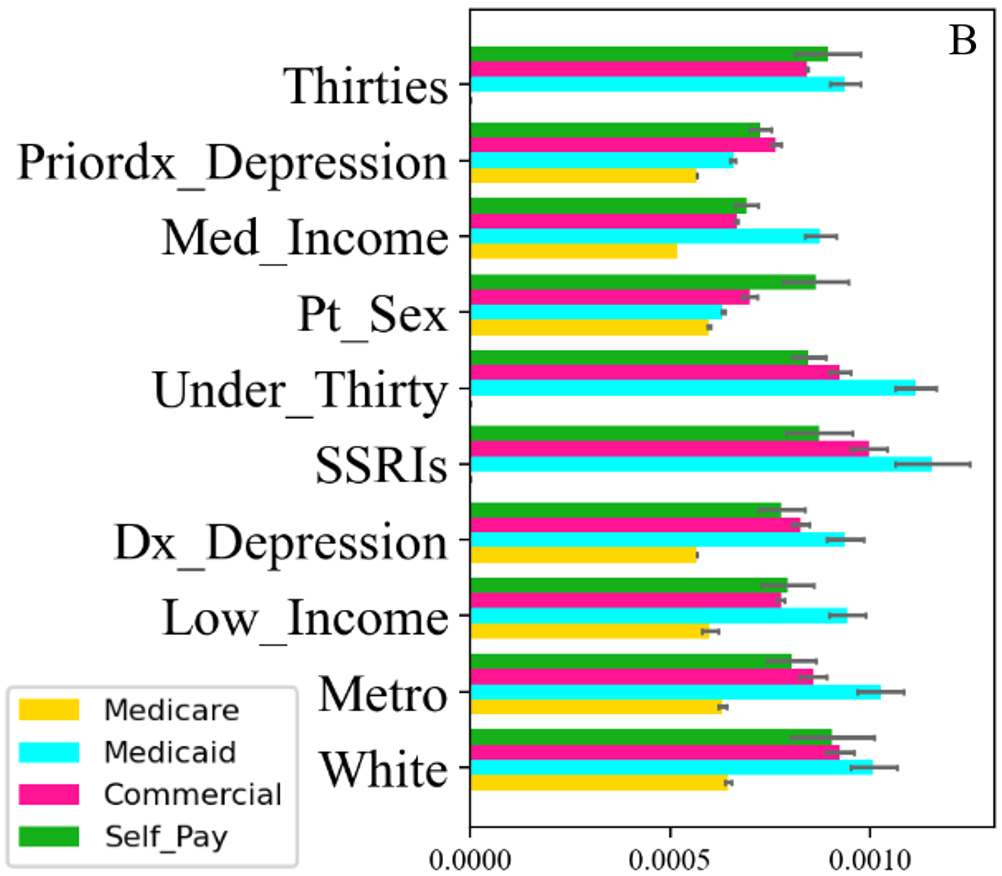}
  
  \end{subfigure}\par\medskip
  \begin{subfigure}{\linewidth}
  \includegraphics[width=.5\linewidth, height=.25\textheight]{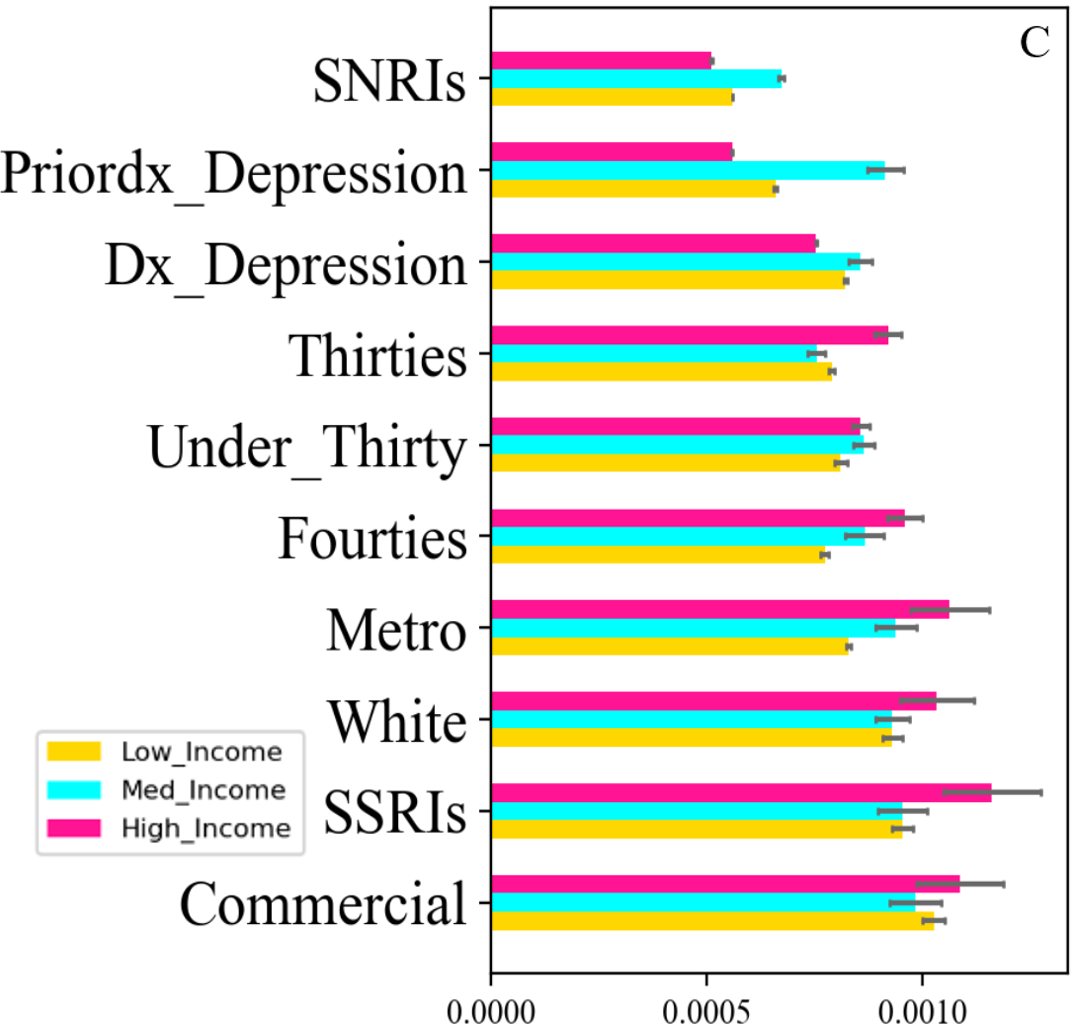}\hfill
  \includegraphics[width=.5\linewidth, height=.25\textheight]{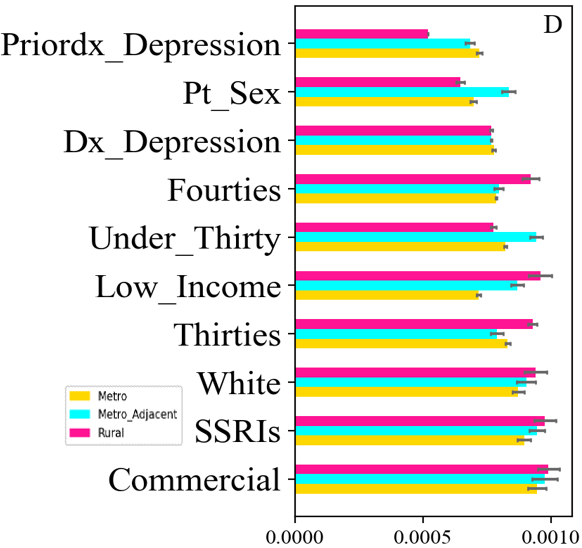}
  
  \end{subfigure}\par\medskip
  \begin{subfigure}{\linewidth}
  \includegraphics[width=.5\linewidth, height=.25\textheight]{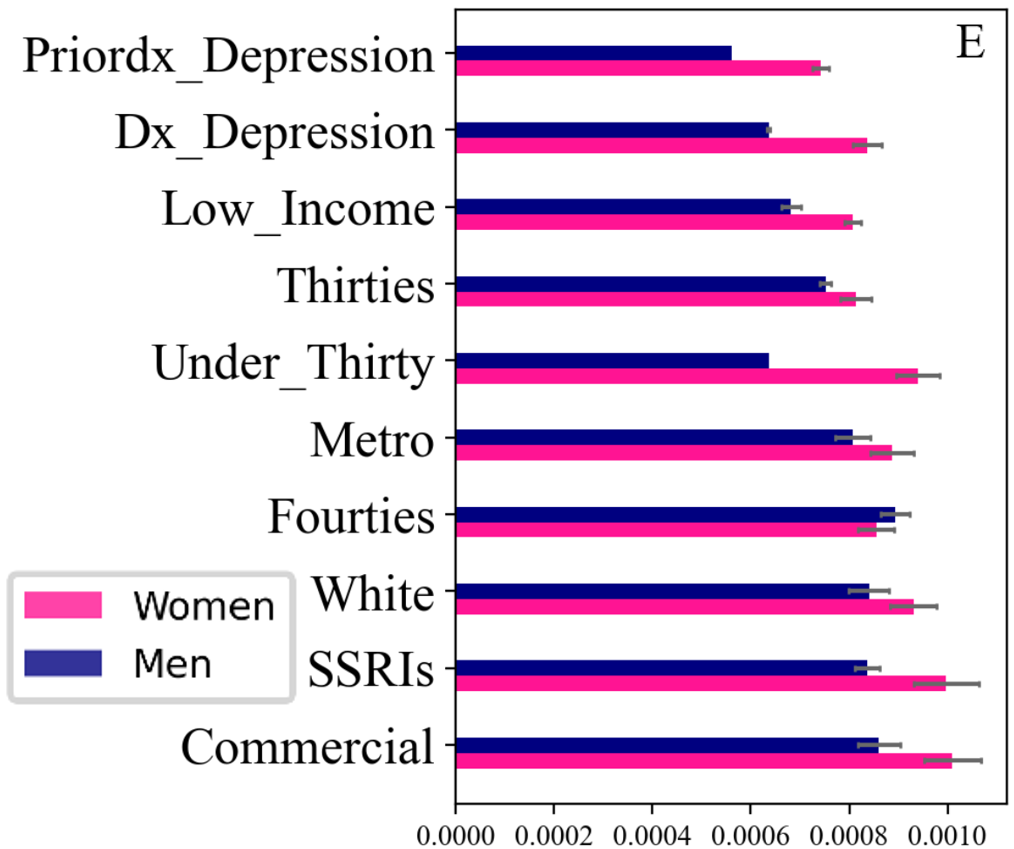}\hfill
  \includegraphics[width=.5\linewidth, height=.25\textheight]{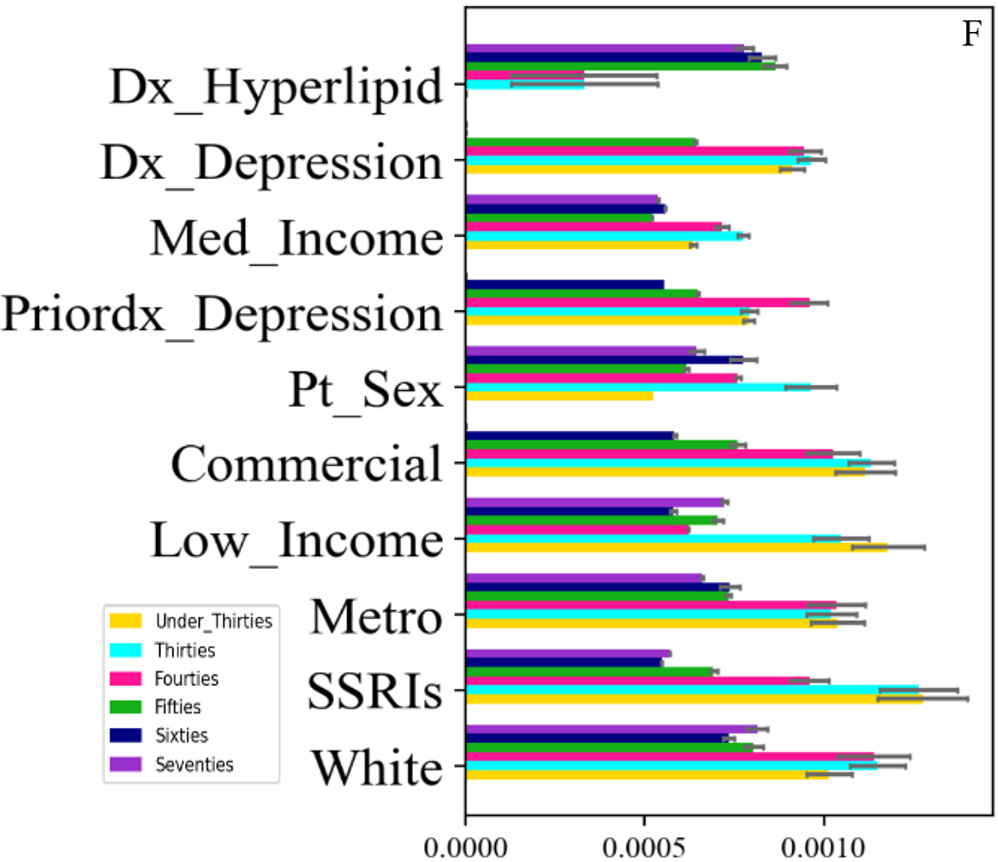}
  
  \end{subfigure}
  \caption{A comparison of weighted average quality of features association to obesity with highest summation across different categories of different groups including A) Race, B) Insurance types, C) Income-Level, D) Neighborhood areas, E) Genders, F) Age categories}
  \label{Fig:7}
\end{figure}




\end{appendices}



\end{document}